%% file: thesis.tex
\documentclass[12pt]{gatechthesis}
\usepackage{makecell}
\usepackage{array}
\usepackage{float}
\usepackage{latexsym}
\usepackage{amsmath}
\usepackage{amssymb}
\usepackage{mathtools}
\usepackage[ruled,vlined]{algorithm2e}
\usepackage{booktabs}
\usepackage{tabularx}
\usepackage{multirow}

\AtBeginDocument{%
}
\usepackage{xurl}
\title{Toward Robust, Contact-aware, and Autonomous Humanoid Loco-manipulation}
\author{Zhaoyuan Gu}
\approvaldate{July 22, 2026}
\school{George W. Woodruff School of Mechanical Engineering}
\department{College of Engineering}
\degree{Doctor of Philosophy}
\bibliography{references}

\begin{document}

\input{abbrevs}

% Register all acronym entries before front matter so the glossary write
% does not create an otherwise blank page before the acronym list.
\glsaddall

\makeTitlePage{August}{2026}
\input{approvalPage.tex}
% \makeEpigraph{I'm not super. Any talents I have, I worked for -- it seems a long time since I thought of myself as a hero.}{Oliver Queen}
% \makeDedication{For my cousin Kara}

\begin{frontmatter}
    \input{acknowledgments}
    \makeTOC
    \makeListOfTables
    \makeListOfFigures
    \makeListOfAcronyms
    \input{summary}
\end{frontmatter}

\begin{thesisbody}
    \input{chapters/introduction}
    \input{chapters/LTL-BT}
    \input{chapters/STL-MPC}
    \input{chapters/tactile-humanoid}

\input{chapters/PPF}
    % \input{chapters/foothold-RL}  % content merged into REFINE-DP sec:RL_train
    \input{chapters/REFINE-DP}
    \input{chapters/conclusion}
    \makeBibliography
\end{thesisbody}

\end{document}

%% file: abbrevs.tex
%
% --- Locomotion / formal methods ---
\newacronym{ltl}{LTL}{Linear Temporal Logic}
\newacronym{stl}{STL}{Signal Temporal Logic}
\newacronym{mpc}{MPC}{Model Predictive Control}
\newacronym{to}{TO}{Trajectory Optimization}
\newacronym{nlp}{NLP}{Nonlinear Program}
\newacronym{lipm}{LIPM}{Linear Inverted Pendulum Model}
\newacronym{com}{CoM}{Center of Mass}
\newacronym{wbc}{WBC}{Whole-Body Control}
\newacronym{pabt}{PABT}{Perturbation-Aware Behavior Tree}
\newacronym{ows}{OWS}{One-Walking-Step}
\newacronym{pd}{PD}{Proportional-Derivative}
\newacronym{caren}{CAREN}{Computer-Aided Rehabilitation Environment}
% --- Learning ---
\newacronym{rl}{RL}{Reinforcement Learning}
\newacronym{rlft}{RLFT}{Reinforcement Learning Fine-Tuning}
\newacronym{bc}{BC}{Behavior Cloning}
\newacronym{dagger}{DAgger}{Dataset Aggregation}
\newacronym{ppo}{PPO}{Proximal Policy Optimization}
\newacronym{gae}{GAE}{Generalized Advantage Estimation}
\newacronym{mdp}{MDP}{Markov Decision Process}
\newacronym{mlp}{MLP}{Multilayer Perceptron}
\newacronym{ood}{OOD}{Out-of-Distribution}
\newacronym{mbc}{MBC}{Model-Based Controller}
\newacronym{mar}{MAR}{Model-Assumption-based Regularization}
\newacronym{ppf}{PPF}{Pre-training and Preservative Fine-tuning}
\newacronym{sr}{SR}{Success Rate}
% --- Diffusion policy ---
\newacronym{dp}{DP}{Diffusion Policy}
\newacronym{dppo}{DPPO}{Diffusion Policy Policy Optimization}
\newacronym{refinedp}{REFINE-DP}{Reinforcement learning Fine-tuning of Diffusion Policy}
% --- Manipulation / tactile ---
\newacronym{umi}{UMI}{Universal Manipulation Interface}
\newacronym{wtumi}{WT-UMI}{Whole-Body Tactile Universal Manipulation Interface}
\newacronym{vr}{VR}{Virtual Reality}
\newacronym{dtw}{DTW}{Dynamic Time Warping}

%% file: approvalPage.tex
% In \begin{approvalPage}{N}, the parameter N is the number of members in the committee. If this is less than 4, the layout of the page is single-column rather than two-column, so change the value accordingly.

\begin{approvalPage}{5}

% Add people in the following format:
% \committeeMember{Member Name}{Member Department/Position}{Member Affiliation}

\committeeMember{Dr. Ye Zhao}{Woodruff School of Mechanical Engineering}{Georgia Institute of Technology}
\committeeMember{Dr. Maegan Tucker}{School of Electrical and Computer Engineering and Woodruff School of Mechanical Engineering}{Georgia Institute of Technology}
\committeeMember{Dr. Sehoon Ha}{School of Interactive Computing}{Georgia Institute of Technology}
\committeeMember{Dr. Patrick Wensing}{Aerospace and Mechanical Engineering}{University of Notre Dame}
\committeeMember{Dr. Guanya Shi}{Robotics Institute and the School of Computer Science}{Carnegie Mellon University}

\end{approvalPage}

%% file: acknowledgments.tex
\begin{acknowledgments}

This dissertation would not have been possible without the support, guidance, and generosity of many people who shaped my journey.

I owe my deepest gratitude to my advisor, Ye Zhao, for giving me the opportunity to pursue my Ph.D. and for supporting me throughout the years, especially during the difficult early period of the pandemic. His guidance on every research paper, from forming the initial idea to polishing the final manuscript, has shaped my research trajectory. He also taught me how to organize workshops and manage a large research group, and I will carry these skills into my future career. Most importantly, his enthusiasm for robotics, intellectual rigor in writing, and genuine commitment to student growth have profoundly influenced the way I collaborate and conduct research.

I sincerely thank the members of my committee, Dr. Maegan Tucker, Dr. Sehoon Ha, Dr. Patrick Wensing, and Dr. Guanya Shi, for their thoughtful feedback and for the time they dedicated to strengthening this dissertation. I have greatly enjoyed our collaborations on papers and shared projects. I am also grateful for the opportunity to collaborate or become friends with your outstanding students, whose excellence reflects the supportive and inspiring research labs you have built.

My research has benefited greatly from close collaborations with many distinguished colleagues. I am deeply grateful to Dr. Yongxin Chen, Dr. Seth Hutchinson, Dr. Greg Sawicki, Dr. Jennifer Leestma, Dr. Aaron Young for sharing robots and equipments across labs within Georgia Tech. 
I also thank my collaborators on the humanoid loco-manipulation survey, including Dr. Junheng Li, Dr. Wenlan Shen, Dr. Wenhao Yu, Dr. Zhaoming Xie, Dr. Stephen McCrory, Dr. Abdulaziz Shamsah, Dr. Xianyi Cheng, Dr. Robert Griffin, Dr. C. Karen Liu, Dr. Abderrahmane Kheddar, Dr. Xue Bin Peng, Dr. Yuke Zhu, Dr. Quan Nguyen, Dr. Gordon Cheng, and Dr. Huijun Gao, for their insightful discussions and contributions to creating a valuable resource for the broader humanoid robotics community.

I am also grateful to many talented colleagues.
When I began my first project, Nathan Boyd, Abdulaziz Shamsah, and Jonas Warnke offered generous help and contributed to the early chapters of my work on formal methods and locomotion. 
I had a fruitful experience with Fukang Liu, Hyunyoung Jung, and Feiyang Wu for their leadership and for bringing me along on the reinforcement learning journey. The work on WT-UMI would not have been possible without the core contributions of Jaehwi Jang. Jaehwi, you are gracious, polite, and selfless, and your hard work will take you far.

When I started my Ph.D., I had little experience leading a research team. I am thankful to Wei Pao, William Yates, Rongming Guo, and Yuntian Zhao for teaching me important lessons in project leadership while in the trenches. 

I am especially grateful to the large team that supported me during the later stage of my Ph.D. Their dedication, generosity, and countless hours of collaborative work brought projects and demonstrations to fruition.
I sincerely thank Yipu Chen, Zimeng Chai, Alfred Cueva, Thong Nguyen, Yifan Wu, Huishu Xue, Minji Kim, Matthew Kim, Junjie Sheng, Ojas Mediratta, Davin Doan, Andrew Collins, Himank Galundia, Isaac Legene, Ayan Barula, and Anand Nagpurkar for their support in hardware prototyping, data collection, and experiments behind our diffusion policy and tactile sensing efforts. I was also fortunate to mentor and collaborate with a large undergraduate team over the years. To the students I had the privilege of mentoring: witnessing your curiosity and growth continually inspired me. I learned as much from you as I hope you learned from our projects, and I am proud of all that you have accomplished.

Beyond research, my labmates and friends taught me lessons no paper could. Xuan Lin and Fukang Liu showed me what passion for research looks like and how to assign credit properly; Abdulaziz Shamsah showed me how to stay calm, courteous, and persistent under pressure. I thank Ziyi Zhou, Jiming Ren, and Wei Zhu for the house parties and karaoke nights; Luke, Nathan, and Jonas for showing me the joy of biking and taking me on adventures, from escape rooms to musicals; Jesse Jiang for his witty humor and kind banter; Max for the casino nights, bonfire talks, and waterfall adventures. Max, you are one of the kindest people I have ever met. I also thank Zhigen for introducing me to biohacking and bodybuilding supplements, and Jeremiah Coholich for teaching me yoga and giving me the courage of cliff jumping. I am especially thankful to Hongchang Zhu for his consistent and selfless support. I am fortunate to have someone who will always stand by my side, and I treasure the moments we shared camping and tubing on the river. I am also deeply grateful for the long-standing friendship with Sida Wang, for the trips we took on together and the countless hours spent sharing the deepest struggles and brightest hopes. To the many friends at the LIDAR Lab and IRIM who offered emotional support, encouragement, and companionship over the years, thank you.

I would also like to thank my mentor and manager, Grant and Corey, at Figure AI. Thank you for supporting the scope of my project and for consistently checking in on me throughout the summer, even when you were extremely occupied. I deeply appreciate your sense of responsibility toward team members, dedication, and professionalism at work.

Last but not least, I am grateful to my parents and my grandmother for their unwavering belief in me and for their emotional and financial support throughout the years. From my parents, I learned the meaning of unconditional love and its immense power. Their diligence has always motivated me to work harder, and their value of honesty has form the foundation of everything I have achieved. My grandmother’s wisdom and warmth have been a constant source of strength, and her influence stays with me in every step forward.

Finally, this dissertation work was supported by the National Science Foundation (NSF), the Office of Naval Research (ONR), the Georgia Tech Research Institute (GTRI), and the Institute for Robotics and Intelligent Machines (IRIM). I am sincerely grateful for the resources and support that made this research possible.

\end{acknowledgments}

%% file: summary.tex
\begin{summary}
% What to solve and why
Humanoid robotics is advancing at an unprecedented pace, raising public expectations that robots may soon operate in human-centered environments and relieve people from dull, dirty, and dangerous tasks. Yet this promise remains limited by a persistent gap between the laboratory demonstrations and the reliability required for real-world deployment. To operate beyond controlled settings, humanoid robots must withstand unexpected disturbances, reason through uncertain contact, and execute reliably with limited sensory feedback. These challenges also reveal a central opportunity: robustness to disturbances provides the foundation for deploying humanoid robots at scale in human environments, while versatile and autonomous manipulation delivers their practical real-world value. To meet the growing demand for a reliable autonomous robotic workforce, this dissertation advances humanoid robots along two complementary dimensions: strengthening robust locomotion as a foundation for reliability and enhancing versatile manipulation as a source of practical value.

% How to solve
The core strategy of pursuing reliable humanoid autonomy is to unify locomotion and manipulation capabilities, but divide the framework into a planner-controller hierarchy. Guided by this strategy, this dissertation advances humanoid capability along three thrusts: robust locomotion, tactile-based whole-body manipulation, and autonomous loco-manipulation. 

The first thrust \emph{quantifies and enhances locomotion robustness} through formal methods. We pursue formal guarantees for locomotion robustness using Linear Temporal Logic (LTL) and Signal Temporal Logic (STL). These formal methods generate provable perturbation-resilient behavior with quantifiable robustness, demonstrating robust locomotion of the Cassie bipedal robot on dynamic terrains, stepping stones, and a naval ship.
%LTL enables reactive planning in a \textit{discrete} center-of-mass (CoM) phase space with a correct-by-construction guarantee, while STL defines a \textit{continuous} robustness degree over the CoM state and incorporates it into an online Model Predictive Control (MPC) problem. 

The second thrust extends traditional table-top finger-tip manipulation to \emph{force-aware whole-body manipulation through large-area tactile sensing}. Through a tactile-sensing suit, WT-UMI, we collect tactile-rich human demonstrations and convert them into robot-executable motion-force plans, enabling the G1 humanoid to manipulate deformable and bulky objects and collaborate with humans during transport.

%We introduce the Whole-Body Tactile Universal Manipulation Interface (WT-UMI) to capture natural human demonstrations with rich distributed tactile feedback. WT-UMI tackles a key bottleneck in learning from human data: the lack of robot-executable action labels. We address this by learning action labels under supervision from robot data while preserving the natural force information. This enables a motion-force planner that supports whole-body manipulation of deformable and bulky objects, as well as human--humanoid collaborative transport, on the G1 humanoid.

The third thrust \textit{enhances unified loco-manipulation through reinforcement learning fine-tuning}. Rather than pre-training on large quantities of demonstrations, this approach achieves a superior success rate over traditional behavior-cloned policies with less data. 
For locomotion, we develop a Pre-training and Preservative Fine-tuning (PPF) approach that adapts to slippery and uneven terrains on the Digit humanoid. For autonomous loco-manipulation, we propose REFINE-DP, which jointly fine-tunes a diffusion-policy planner and an RL-based whole-body controller to improve both task success rate and controller tracking accuracy on the Booster T1 humanoid.
% that coordinates whole-body motion, adapts to robot–environment interactions, and recovers from errors during long-horizon execution. Traditional Behavior Cloning (BC) approaches are often limited by the quantity and quality of the demonstration data. To overcome this limitation, we use reinforcement learning fine-tuning to expand the capabilities of pre-trained BC policies, achieving stronger performance while relying on substantially less demonstration data. 

Overall, this dissertation advances a broad spectrum of humanoid capabilities through a set of contributions: guaranteeing locomotion robustness, unlocking human data for large-area tactile-based whole-body manipulation, and enhancing loco-manipulation skills via reinforcement learning fine-tuning. Together, these contributions chart a path toward humanoid robots that move with stability, manipulate with autonomy, and interact with the world through rich physical contact.

%The first thrust leverages human knowledge through formal methods to achieve robust locomotion. The second thrust adopts human demonstrations through tactile imitation learning to enable contact-rich whole-body manipulation. The third thrust uses robot demonstrations for pre-training and rolls out interactions through reinforcement learning fine-tuning to improve locomotion and loco-manipulation.
\end{summary}

%% file: chapters/introduction.tex
\chapter{Introduction}

\section{Introduction}

Humanoid robots hold the promise of physical intelligence combining human-level mobility and manipulation capability. Their legged morphology carries them through spaces designed for humans, while their hands and arms equip them to operate tools, equipment, and facilities. This multi-purpose capability makes humanoids a compelling platform for relieving humans from dull, dirty, and dangerous work. 

Yet despite recent advances in hardware reliability and motion agility, reliable real-world autonomy remains elusive. Humanoid robots must move beyond following human-specified behaviors toward understanding diverse tasks, proactively perceiving their surroundings, and making and executing plans via whole-body motions and rich contacts. Such autonomy depends on the ability to maintain stability, engage in rich contact with surrounding objects and environments, and chain diverse skills over long horizons. The goal of this dissertation is therefore to \textit{advance robust, contact-aware, and autonomous humanoid loco-manipulation}.

Motivated by this goal, this dissertation develops methods that progressively expand humanoid capabilities from robust locomotion, to tactile-based whole-body manipulation, and ultimately to autonomous loco-manipulation. Each stage addresses a key challenge: for locomotion, we seek formal guarantees of robustness; for contact-rich whole-body manipulation, we explore the integration of large-area tactile sensing and force regulation; and for loco-manipulation, we address the combined challenges of limited task success and prohibitive data requirements.
% Together, these efforts advance humanoid robots toward reliable autonomy in human-centered environments.

\begin{figure*}[t]
\centerline{\includegraphics[width=0.99\textwidth]{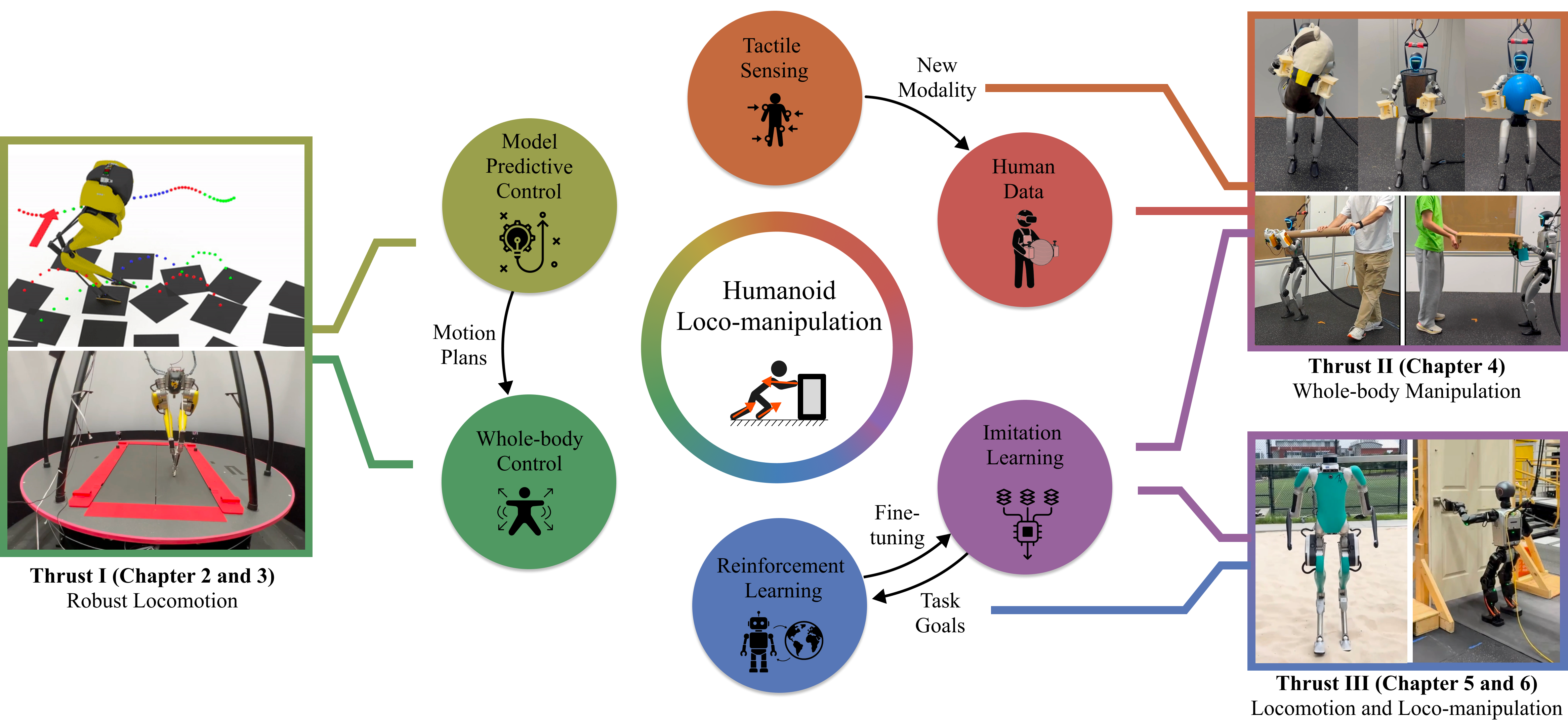}}
\caption{Roadmap of the dissertation toward advancing humanoid loco-manipulation. The three thrusts progress from robust locomotion, to tactile-based whole-body manipulation, and finally to autonomous loco-manipulation.}
\label{fig:roadmap}
\vspace{-0.05in}
\end{figure*}

\subsection{Thrust I: Quantifying and Enhancing Locomotion Robustness}

Robust locomotion is a prerequisite for humanoid autonomy, yet disturbance recovery remains challenging because the robot must plan recovery actions, reason about dynamic feasibility, and preserve sufficient safety margins in real time. Conventional locomotion controllers often stabilize walking gaits around periodic trajectories, but they typically lack an explicit representation of how much disturbance the system can tolerate before failure. As a result, robustness is often assessed after deployment rather than specified and optimized during planning.

The first thrust addresses this gap by embedding bipedal stability theory and locomotion dynamics into the planning process. Its guiding principle is that robustness can be systematically achieved by specifying, optimizing, and validating a desired system property. Specifically, this thrust derives a stability specification based on the center-of-mass keyframe, optimizes its boundedness as a measure of robustness, and validates that the resulting locomotion behavior is robust under comprehensive external disturbances.

\subsection{Thrust II: Force-Aware Whole-Body Manipulation through Large-Area Tactile Sensing}

The second thrust extends the problem from locomotion to contact-rich whole-body manipulation. Many everyday objects are bulky, deformable, or difficult to grasp with the hands alone. Manipulating such objects requires humanoids to coordinate distributed contact across the arms, torso, and hands. In these settings, success depends not only on how the robot moves, but also on where contact occurs and how force is applied.

Human demonstrations provide a natural source of guidance because they capture coordinated whole-body motion and intuitive contact strategies. However, the embodiment gap between humans and humanoids makes these demonstrations difficult to execute directly on the robot, and kinematic demonstrations alone lack the force information needed for stable physical interaction. To bridge this gap, this thrust converts human demonstrations into robot-executable supervision while preserving the tactile and force information required for contact-rich whole-body manipulation.

\subsection{Thrust III: Improving Pre-trained Humanoid Capabilities through Reinforcement Learning Fine-Tuning}
%Reinforcement learning fine-tuning can improve policies through interaction, but it benefits from structured initialization and mechanisms that preserve useful prior behavior. 

The third thrust addresses the limitations of imitation learning from demonstrations. Pure imitation learning requires large quantities of data, and the resulting policies remain limited by the distribution shifts encountered during deployment. These limitations become more pronounced in humanoid loco-manipulation, where locomotion drift and manipulation inaccuracy compound over long horizons and frequently lead to task failure.

Improving policy performance by scaling demonstrations alone is prohibitively expensive and cannot cover the full range of task variations encountered in deployment. This thrust therefore adopts reinforcement learning fine-tuning as a more effective and practical approach for improving pre-trained locomotion and loco-manipulation capabilities beyond their initial priors.

\subsection{The Unifying Theme}

This dissertation is organized around three thrusts that progressively advance humanoid capabilities toward autonomous loco-manipulation. This progression is reflected in the dissertation title, \emph{Toward Robust, Contact-Aware, and Autonomous Humanoid Loco-Manipulation}: robustness motivates the first thrust on locomotion, contact awareness motivates the second thrust on whole-body manipulation, and autonomy motivates the final thrust on loco-manipulation.

\ref{fig:roadmap} summarizes the roadmap for advancing humanoid loco-manipulation and provides a high-level overview of the major techniques used in each thrust. It also highlights two complementary methodological directions: model-based and learning-based approaches. Thrust 1 relies heavily on model-based methods, as models provide robot dynamics, formal specifications, and verifiable performance. In contrast, Thrust 2 and Thrust 3 both address manipulation, where contact-rich behaviors are difficult to specify. These thrusts therefore adopt learning-based methods to extract behaviors from demonstrations rather than manually designing detailed contact interactions. %In Thrust 2, tactile sensing further complements kinematic demonstrations by capturing force information that motion trajectories alone cannot represent.

The central theme of the dissertation is to convert structured model-based knowledge and experience into deployable robot capabilities. The structured knowledge includes formal specifications and robot dynamics, while the experience comes from both human and robot demonstrations. Each distinct source provides a unique benefit: (i) formal specifications and robot dynamics provide locomotion robustness guarantees; (ii) human demonstrations provide natural force supervision for contact-rich manipulation; and (iii) robot demonstrations provide high-quality loco-manipulation priors for behavior cloning. Together, these approaches demonstrate that structured knowledge can yield capabilities with rigor and reliability, while demonstration-based learning can produce behaviors that are versatile and adaptable.

\subsection{Dissertation Contributions}

The overarching contribution of this dissertation is a set of methods that convert distinct sources of specifications and experience into deployable humanoid capabilities, advancing robust, contact-aware, and autonomous loco-manipulation. The three thrusts contribute the following:

\textbf{Enhancing and guaranteeing locomotion robustness.} We guarantee locomotion stability through specifying, planning, and validating robot motions based on a formal specification. The formal specification is designed based on bipedal stability theory and locomotion dynamics and encoded as a quantifiable robustness degree in a trajectory optimization. This formulation generates dynamic recovery behaviors on Cassie under diverse perturbations and real-world field tests.

\textbf{Collecting and correcting human data for force-aware whole-body manipulation.} We show that human data provides a valuable prior for force-aware whole-body manipulation. We introduce WT-UMI~\cite{jang2026wtumi}, a system for collecting tactile-rich human demonstrations and humanoid teleoperation data. To bridge the embodiment gap between humans and humanoids, we develop a correction module that maps human motions into robot-executable actions while preserving calibrated contact forces. The resulting force-prediction planner supports contact-rich manipulation of bulky and deformable objects.

\textbf{Reinforcement learning fine-tuning of pre-trained humanoid policies.} We show that reinforcement-learning fine-tuning offers a more effective path to improving pre-trained policies than data scaling alone. Our methods, PPF~\cite{jung2025ppfpretrainingpreservativefinetuning} and REFINE-DP~\cite{REFINEDP}, improve locomotion and loco-manipulation policies beyond the demonstrated capabilities. On the Digit and Booster T1 humanoids, fine-tuning improves terrain robustness, task success, and controller tracking accuracy while significantly reducing pre-training data requirements.

\section{Dissertation Organization}

The remainder of this dissertation is organized around five technical chapters grouped into three thrusts.
\ref{chapter:LTL-BT} and~\ref{chapter:STL-MPC} develop the first thrust: enhancing and guaranteeing locomotion robustness through formal specifications and model-based control. \ref{chapter:LTL-BT} uses Linear Temporal Logic (LTL) to synthesize reactive recovery behaviors for bipedal locomotion, while \ref{chapter:STL-MPC} uses Signal Temporal Logic (STL) robustness as an optimization objective within Model Predictive Control (MPC).
\ref{chapter:tactile-humanoid} develops the second thrust: force-aware whole-body manipulation through large-area tactile sensing. It introduces WT-UMI, a whole-body tactile interface and learning framework that converts human tactile demonstrations into robot-executable, force-supervised manipulation policies.
\ref{chapter:PPF} and~\ref{chapter:DPFT} develop the third thrust: improving pre-trained humanoid capabilities through reinforcement learning fine-tuning. \ref{chapter:PPF} studies fine-tuning for robust locomotion from a model-based controller prior, and \ref{chapter:DPFT} extends this paradigm to autonomous loco-manipulation by jointly fine-tuning a diffusion-policy planner and a whole-body controller.
Finally, \ref{chapter:conclusion} synthesizes the capability progression across these thrusts, discusses how formal specifications, human demonstrations, and robot experience serve as sources of prior knowledge, and identifies future directions toward generalizable humanoid loco-manipulation.

%% file: chapters/LTL-BT.tex
\chapter{Linear Temporal Logic for Robust Bipedal Locomotion}
\label{chapter:LTL-BT}

\noindent\textbf{Dissertation Roadmap and Contribution of This Chapter.}
This chapter begins the first thrust of the dissertation: \emph{enhancing and formally guaranteeing locomotion robustness}. Robust bipedal locomotion under external perturbations requires more than motion planning and tracking. The robot must select a discrete recovery strategy and walking steps that return the system to stable walking while preserving whole-body dynamic feasibility and self-collision avoidance. This problem is challenging because typical task planning is symbolic and does not directly reason about continuous dynamics, while motion planning generates feasible motions but cannot guarantee push recovery.

To address this challenge, this chapter formulates push recovery as a hierarchical task-and-motion planning problem. Linear Temporal Logic (LTL) encodes recovery objectives and feasible walking-step transitions, and reactive synthesis produces correct-by-construction recovery decisions under bounded perturbations. Perturbation-aware behavior trees adapt these decisions between keyframes, while collision-aware full-body trajectory optimization verifies and refines the feasible transition set used by the symbolic planner. Together, these components connect formal task-level guarantees with executable recovery motions to achieve stable walking.

This chapter establishes the dissertation's first contribution: building the locomotion robustness as a foundation for humanoid autonomy. We transform structured model-based knowledge about recovery strategies, transition feasibility, and full-body dynamics constraints into a reactive planner with formal guarantees. The resulting framework motivates the next chapter, where temporal-logic specifications are extended from discrete correctness guarantees to quantitative robustness measures that directly guide continuous trajectory optimization.

\vspace{0.5em}
\section{Introduction}
% Original short introduction paragraph for reference:
% Robust bipedal locomotion requires correctness at both the task and motion levels. Task-level correctness requires the robot to select, in response to a perturbation, a sequence of recovery actions that eventually restores stability. Motion-level correctness requires each selected action to remain feasible under the robot's continuous full-body dynamics. 
% To achieve task-level correctness, symbolic task planning~\cite{Kress2009temporal} is often used to guarantee high-level objectives, but its decisions may be dynamically infeasible. Conversely, 
% reduced-order models, such as inverted pendulum models~\cite{Kajita2001}, support effective motion planning for push recovery, but it does not guarantee task completion or whole-body feasibility under aggressive perturbations. This chapter links both levels by encoding symbolic recovery strategy and motion feasibility into LTL specifications. The motionn feasibility refines the symbolic task transition through offline collision-aware trajectory optimization.

As legged robots are increasingly deployed in complex environments, the need for robots to accomplish tasks through task planning and decision making becomes more apparent. Although bipedal locomotion has been extensively explored at the motion-planning level, resilience to uncertainties and external disturbances at the task-planning level has been less developed. 

At the motion-planning level, ensuring feasibility and safety during online push recovery remains challenging. Many hip, ankle, and foot-placement strategies rely on reduced-order models~\cite{pratt2006capture,yi2011online,shafiee2019online}, which omit full-leg kinematics and therefore cannot guarantee self-collision avoidance during dynamic recovery. Although prior work has addressed self-collision, reactive safe motion, and robust walking behaviors~\cite{Chengju_et_al2017,Zhou_reactive2016,Hildebrandt2014,Dietrich2011,Nguyen2016,gong2019feedback}, real-time integration of these full-body safety constraints remains limited.

% At the motion-planning level, online reasoning of feasibility and safety during push recovery is still a challenge. Although bipedal locomotion has been extensively studied and various strategies, such as hip, ankle, and foot-placement strategies, have been developed to handle external perturbations~\cite{pratt2006capture,yi2011online,shafiee2019online}. However, many of these push-recovery strategies use reduced-order models, such as inverted pendulum or centroidal momentum models, making it difficult to guarantee leg self-collision avoidance. This challenge arises because full-leg kinematic constraints are difficult to solve directly in these reduced-order models. It is a strong assumption that a robot will never be in close contact with itself during highly dynamic locomotion. Prior work has considered self-collision under disturbances~\cite{Chengju_et_al2017}, reactive safe motions for high-dimensional robots~\cite{Zhou_reactive2016,Hildebrandt2014,Dietrich2011}, and behavior libraries for robust real-time walking~\cite{Nguyen2016,gong2019feedback}; however, few motion-planning strategies incorporate higher-level task planning.

At the task-planning level, temporal-logic-based reactive synthesis can generate formally guaranteed robot actions under environmental events~\cite{Kress2009temporal,liu2013synthesis}, and recent work has applied LTL to locomotion navigation over rough terrain~\cite{LTL_Nav_Kulgod,LTL_Nav_Warnke,shamsah2023TRO}. Applying reactive synthesis to push recovery, however, presents two challenges. First, symbolic foot-placement decisions must remain executable under the continuous whole-body dynamics of a high-degree-of-freedom robot. Second, the discrete transition system used for reactive synthesis represents perturbations at keyframe instants, leaving perturbations at non-keyframe instants outside the symbolic abstraction. Formal recovery decisions must therefore be coupled with motion-feasibility verification and a reactive interface between discrete keyframe decisions and continuous locomotion execution.

Hierarchically integrated task and motion planning (TAMP) can handle logical objectives and whole-body dynamics objectives simultaneously. However, unexpected errors or failures at the lower level can lead to expensive replanning at the task-planning level, while discrete task plans can result in infeasible motion plans. To address these challenges, we formulate nonperiodic, multistep recovery planning as a hierarchical TAMP problem for a bipedal robot subject to bounded external perturbations. Given the current center-of-mass (CoM) state and stance-foot information, LTL reactive synthesis generates a correct-by-construction keyframe planner that returns the robot to steady walking. Collision-aware full-body trajectory optimization verifies the dynamics and self-collision feasibility of the keyframe transitions and refines the transitions used for reactive synthesis. For perturbations between keyframes, a perturbation-aware Behavior Tree (BT) uses Riemannian recalculation to update the next keyframe in real time, reducing the need for keyframe replanning while maintaining consistency with the planned transition. The main contributions are as follows:
\begin{itemize}
    \item \textbf{Hierarchical LTL-BT task and motion planning:} We present a hierarchically integrated LTL-BT TAMP framework for dynamic locomotion that reacts to continuous environmental perturbations for resilient task execution.
    \item \textbf{Riemannian robustness margins:} We employ Riemannian manifolds to quantify locomotion keyframe robustness margins and design robust transitions enabled by the reactive task planner.
    \item \textbf{Collision-aware trajectory optimization:} We propose a collision-aware, kinodynamic TO that generates collision-free and non-periodic full-body motions and use this TO to refine feasibility specifications in reactive synthesis.
\end{itemize}

\section{Methods}
\subsection{Framework Overview}
\begin{figure*}[t]
\centerline{\includegraphics[width=0.99\textwidth]{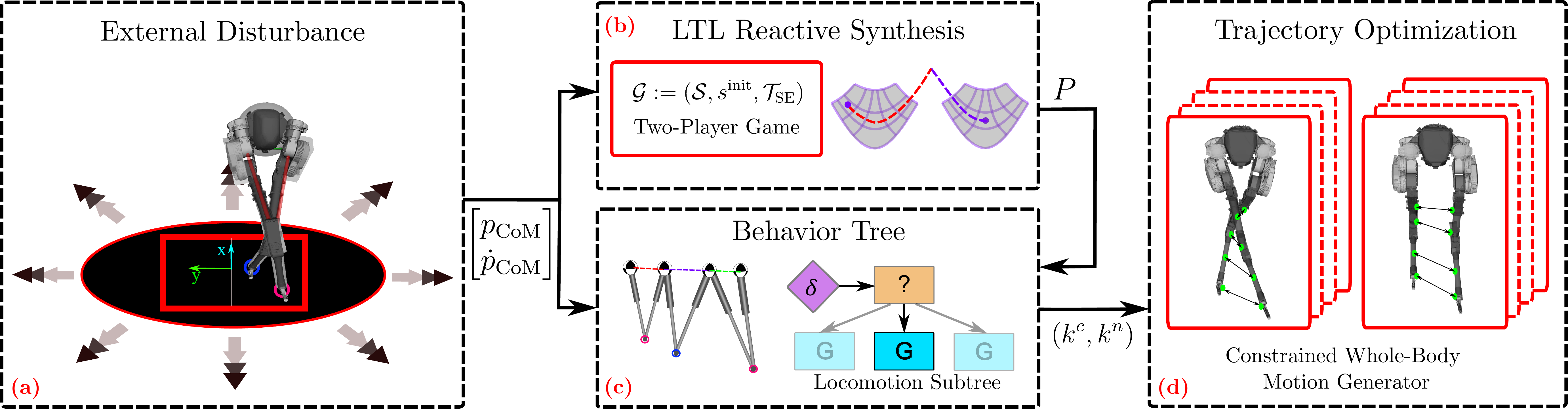}}
\caption{Block diagram of the LTL-BT framework. a) Experiments of Cassie disturbed during stable walking; b) The high-level task planner synthesis, employing an LTL two-player game; c) The behavior trees act as a middle layer that reactively executes subtrees based on real-time environmental disturbances; d) A whole-body motion planner is used to generate feasible motions and refine LTL specifications $\psi$. The high-level task planner and the phase-space planner are integrated in an \textit{online} fashion as shown by the solid black arrows.}
\label{fig:ltl_framework_overview}
\vspace{-0.05in}
\end{figure*}

\ref{fig:ltl_framework_overview} summarizes the four modules of our LTL-BT framework:
\begin{enumerate}
  \item \textbf{Keyframe Abstraction for Nonperiodic Locomotion:} Continuous walking is discretized into \emph{one-walking-step} (OWS) phases separated by apex keyframes. This abstraction enables symbolic reasoning over keyframes and supports nonperiodicity with variable step timing and apex velocity.
  \item \textbf{LTL Reactive Synthesis:} LTL specifications define a correctness-by-construction strategy that recovers from bounded perturbations while enforcing trajectory-optimization-refined feasibility specifications, as shown in~\ref{fig:ltl_framework_overview} (a).
  \item \textbf{Perturbation-Aware Behavior Tree (PABT):} A BT execution layer handles disturbances between keyframes by recalculating target keyframes using phase-space Riemannian metrics, allowing continuous-valued adjustments of the keyframe transition and its corresponding full-body motion.
  \item \textbf{Collision-Aware Kinodynamic Trajectory Optimization:} Offline trajectory optimization generates dynamically feasible and self-collision-free whole-body motion primitives. The resulting feasible transitions are encoded into specifications and used in the reactive synthesis.
\end{enumerate}

% \begin{figure}[t]
%   \centering
%   % \includegraphics[width=0.98\textwidth]{figures/ltl_framework_overview.pdf}
%   \caption{LTL–BT–TO framework for resilient push-recovery: (i) GR(1) reactive synthesis over keyframes, (ii) PABT for non-keyframe perturbations via Riemannian recalculation, and (iii) TO for collision-aware whole-body feasibility. (Figure to be created following the structure in the source paper’s Fig.~2.) :contentReference[oaicite:8]{index=8}}
%   \label{fig:ltl_framework_overview}
% \end{figure}

\vspace{0.5em}
\subsection{Keyframe-Based Nonperiodic Locomotion}
Compared to periodic walking, in which the robot repeats a cyclic motion, keyframe-based walking supports \emph{nonperiodic} locomotion with varying step length, velocity, and duration. The nonperiodic locomotion adapts more flexibly to external disturbances. In this study, nonperiodic locomotion is partitioned into one walking step (OWS)~\cite{shamsah2023TRO}, each bounded by apex keyframes. These apex keyframes encode the essential state of an OWS, including the sagittal and lateral apex position and velocity, as well as the stance foot index.

The keyframe-based locomotion allows a task planner to make decisions on the next keyframes, thereby satisfying specifications and guaranteeing locomotion stability (detailed in ~\ref{sec:LTL}). Once the next keyframe is determined, the contact-switch times $(t_1,t_2)$ can be obtained by forward/backward integrating reduced-order Center of Mass (CoM) dynamics (e.g., linear inverted pendulum model~\cite{Kajita2001}) between two consecutive keyframes~\cite{Zhao2017IJRR}. Because sagittal and lateral motions must switch foot contact simultaneously, the corresponding lateral foothold can be derived analytically from the lateral keyframe decision $(p_{\rm switch,l},\dot{p}_{\rm switch,l})$:
\begin{equation}
p_{\text{foot},l}
=
p_{\text{switch},l}
+
\frac{\left(e^{2\omega_{\text{asym}}t_2}-1\right)\,\dot{p}_{\text{switch},l}}
{\left(e^{2\omega_{\text{asym}}t_2}+1\right)\,\omega_{\text{asym}}},
\quad 
\omega_{\text{asym}}=\sqrt{\frac{g}{h_{\text{apex}}}},
\label{eq:lateral_foothold}
\end{equation}
where $h_{\text{apex}}$ is the apex CoM height relative to the stance foot. This keyframe abstraction and planning strategy naturally support nonperiodic recovery behaviors, which are essential for handling external disturbances.

\begin{figure}[t]
\centerline{\includegraphics[width=.7\textwidth]{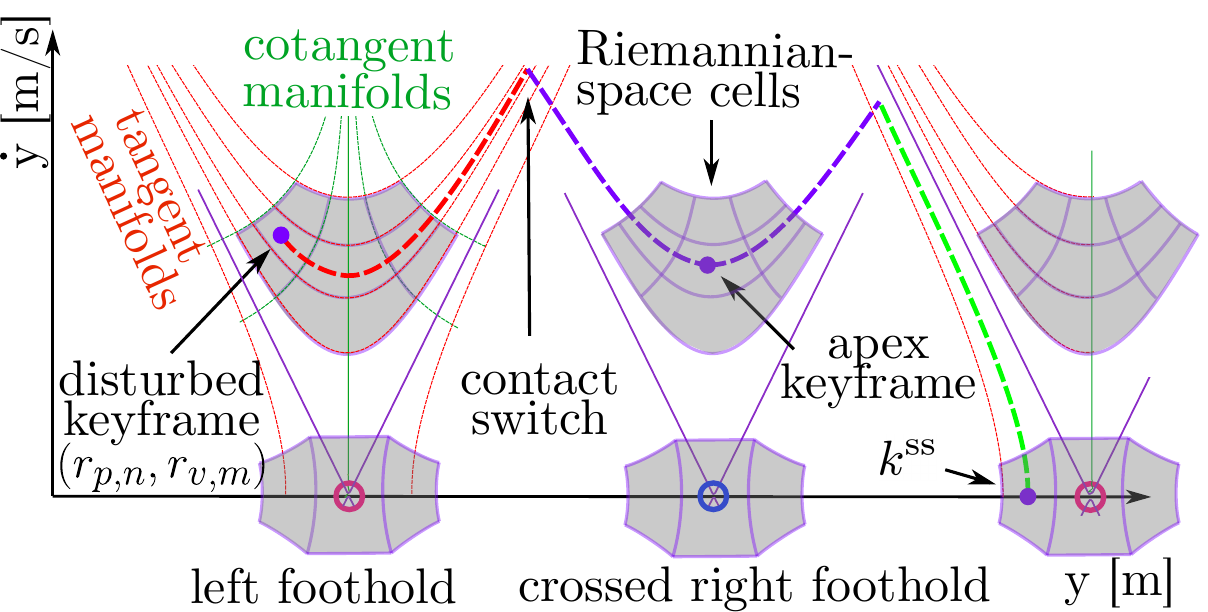}}
\caption{
An illustration of a phase-space Riemannian partition and non-deterministic lateral keyframe transition for disturbance recovery.}
\label{fig:riem}
\vspace{-0.05in}
\end{figure}

\subsection{Riemannian Partition and Keyframe Definitions}

To ensure that locomotion keyframes are dynamically feasible and directly executable by the low-level motion planner, we discretize the CoM phase-space around apex states into \emph{Riemannian partitions} (\ref{fig:riem}). These partitions are derived from the tangent and cotangent manifolds of the prismatic inverted pendulum dynamics, such that the discretization conforms to the intrinsic geometry of locomotion trajectories. Formally, we define the partition set as
\[
\mathcal{R}^{\mathrm{cell}} := \mathcal{R}^{\mathrm{cell}}_{\text{position}} \times \mathcal{R}^{\mathrm{cell}}_{\text{velocity}}
= \{r^{\mathrm{cell}}_{p,n},r^{\mathrm{cell}}_{p,z},r^{\mathrm{cell}}_{p,p}\}\times \{r^{\mathrm{cell}}_{v,z},r^{\mathrm{cell}}_{v,s},r^{\mathrm{cell}}_{v,m},r^{\mathrm{cell}}_{v,f}\},
\]
where $\mathcal{R}^{\mathrm{cell}}_{\text{position}}$ encodes negative/zero/positive CoM position relative to the stance foot, and $\mathcal{R}^{\mathrm{cell}}_{\text{velocity}}$ categorizes apex velocities into four ranges (zero, slow, medium, fast). The superscript ``cell'' indicates that these symbols denote discrete Riemannian cells designed for LTL planning. An LTL task planner then makes decisions on the next keyframe as a combination of sagittal and lateral partitions together with the stance foot index:
\[
\mathcal{K} := \mathcal{R}^{\mathrm{cell}}_s \times \mathcal{R}^{\mathrm{cell}}_l \times \{ \texttt{left}, \texttt{right}\}.
\]
Among these keyframes, we define a special set of \emph{steady-state keyframes} $\mathcal{K}_{\text{ss}}$, which correspond to apex states of nominal walking manifolds (with zero lateral apex velocity). The goal of the LTL task planner is to return to $\mathcal{K}_{\text{ss}}$ under bounded perturbations.

The Riemannian representation offers advantages over naive Euclidean discretization. By aligning partition boundaries with the tangent and cotangent manifolds of the locomotion dynamics, the resulting cell centers represent CoM keyframes connected by dynamically feasible transitions. 
The distance between Riemannian manifolds provides intuitive robustness margins, allowing middle-level behavior trees to execute keyframe transitions using systematic recalculation of keyframes (detailed in \ref{Sec:PABT}). 
This structure bridges symbolic and continuous planning: the LTL task planner reasons over discrete partitions, preserving high-level guarantees, while dynamics-informed trajectory optimization solves continuous-valued full-body motion between keyframe transitions. 
As a result, recovery strategies such as wider or crossed-leg steps emerge naturally from the manifold geometry rather than from heuristic rules, yielding a principled abstraction that unifies feasibility, robustness, and efficiency in reactive locomotion planning.

\subsection{LTL Specifications for Push Recovery}
\label{sec:LTL}
To formally guarantee locomotion task completion under environmental disturbances, we adopt the General Reactivity of Rank 1 (GR(1)) \cite{gr1}, a fragment of LTL. GR(1) provides correct-by-construction guarantees of the realizability of LTL specifications. An abstraction of the dynamic system provides a discrete description of the system and the dynamics of the environment. Built upon task specifications and an abstraction of dynamical systems \cite{kress2011correct, liu2013synthesis}, reactive synthesis leverages the discretized Riemannian partitions (apex keyframes) and reasons about next keyframes as system actions in response to environment perturbations. 

The tasks are represented by LTL specifications, which describe the temporal and logical goals of the system properties. An LTL formula operates over atomic propositions (APs) that can be \textsf{True} ($\varphi \vee \neg\varphi$) or \textsf{False} ($\neg \textsf{True}$). The formulas use logical symbols of negation ($\neg$), disjunction ($\vee$), and conjunction ($\wedge$). Temporal operators such as ``next" ($\bigcirc$), ``eventually" ($\Diamond$), and ``always" ($\square$) are used as extensions to the propositional logic.

The specifications describe two aspects: (i) \textbf{environment assumptions}, where external perturbations are modeled as discrete jumps of the CoM state into neighboring partitions, occurring only at keyframe instants and not more than once within two steps; and (ii) \textbf{system guarantees}, in the presence of perturbations, the keyframe state must return to a steady-state walking keyframe set $\mathcal{K}_{\text{ss}}$ within two steps: 
$$\square \big(k = \neg k^{\rm ss} \Rightarrow (\bigcirc k = k^{\rm ss})\vee(\bigcirc \bigcirc k = k^{\rm ss})\big).$$
\paragraph{Trajectory-Optimization-Refined Specifications.}
Feasibility of keyframe transitions is \emph{refined} by offline whole-body trajectory optimization (see \ref{sec:LTL_TO}); dynamically or kinematically infeasible transitions are pruned from the reactive synthesis. After TO refinement, we obtain all full-body-dynamics-feasible transitions offline and encode TO-refined specifications. An example of a TO-refined specification defines feasible next actions $a$ for a certain keyframe:
$$
\begin{aligned}
\square \big(k&=((r^{\mathrm{cell}}_{p,z},r^{\mathrm{cell}}_{v,m})_s,(r^{\mathrm{cell}}_{p,z},r^{\mathrm{cell}}_{v,m})_l, \textsf{right}) \Rightarrow \\ a&=((r^{\mathrm{cell}}_{p,z},r^{\mathrm{cell}}_{v,m})_s,(r^{\mathrm{cell}}_{p,z},r^{\mathrm{cell}}_{v,s})_l, \textsf{small}, \textsf{small})\\
&...\\
\vee a&=((r^{\mathrm{cell}}_{p,z},r^{\mathrm{cell}}_{v,m})_s,(r^{\mathrm{cell}}_{p,z},r^{\mathrm{cell}}_{v,m})_l,\textsf{small}, \textsf{medium}) \big)
\end{aligned}
$$
\vspace{-0.1in}

This induces a feasibility feedback loop: solving TO offline provides knowledge about feasible transitions, which inform the synthesized automaton to select only feasible low-level motions, ensuring consistency of the entire task and motion planning framework.

The reactive synthesis procedure constructs a two-player game between the system and the environment, where the environment applies bounded perturbations and the system selects feasible keyframe transitions. A \emph{winning strategy} for the system corresponds to a correct-by-construction task planner: for any admissible perturbations, the generated sequence of keyframes will always satisfy the specifications for safety and recovery.

\subsection{Perturbation-Aware Behavior Tree (PABT)}
\label{Sec:PABT}
BTs provide modular, reactive task execution under unexpected environmental changes~\cite{marzinotto2014towards,li2021reactive,colledanchise2018behavior,iovino2020survey} without requiring the large hand-designed state machines often used in finite-state-machine approaches~\cite{Park2013FSMterrain}. We therefore employ a perturbation-aware BT as the middle-level execution layer for handling perturbations between discrete keyframe decisions. Its fallback and sequence structure supports high-frequency action updates while maintaining consistency with the correctness guarantees of the discrete action plan.
The BT (i) checks whether the currently commanded transition $(k^c\!\rightarrow\!k^n)$ remains feasible given the observed $(p_{\text{CoM}},\dot{p}_{\text{CoM}})$, and (ii) when deviations occur, it \emph{locally recalculates} the next keyframe using a Riemannian guard to preserve feasibility.

\begin{figure}[t]
\centerline{\includegraphics[width=.7\textwidth]{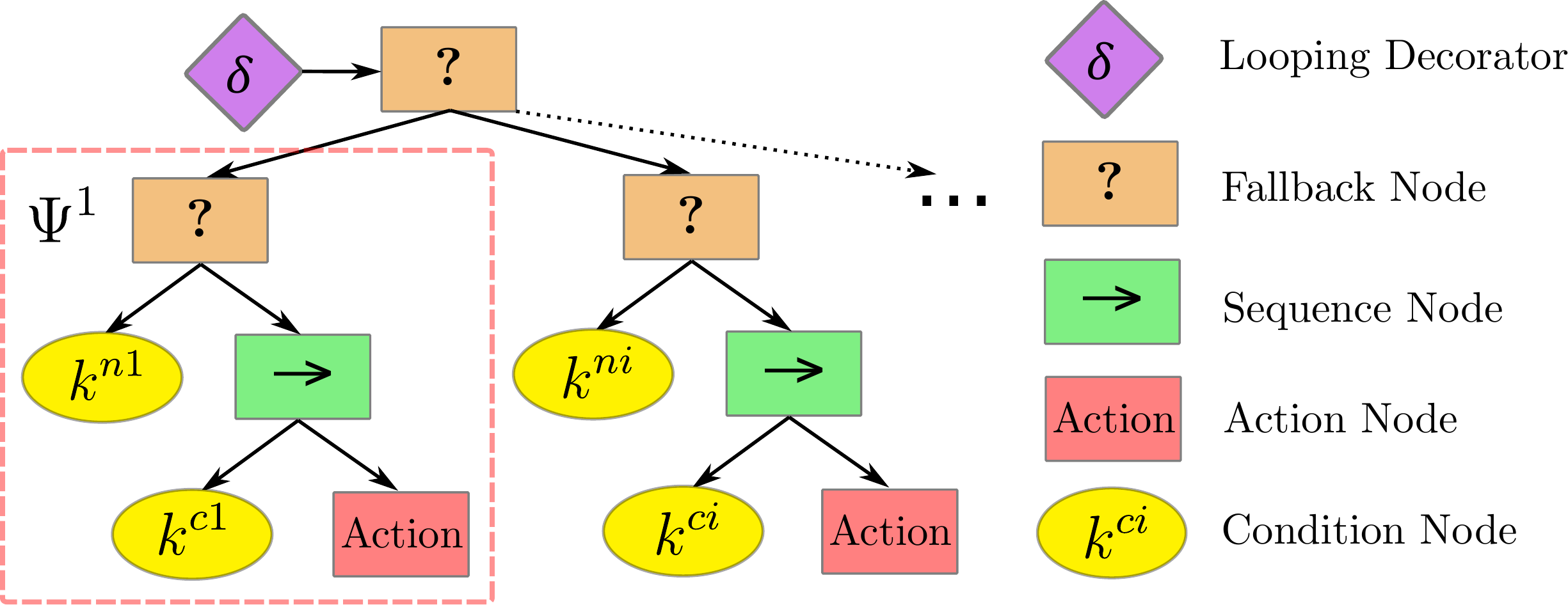}}
\caption{
An illustration of the PABT structure. The PABT groups a set of locomotion subtrees $\Psi^i$. Each subtree is a fallback tree that encodes a keyframe transition $(k^{c,i}, k^{n,i})$ and a Riemannian recalculation action.}
\label{fig:bt}
\vspace{-0.15in}
\end{figure}

As shown in \ref{fig:bt}, each locomotion subtree in the BT encodes the pre- and post-conditions of a target keyframe transition, together with an action node that either executes the associated motion primitive or triggers a keyframe recalculation. For each newly synthesized LTL action, a subtree is inserted, allowing the BT to expand online and reduce high-level decision-making overhead when the corresponding action subtree already exists in the tree. 

\subsection{Riemannian Recalculation (Position Guard)}
When an external perturbation drives the CoM state onto a new tangent manifold $\sigma'$, the BT invokes a Riemannian recalculation to update the desired apex keyframe. The recalculation follows the \emph{position guard}~\cite{Zhao2017IJRR}, which is a phase-space recovery rule used by the BT when the measured CoM state deviates from the nominal walking manifold. The next CoM apex is computed analytically as
\begin{equation}
\big(p_{\text{apex}},\dot{p}_{\text{apex}}\big)
=
\left(
p_{\text{foot}},\;
\sqrt{
\frac{\dot{p}_{\text{CoM}}^{\prime\,2} \pm \sqrt{ \dot{p}_{\text{CoM}}^{\prime\,4} - 4\,\omega_{\text{asym}}^{2}\,\sigma'} }{2}
}
\right),
\label{eq:riemannian_guard}
\end{equation}
where $p_{\text{foot}}$ is the stance-foot location and $\omega_{\text{asym}}$ is the asymptote slope defined in~\ref{eq:lateral_foothold}.

\begin{figure}[t]
\centerline{\includegraphics[width=.88\textwidth]{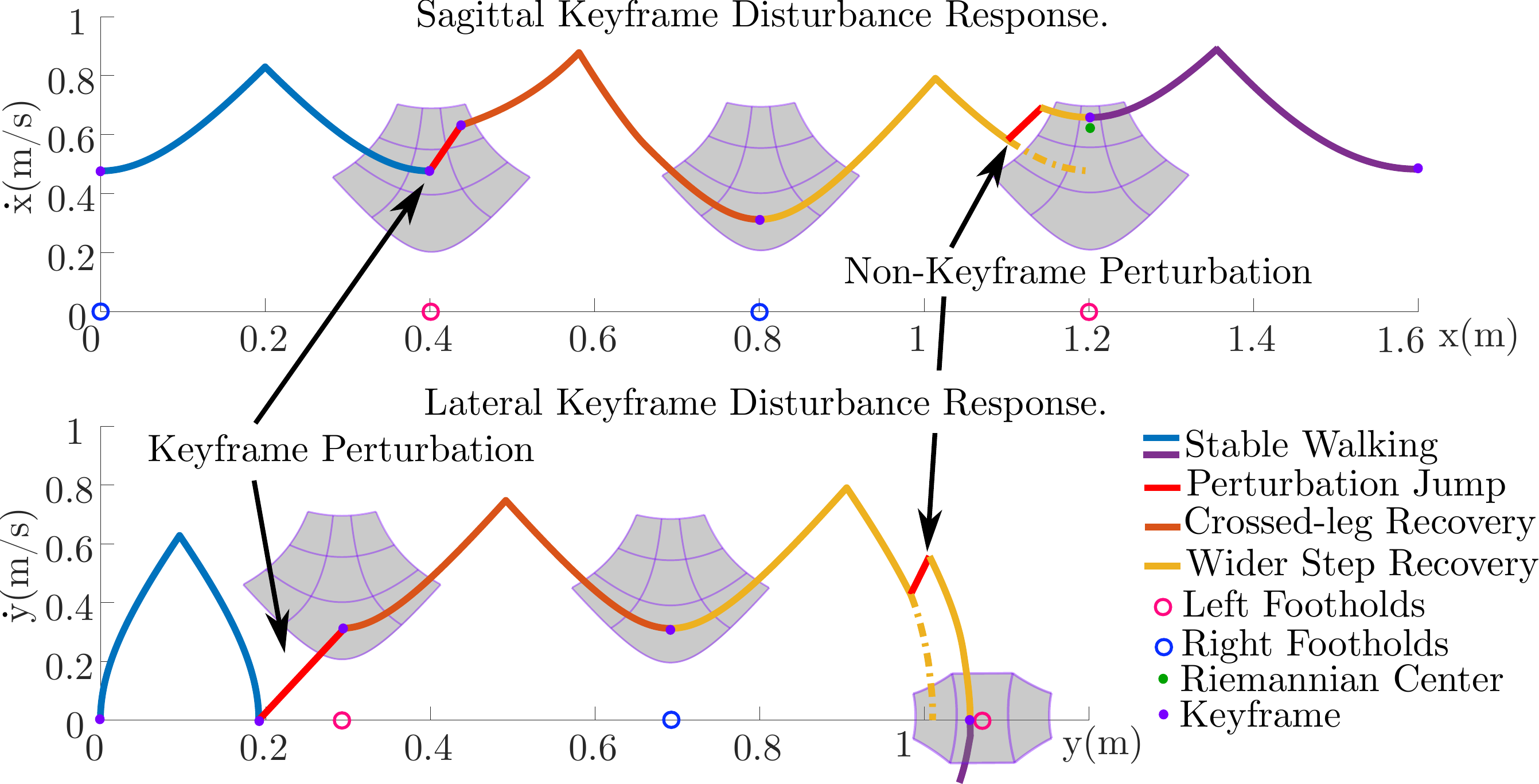}}
\caption{
Lateral and sagittal responses to diagonal disturbances at keyframe and non-keyframe instants while walking at $0.5$ m/s apex velocity. Each color represents a single step generated by the LTL-BT. }
\label{fig:ps_result}
\vspace{-0.05in}
\end{figure}

This recalculation has two important effects. First, it guarantees that the disturbed CoM trajectory is smoothly rerouted to a new feasible apex state rather than being forced back onto a predefined discrete cell center, as shown in~\ref{fig:ps_result}. Second, the updated apex state is still mapped to the nearest valid keyframe partition, ensuring that the high-level LTL strategy remains valid and still holds recovery guarantees. In this way, the BT and Riemannian guard preserve the correctness guarantees of the discrete logic layer while providing the flexibility to handle arbitrary perturbations occurring between keyframes. 

\subsection{Collision-Aware Kinodynamic Trajectory Optimization}
\label{sec:LTL_TO}
We solve trajectory optimization (TO) problems formulated as Nonlinear Programs (NLPs) to generate motion primitives and verify feasibility. 
\begin{align}
\min_{\{x_i^j,u_i^j\}} \;& \sum_{j=1}^{D}\sum_{i=0}^{N_j} \Omega_j \,\|u_i^j\|_2^2 
\label{eq:nlp_obj}\\
\text{s.t.}\;& H^j(x_i^j)\dot{x}_i^j+V^j(x_i^j)+G^j(x_i^j)=u_i^j && \text{(dynamics)} \nonumber\\
& x_{0}^{j+1}=\Delta^j\!\left(x_{N_j}^{j}\right) && \text{(impact/reset)} \nonumber\\
& \lambda_{c,z}\ge 0,\;\; \|\lambda_{c,xy}\|_\infty \le \mu \lambda_{c,z} && \text{(friction)} \nonumber\\
& C_{\text{kin}}^{j}(x_i^j)\le 0 && \text{(kinematics)} \nonumber\\
& C_{\text{col}}^{j}(x_i^j)\le 0 && \text{(self-collision)} \nonumber\\
& C_{\text{key}}^{j}(x_i^j,u_i^j)=0 && \text{(keyframe boundary)}. \nonumber
\end{align}
where the domains include $D=2$ continuous single-stance phases and one velocity reset map. Each single stance contains $N_j$ nodes, which represent a state-control pair $n^j_i=(x^j_i, u^j_i)$ at the $i^{\rm th}$ instant; the state is $x = [q;\dot{q}]$, with $q$ denoting the robot's generalized coordinates. 
The NLP above solves the optimal state-control trajectory $X^* = \{n^{j*}_i\}$ by minimizing the pseudo energy $\mathcal{L}(\cdot) = || u_i ||^2$ with weights $\Omega_j$ while enforcing the physical constraints of the robot.
We use the full-body dynamics of a bipedal robot as dynamics constraints. 
Self-collision is prevented by enforcing signed-distance inequalities between selected geometric pairs on the two legs, as shown in~\ref{fig:ltl_framework_overview} (c):
\begin{equation}
C_{\text{col}}^{j}(x_i) \;=\; 
\|d_{\min}\|_2^2 - \big\|\,\mathrm{FK}_\ell(x_i)-\mathrm{FK}_r(x_i)\,\big\|_2^2 \;\;\le 0,
\end{equation}
where $\mathrm{FK}_\ell$ and $\mathrm{FK}_r$ denote the forward kinematics of the chosen points and $d_{\min}$ is the minimum allowable clearance. 
In addition, a keyframe boundary constraint ensures that the keyframe transition $(k^c \!\rightarrow\! k^n)$ prescribed by the task planner is dynamically realizable. If the TO problem is infeasible for a given keyframe pair, that transition is pruned from the set of admissible motions and excluded from the automaton during decision making.

% and keyframe boundary conditions match the LTL-commanded $(k^c\!\rightarrow\!k^n)$ transition. 
% The feasible transition set extracted from \eqref{eq:nlp_obj}–\eqref{eq:collision} prunes the GR(1) game graph.

\subsection{Task Planner Synthesis and Online Execution}
At each keyframe instant, the synthesized \emph{decision maker} estimates the current keyframe $k^c$ and rolls out a finite plan to a steady-state $k_{\text{ss}}$. Between keyframes, the PABT \emph{ticks} at the control rate, executing the matched subtree or invoking recalculation. Concise pseudocode summarizes the synthesis and execution process in \ref{alg:LTL_BT}.

\setlength{\textfloatsep}{0.5cm}
\setlength{\floatsep}{0.5cm}
\begin{algorithm}[t]
\SetAlgoLined
\SetNoFillComment
\textbf{Input:} PABT $\Psi$, Decision Maker $DM$, current $time$\;
\textbf{Set:} status = \textsf{success}\;
\While{ status == \textnormal{\textsf{success}} }{
    $k^c, time$ = StateEstimation()\;
    \If{ $time$ == \textnormal{\textsf{keyframe\_instant}}} {
        $P = DM(k^c)$\;
        \For {$(k^c, k^n)$ \textnormal{in} $P$} {
            $\Psi^c$ = LocomotionSubtree$(k^c, k^n)$\;
            $\Psi$.Insert($\Psi^c$)\;
        }
    }
    \tcc{PABT Riemannian Recalculation}
    $status$ = $\Psi$.Tick()\;
    %\tcp*{Riemannian Recalculation} %\Comment*[r]{Riemannianrecalculation}%\COMMENT{Riemannian Recalculation}
    $(k^{c}, k^{n})'$ = $\Psi$.GetModifiedTransition()\; 
}
\textbf{Output:} updated PABT $\Psi$, modified keyframe transition $(k^{c}, k^{n})'$.
\caption{Keyframe Decision Making and PABT Execution}
\label{alg:LTL_BT}
\end{algorithm}

% \vspace{0.5em}
\section{Results and Case Studies}

We validate the LTL-BT framework in simulation on a Cassie bipedal robot. The trajectory optimization (TO) is constructed using FROST-generated dynamics~\cite{Frost} and solved with IPOPT~\cite{IPOPT}. While TO refinement and reactive synthesis are performed offline, the perturbation-aware behavior tree (PABT) operates online at a fast control rate of $2$ {kHz}. Disturbances are modeled as instantaneous CoM velocity jumps and injected at different phases and directions of the gait cycle.

\subsection{Direction--Phase Sensitivity}
\begin{figure}[t]
\centerline{\includegraphics[width=.7\textwidth]{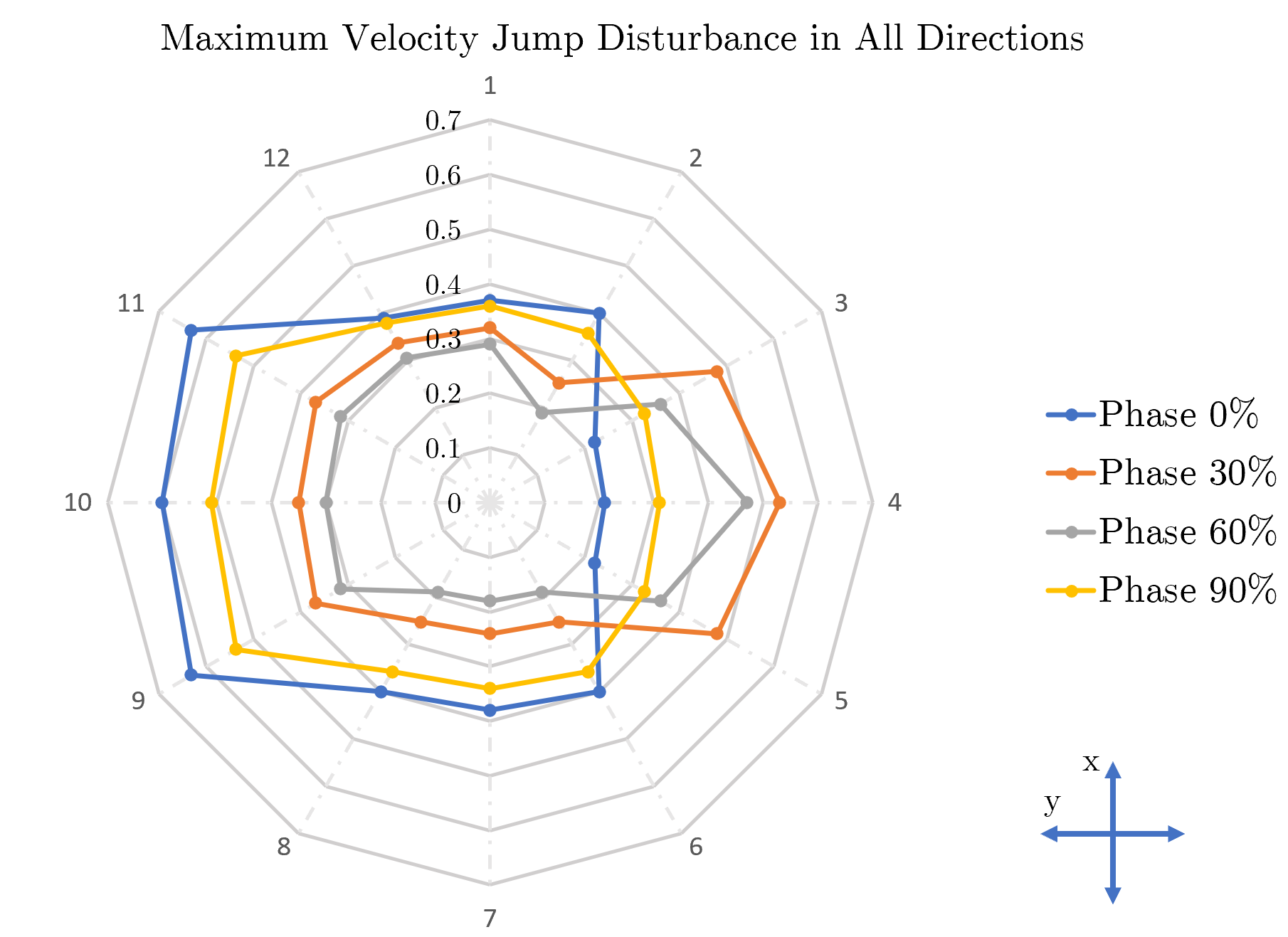}}
\caption{
Maximum allowable velocity change exerted on the CoM for a single step at $30^\circ$ increments. {The perturbation occurs at different phases during a right-leg stance.} Values on the left half result in single, wider-step recoveries, and values on the right half require crossed-leg maneuvers.}
\label{fig:LTL_spider}
\vspace{-0.05in}
\end{figure}

We evaluate the performance of our framework through a series of push-recovery studies. As shown in~\ref{fig:LTL_spider}, we compare the maximum recoverable impulse velocity across $12$ perturbation directions during one-walking-step (OWS). Perturbations are applied at four distinct phases: $\phi=0\%$ and $90\%$, which correspond to boundary phases near keyframe states, and $\phi=30\%$ and $60\%$, which lie closer to the contact-switch phase ($50\%$). The results reveal a clear phase-dependent recovery pattern: pushes occurring near keyframe states are better absorbed in the leftward direction, while disturbances applied closer to the contact switch are more effectively countered on the right, aided by increased lateral velocity at contact time. The observed asymmetry in lateral disturbance tolerance is primarily due to the kinematic limitations of the swing leg in crossed-leg recovery scenarios.

\subsection{Probabilistic Success Rates}

\begin{figure}[t]
\centerline{\includegraphics[width=.75\textwidth]{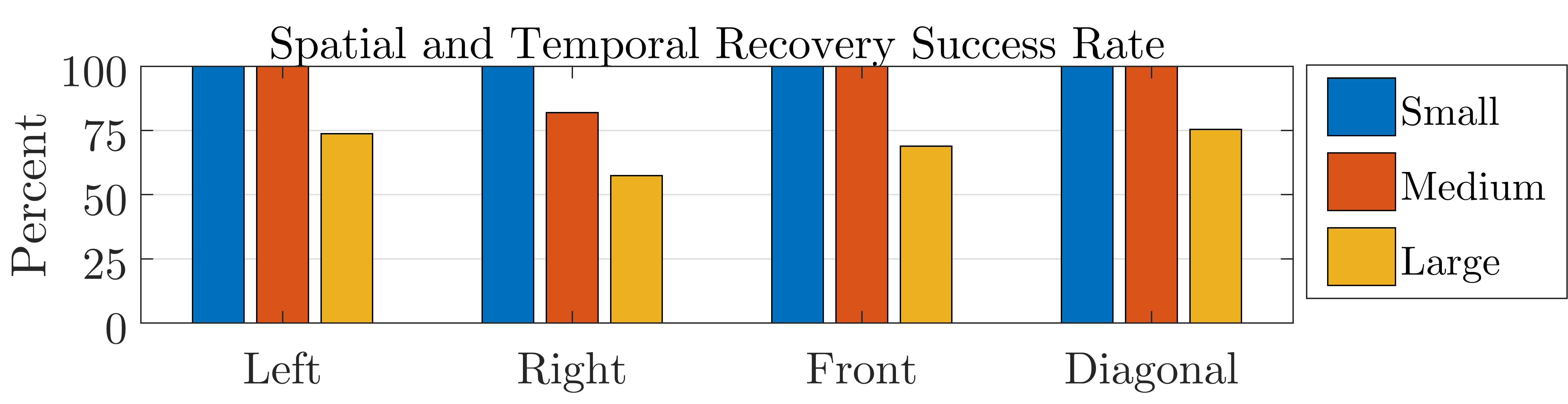}}
\caption{
Success rate of recovery motion when disturbances occur at any time during an OWS and from multiple directions. Three disturbance magnitudes are used: a) small $0.1$ m/s, b) medium $0.2$ m/s, and c) large $0.3$ m/s.}
\label{fig:success_rate}
\vspace{-0.05in}
\end{figure}

To systematically evaluate robustness, we perform $100$ Monte Carlo trials for diagonal disturbances of varying magnitudes ($0.1$, $0.2$, and $0.3$~m/s), evenly distributed across gait phases. The results in \ref{fig:success_rate} reveal that small perturbations are reliably rejected, while larger impulses cause failures concentrated near phases where the CoM velocity is already large in the disturbance direction (right stance: $\phi\leq10\%$ and $\phi\geq90\%$, left stance: $40\%\leq \phi \leq60\%$). In these cases, the imposed velocity jump adds to the ongoing CoM motion and can exceed the range that can be recovered within the planned stepping maneuver.

\section{Conclusion}

This chapter encoded push-recovery objectives and feasible locomotion transitions through LTL specifications. Reactive synthesis selected recovery actions with correctness-by-construction guarantees under bounded disturbances, while perturbation-aware behavior trees and collision-aware trajectory optimization connected discrete decisions to continuous full-body motion. The resulting task-and-motion-planning hierarchy enhanced locomotion robustness by selecting nonperiodic recovery strategies, including wide and crossed-leg steps, according to the robot state and perturbation.

This hierarchy, however, also exposes a limitation of LTL-based abstraction. The synthesized strategy reasons over discrete recovery actions and enforces specification satisfaction at the symbolic level, while the continuous foot trajectories, keyframe timing, and full-body feasibility are handled by downstream TO. As a result, the planner can certify whether a discrete transition is allowed, but it cannot measure how far a continuous trajectory is from violating a locomotion specification or use that margin as an optimization objective. This becomes restrictive when multiple continuous recovery trajectories satisfy the same symbolic action or when strict satisfaction is infeasible and the controller should still return the least-violating recovery motion.

The next chapter addresses this limitation using Signal Temporal Logic (STL), which specifies tasks directly over real-valued continuous trajectories and admits a \textit{robustness degree} that quantifies the degree of specification satisfaction. This robustness degree shifts the formal-methods formulation from selecting a correct symbolic recovery transition to optimizing continuous recovery trajectories, deeply coupling task and motion planning.

%% file: chapters/STL-MPC.tex
\chapter{Robust Locomotion by Logic: Signal Temporal Logic Guided Model Predictive Control}
\label{chapter:STL-MPC}

\noindent\textbf{Dissertation Roadmap and Contribution of This Chapter.}
This chapter completes the first thrust of the dissertation: \emph{enhancing and guaranteeing locomotion robustness}. Whereas LTL-BT in \ref{chapter:LTL-BT} achieves push recovery via planning on \textit{discrete} recovery transitions, this chapter employs Signal Temporal Logic (STL) to specify stability directly over \textit{continuous} trajectories at the motion-planning level. By embedding the robustness degree of stability specification into a trajectory optimization, the framework makes locomotion robustness both quantifiable and optimizable. The optimization additionally enforces dynamic feasibility and self-collision avoidance, eliminating the need for a separate offline feasibility check in \ref{chapter:LTL-BT}. It also solves motions in a real-time Model Predictive Control (MPC), obviating a separate reactive execution layer in LTL-BT. We validate our STL-guided MPC through comprehensive simulation and hardware perturbation experiments, as well as a field test aboard the \textit{Stiletto} vessel. Together, \ref{chapter:LTL-BT} and \ref{chapter:STL-MPC} demonstrate how formal specifications and dynamics models yield robust locomotion with verifiable guarantees grounded in locomotion stability theory.

%The optimized trajectories have guaranteed stability and ensure safety from self-collision.

\section{Introduction}
Robust locomotion requires more than reactive balance; it further requires guaranteed stability of a planned trajectory and how far it remains from failure. Traditional push-recovery methods use angular-momentum regulation~\cite{gong2019feedback} or predictive foot placement~\cite{Feng_dynamic_online, HLIP_TRO_XIONG}, but they generally lack formal task guarantees or a quantitative measure that can be optimized online. STL provides both dense-time task semantics~\cite{Donze_STL_origin} and a robustness degree that measures tolerance to specification violation~\cite{JerryThesis,Byl_robust_quantify}. Building on the task planning and robustness measurement capability of STL, this chapter embeds locomotion robustness in trajectory optimization (TO) for robust locomotion. The TO plans recovery motions in real time by maximizing locomotion robustness derived from the Riemannian safety region of Zhao et al.~\cite{Zhao2017IJRR}.

Among formal methods, Linear Temporal Logic (LTL) has been previously applied to bipedal locomotion planning~\cite{Gu_push} and reactive legged navigation~\cite{shamsah2023TRO}, but its hierarchical formulation decouples discrete decision-making from continuous motion planning, so a high-level recovery plan may become infeasible when executed by the low-level planner. STL instead encodes the task specification within TO that solves continuous motions online; this integration reduces the mismatch between task-level recovery decisions and dynamically feasible motion. 

To our knowledge, this work~\cite{gu2024robustlocomotionbylogic} is the first framework using STL-guided trajectory optimization for bipedal locomotion. By embedding STL specifications into a TO and solving it online in a Model Predictive Control (MPC) style, the framework generates recovery motions with guaranteed locomotion stability while enabling advanced behaviors such as crossed-leg maneuvers and terrain-agnostic push recovery. 
In addition, we explore omnidirectional perturbations using a Computer-Aided Rehabilitation Environment (CAREN) system~\cite{CAREN} and Bump'Em~\cite{BumpEm}, and comprehensively investigate their impact on locomotion robustness. The hardware validation further includes a real-world field test aboard the \textit{Stiletto} vessel, where Cassie walked on a moving ship deck under vessel-induced motion, external pushes, and uneven terrain blocks.

\begin{figure}[t]
\centerline{\includegraphics[width=0.99\textwidth]{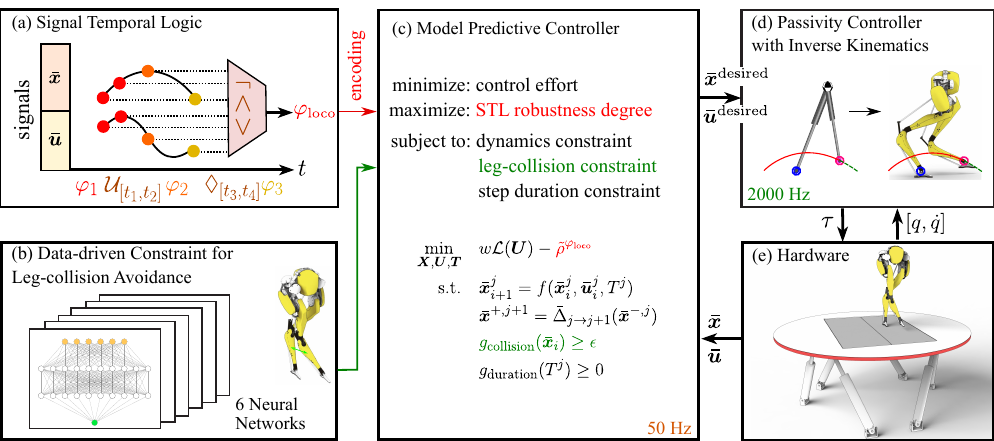}}
\caption{Block diagram of the STL-MPC framework. (a) The STL specification $\varphi_{\rm loco}$ specifies robust locomotion tasks such as recovery within two steps after perturbation. (b) A set of neural-network-based constraints is trained with kinematics data to enforce the leg self-collision. (c) STL-MPC encodes the STL specification as a cost function and solves subsequent reduced-order keyframes at $50$ Hz. (d) A whole-body passivity-based controller tracks the desired trajectory at $2000$ Hz. (e) Hardware walking experiments on our bipedal robot Cassie.}
\label{fig:STL_framework}
\vspace{-0.05in}
\end{figure}

\section{Methods}
The core components of the STL-MPC approach are illustrated in \ref{fig:STL_framework}. We formulate locomotion recovery as a trajectory optimization (TO) problem guided by STL specifications. To ensure safety during dynamic walking, we incorporate data-driven constraints for leg-collision avoidance. The TO problem is solved online in a model predictive control (MPC) fashion, continuously updating the foot and center-of-mass (CoM) trajectories, which are tracked by a whole-body passivity-based controller.

%The specification design is summarized in \ref{fig:specification}.

\subsection{Reduced-Order Modeling}
\label{sec:augmented_LIPM}
\begin{figure}[t]
\centering
\includegraphics[width=0.57\textwidth]{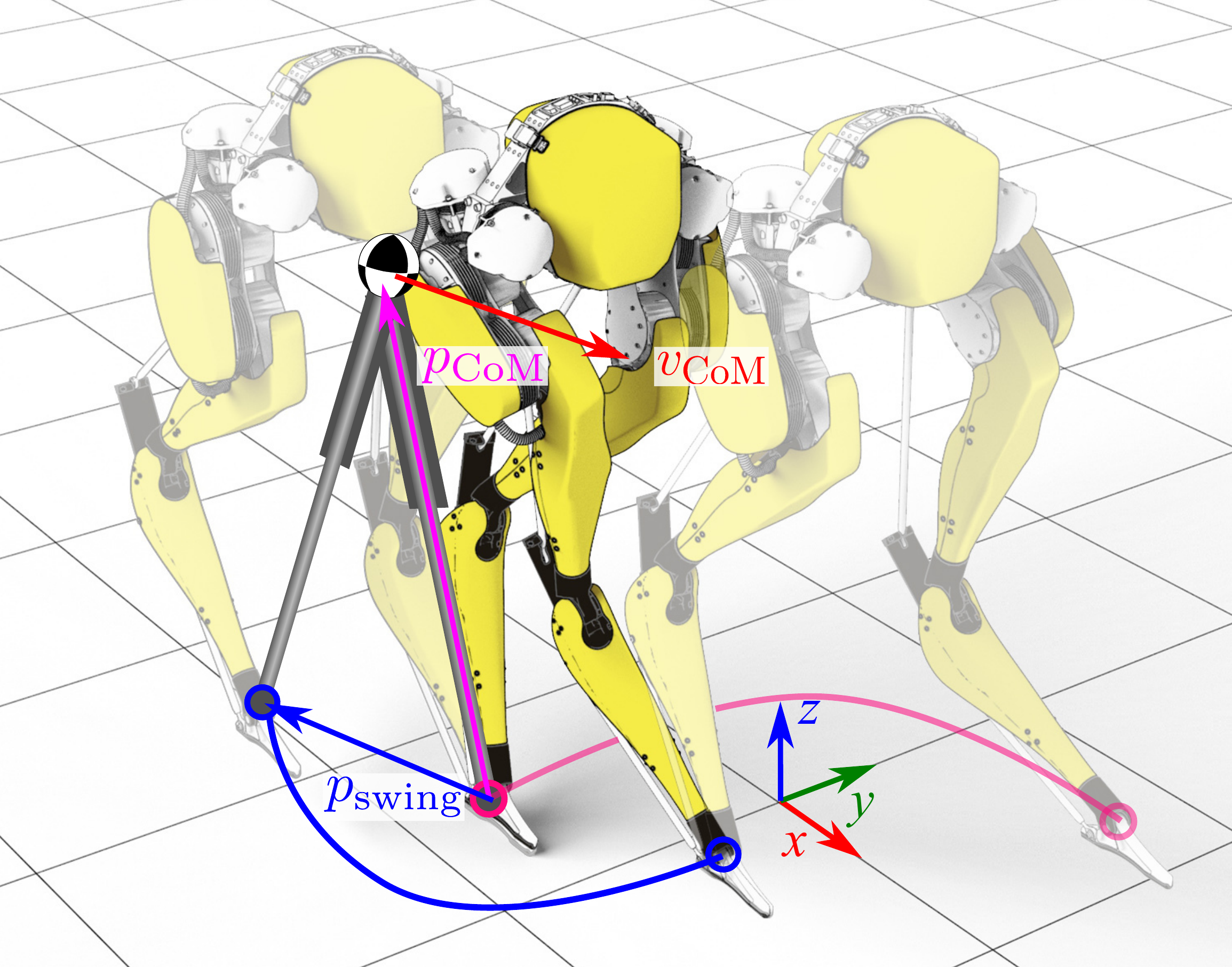}
\caption{Our reduced-order modeling of the Cassie robot as a 3D inverted pendulum model with all of its mass concentrated on its CoM. The massless telescopic leg maintains a constant CoM height. The swing foot position is included in the augmented model to handle leg self-collision avoidance.}
\label{fig:lipm}
\vspace{-0.1in}
\end{figure}

We design a variant of the Linear Inverted Pendulum Model (LIPM)~\cite{Kajita2001} that balances computational efficiency with the ability to enforce kinematic safety constraints, e.g., self-collision avoidance. To achieve this, we augment LIPM with additional swing-foot position and velocity to capture the swing leg behavior, as shown in \ref{fig:lipm}. The augmented state vector is $\boldsymbol{\bar{x}} \coloneqq [\boldsymbol{p}_{\rm CoM}; \boldsymbol{v}_{\rm CoM}; \boldsymbol{p}_{\rm swing}]$, $\boldsymbol{p}_{\rm swing} \in \mathbb{R}^{3}$. We then define the swing foot velocity $\boldsymbol{\dot{p}}_{\rm swing}$ as the control input $\boldsymbol{\bar{u}} = \boldsymbol{\dot{p}}_{\rm swing}$. At contact time, the augmented \textit{reset map} $\boldsymbol{\bar{x}}^+ = \bar{\Delta}_{j \rightarrow j+1} (\boldsymbol{\bar{x}}^{-})$ uses the swing foot location to reset the system states and transition to the next walking step: 
\begin{equation}
\left[ \begin{matrix}
      \boldsymbol{p}_{\rm CoM}^+\\
      \boldsymbol{v}_{\rm CoM}^+\\
      \boldsymbol{p}_{\rm swing}^+
\end{matrix} \right] 
= 
\left[ \begin{matrix}
      \boldsymbol{p}_{\rm CoM}^- - \boldsymbol{p}_{\rm swing}^-\\
      \boldsymbol{v}_{\rm CoM}^- \\
      - \boldsymbol{p}_{\rm swing}^-
\end{matrix} \right].
\end{equation}
The locomotion model switches to the next step when the augmented state reaches the switching condition $\mathcal{S} \coloneqq \{\bar{\boldsymbol{x}} \mid {p}_{{\rm swing},z} = h_{\rm terrain} \}$, where $h_{\rm terrain}$ is the terrain height. At this contact event, the swing foot becomes the new stance foot, and the reset map above updates the state in the new local stance-foot frame.

The augmented LIPM and reset map describe the continuous CoM and swing-foot evolution across contact switches. In order to systematically evaluate the stability of continuous multi-step walking, we use a keyframe to describe each walking step. Following the keyframe representation in~\cite{Zhao2017IJRR,zhao2016robust}, a locomotion keyframe is the robot's CoM phase-space state $\boldsymbol{x} = [\boldsymbol{p}_{\rm CoM}; \boldsymbol{v}_{\rm CoM}]$ at the apex, i.e., when the sagittal projection of the CoM coincides with the stance foot $p_{\mathrm{CoM},x}=0$ in the local stance-foot frame. Expressing this apex state in the local frame makes the representation invariant to the robot's global position and ties each keyframe to the current support foot. This abstraction is useful for non-periodic locomotion: perturbations and terrain variations can change step timing and foot placement, while the sequence of keyframes still provides a compact set of reduced-order states at which stability can be evaluated. Based on these keyframes, we construct Riemannian safety regions and define a signed safety margin, which is subsequently encoded as an STL specification for stability.

\subsection{Riemannian Space in Phase Space and the Safety Margin based on Safety Region}
To quantify robustness in locomotion, we use the concept of a \emph{Riemannian safety region} and \emph{Riemannian safety margin}. 
The Riemannian space reparametrizes the Euclidean CoM phase space using two manifolds derived from the Linear Inverted Pendulum Model (LIPM): the \emph{phase-space tangent manifold} and the \emph{phase-space cotangent manifold}. The tangent manifold $\{(p,v) \in \mathbb{R}^2 \mid \sigma(p,v) = \sigma_{\text{val}}\}$ represents the phase-space manifold along which the CoM dynamics evolve, while the cotangent manifold $\{(p,v) \in \mathbb{R}^2 \mid \zeta(p,v) = \zeta_{\text{val}}\}$ is orthogonal to it~\cite{Zhao2017IJRR}. Here, $\sigma_{\text{val}} \in \mathbb{R}$ and $\zeta_{\text{val}} \in \mathbb{R}$ are scalar values, and each choice of these values defines one tangent or cotangent manifold. The variables $p$ and $v$ are the CoM position and velocity in the stance foot frame.

\begin{figure}
    \centering
    \includegraphics[width=0.6\linewidth]{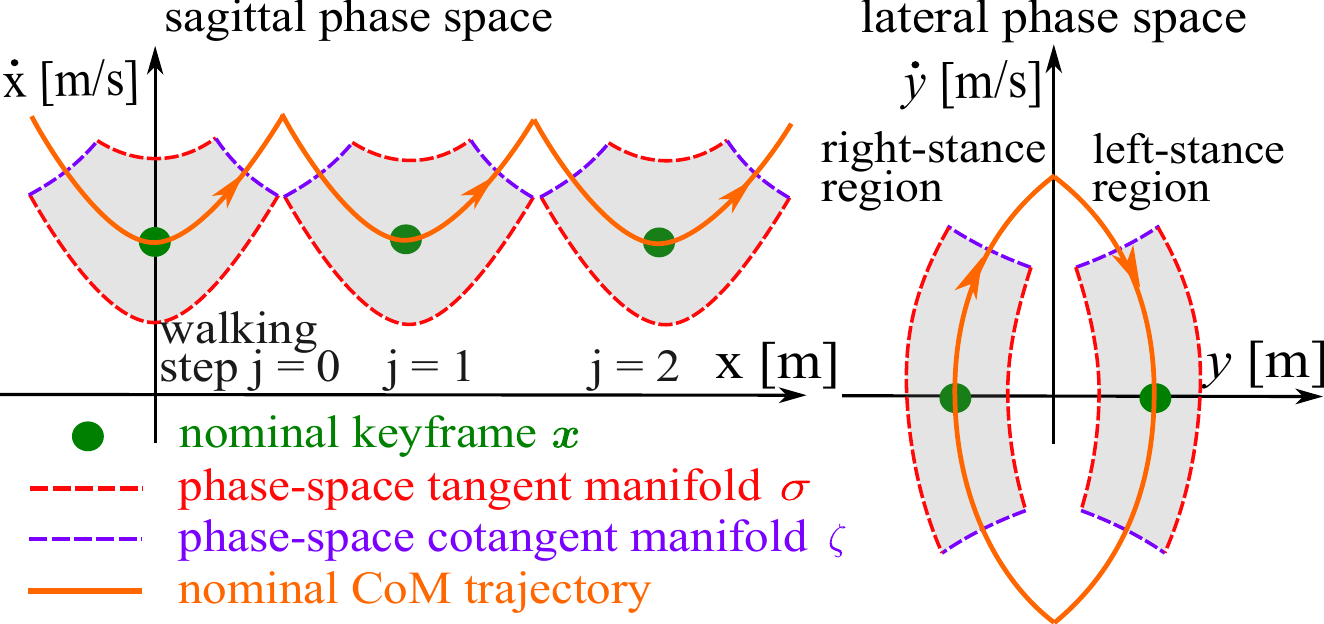}
    \caption{An illustration of a phase-space Riemannian partition and a periodic walking trajectory. The reference frame is the global frame whose origin is located at the apex of the first step.}
    \label{fig:riem_concept}
\end{figure}

\paragraph{Riemannian Safety Region.}
A keyframe is defined as the apex CoM state in a nominal walking manifold, as shown in~\ref{fig:riem_concept}. 
Based on the keyframe definition, the Riemannian safety region $\mathcal{R}$ is defined as a region in phase space around a nominal keyframe state $(\sigma_{\text{nom}}, \zeta_{\text{nom}})$ such that a keyframe inside the region has a positive robustness degree, while one outside has a negative robustness degree. Formally, it is expressed as~\cite{Gu_STL}:
\begin{align}
\mathcal{R}_x &= \{ (p_{\text{CoM},x}, v_{\text{CoM},x}) \mid \sigma(p_{\text{CoM},x}, v_{\text{CoM},x}) \in \Sigma_x, \; \zeta(p_{\text{CoM},x}, v_{\text{CoM},x}) \in Z_x \}, \\
\mathcal{R}_y &= \{ (p_{\text{CoM},y}, v_{\text{CoM},y}) \mid \sigma(p_{\text{CoM},y}, v_{\text{CoM},y}) \in \Sigma_y, \; \zeta(p_{\text{CoM},y}, v_{\text{CoM},y}) \in Z_y \},
\end{align}
where $\Sigma = [\sigma_{\text{nom}} - \delta \sigma, \, \sigma_{\text{nom}} + \delta \sigma]$ and $Z = [\zeta_{\text{nom}} - \delta \zeta, \, \zeta_{\text{nom}} + \delta \zeta]$ define the allowable ranges for the manifold values. The Cartesian product $\mathcal{R} = \mathcal{R}_x \times \mathcal{R}_y$ forms the full Riemannian safety region in both sagittal and lateral phase space.

The bounds of this region are selected from the nominal walking behavior and kinematic limits of Cassie. In the sagittal direction, $\delta \sigma$ is chosen from a desired apex-velocity range of $[0.1, 0.9]$~m/s around the nominal apex velocity of $0.6$~m/s, while in the lateral direction it is chosen from the feasible lateral CoM range allowed by Cassie's leg kinematics. The cotangent margin $\delta \zeta$ specifies the allowable phase window for the apex and is selected to cover the middle portion of a nominal walking step. Thus, these bounds are empirical robot-specific design choices. Compared with Euclidean box bounds around the nominal CoM position and velocity, the Riemannian region aligns its boundary with the LIPM phase-space dynamics: deviations along the tangent manifold correspond to deviations in orbital energy, whereas deviations along the cotangent manifold correspond to deviations in step phase. This parameterization provides a stability margin that is more directly tied to the robot's walking dynamics.

\paragraph{Riemannian Safety Margin.}
The Riemannian safety margin $\rho_{\text{riem}}$ measures locomotion
stability by assessing a keyframe against a Riemannian safety region. Specifically, it is defined as the minimum signed distance of a CoM state $x$ to all the bounds of the Riemannian safety region:
\begin{equation}
\rho_{\text{riem}} := \min_{l=1,\dots,8} r_l(x),
\end{equation}
where $r_l(x)$ denotes the signed distance to the $l$-th bound of a Riemannian safety region. Since both sagittal and lateral regions each contribute four bounds, there are eight in total. A positive $\rho_{\text{riem}}$ indicates the keyframe state resides within the Riemannian safety region, representing a stable walking step, whereas a negative value indicates unstable deviation from the nominal locomotion manifold.

\begin{figure}[t]
\centering
\includegraphics[width=0.99\textwidth]{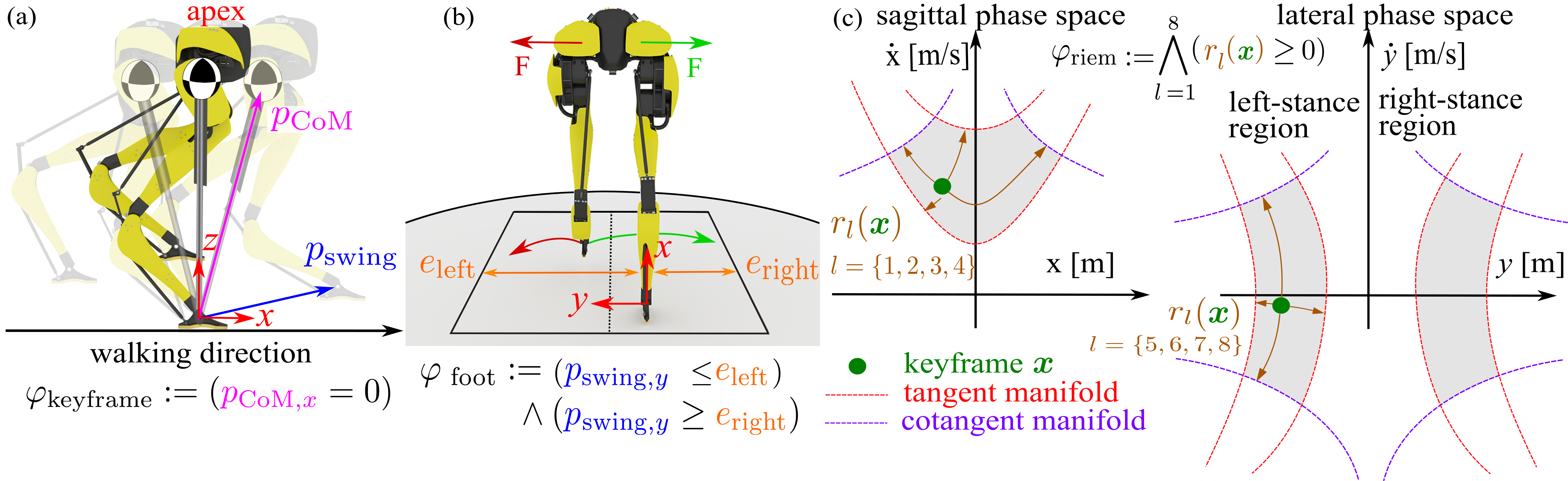}
\caption{Illustration of the locomotion specifications. (a) The highlighted apex state defines the keyframe of a walking step. (b) The swing foot is constrained to step within the lateral bounds of the treadmill. (c) The gray regions denote the predefined Riemannian safety regions in sagittal and lateral phase spaces, and the curved arrows indicate the signed distances to the region bounds. The reference frames in (a) and (b) are local frames fixed to the stance foot and alternate between walking steps.}
\label{fig:specification}
\vspace{-0.15in}
\end{figure}

\subsection{Signal Temporal Logic: Syntax and Robustness Degree}
\label{Sec:STL_detail}
STL \cite{STL_Origin} uses logical symbols of negation ($\neg$), conjunction ($\wedge$), and disjunction ($\vee$), as well as temporal operators such as eventually ($\Diamond$), always ($\square$), and until ($\mathcal{U}$) to construct specifications. A specification formula is defined with the following syntax:
\begin{equation}\begin{split}
\varphi \coloneqq \;
& \pi \;|\; \neg\varphi \;|\; \varphi_1 \wedge \varphi_2 \;|\; \varphi_1 \vee \varphi_2 \;|\;  \Diamond_{[t_1,t_2]}\;\varphi \;|\; \square_{[t_1,t_2]}\;\varphi \;|\; 
\varphi_1 \; \mathcal{U}_{[t_1,t_2]}\; \varphi_2
\end{split}\end{equation}
where $\varphi$, $\varphi_1$, and $\varphi_2$ are STL specifications. $\pi := (\mu^\pi(\boldsymbol{y}) - c \geq 0)$ is a Boolean predicate, where $\mu^\pi: \mathbb{R}^p \to \mathbb{R}$ is a {real}-valued function, $c \in \mathbb{R}$, and the signal $\boldsymbol{y}(t) : \mathbb{R}_+ \to \mathbb{R}^p$ is a $p$-dimensional vector at time $t$. For a dynamical system, a signal $\boldsymbol{y}(t)$ is the system output (in this work, $\boldsymbol{y} = [\boldsymbol{\bar{x}}; \boldsymbol{\bar{u}}] \in \mathbb{R}^{12}$).

The time bounds of an STL formula are represented with $t_1$ and $t_2$, where $0 \leq t_1 \leq t_2 \leq t_{\rm end}$ and $t_{\rm end}$ is the end of a planning horizon. We denote a specific segment of a signal within the interval $[t_1, t_2]$ as $\boldsymbol{y}([t_1:t_2])$. 
The STL semantics $(\boldsymbol{y},t) \models \varphi$ indicates that the signal segment $\boldsymbol{y}([t:t_{\rm end}])$ satisfies $\varphi$. 
The validity of an STL specification is inductively defined using the rules in \ref{tab:STL_satisfy}.

\begin{table}[H]
\centering
\caption{Validity semantics of signal temporal logic.}
\label{tab:STL_satisfy}
\begin{tabular}{l c c}
\hline

% $y =\varphi$ &  $\Leftrightarrow$ & $\rho ^ \varphi((y,0))$   \\ 
$(\boldsymbol{y},t) \models \pi$ & $\Leftrightarrow$ & $\mu ^ \pi  (\boldsymbol{y}(t)) - c \geq 0$ \\

$(\boldsymbol{y},t) \models \neg\varphi$ & $\Leftrightarrow$ & $(\boldsymbol{y},t) \not\models \varphi$ \\

$(\boldsymbol{y},t) \models \varphi_1 \wedge \varphi_2$ & $\Leftrightarrow$ & $(\boldsymbol{y},t) \models \varphi_1 \wedge (\boldsymbol{y},t) \models \varphi_2$ \\

$(\boldsymbol{y},t) \models \varphi_1 \vee \varphi_2$ & $\Leftrightarrow$ & $(\boldsymbol{y},t) \models \varphi_1 \vee (\boldsymbol{y},t) \models \varphi_2$ \\

$(\boldsymbol{y},t) \models \Diamond_{[t_1,t_2]}\varphi$ & $\Leftrightarrow$ & $\exists {t^{'}\in[t+t_1,t+t_2]}, (\boldsymbol{y},t^{'}) \models \varphi$ \\

$(\boldsymbol{y},t) \models \square_{[t_1,t_2]}\varphi$ & $\Leftrightarrow$ & $\forall {t^{'}\in[t+t_1,t+t_2]}, (\boldsymbol{y},t^{'}) \models \varphi$\\

$(\boldsymbol{y},t) \models {\varphi_1}\mathcal{U}_{[t_1,t_2]}{\varphi_2}$ & $\Leftrightarrow$ & \makecell{$\exists {t^{'}\in[t+t_1,t+t_2]}, (\boldsymbol{y},t^{'}) \models \varphi_2 \wedge $\\$ \forall {t^{''}\in[t+t_1,t^{'}]} (\boldsymbol{y},t^{''}) \models \varphi_1$} \\ 
 
\hline
\end{tabular}
\end{table}
\vspace{-0.15in}

In addition to the symbolic syntax, STL admits \textit{quantitative semantics}, which denotes the \textit{robustness degree} \cite{MTL_Papas}, i.e., the extent to which a formula is satisfied by a signal \cite{Belta_STL_review}. 
A positive robustness value indicates satisfaction, and the magnitude represents the degree of task satisfaction. 
For example, in a dynamic environment involving terrain perturbations, the robustness degree indicates the locomotion maneuverability (i.e., the degree of adaptability) to react without violating robot task specifications \cite{fly-by-logic}. When incorporating the robustness degree into TO, it also helps to generate a minimally specification-violating trajectory if the task specification cannot be satisfied strictly \cite{Robust_STL_MPC}. \ref{tab:robustness} shows the semantics of the robustness degree of STL.
\begin{table}[H]
\centering
\caption{Robustness degree semantics of signal temporal logic.}
\label{tab:robustness}
\begin{tabular}{l c c}
\hline

$\rho ^ \pi (\boldsymbol{y},t)$ & $=$ & $\mu ^ \pi  (\boldsymbol{y}(t)) - c$ \\

$\rho ^ {\neg\varphi} (\boldsymbol{y},t)$ & $=$ & $-\rho ^ \varphi  (\boldsymbol{y},t)$ \\

$\rho ^ {\varphi_1 \wedge \varphi_2} (\boldsymbol{y},t)$ & $=$ & ${\rm min} (\rho ^ {\varphi_1} (\boldsymbol{y},t),\rho ^ {\varphi_2} (\boldsymbol{y},t))$ \\

$\rho ^ {\varphi_1 \vee \varphi_2} (\boldsymbol{y},t)$ & $=$ & ${\rm max} (\rho ^ {\varphi_1} (\boldsymbol{y},t),\rho ^ {\varphi_2} (\boldsymbol{y},t))$ \\

$\rho ^ {\Diamond_{[t_1,t_2]}\varphi}(\boldsymbol{y},t)$ & $=$ & ${\rm max}_{t^{'}\in[t+t_1,t+t_2]} (\rho^{\varphi} (\boldsymbol{y},t^{'}))$ \\

$\rho ^ {\square_{[t_1,t_2]}\varphi}(\boldsymbol{y},t)$ & $=$ & ${\rm min}_{t^{'}\in[t+t_1,t+t_2]} (\rho^{\varphi} (\boldsymbol{y},t^{'}))$\\

$\rho ^ {{\varphi_1}\mathcal{U}_{[t_1,t_2]}{\varphi_2}} (\boldsymbol{y},t)$ & $=$ & \makecell{${\rm max}_{t^{'}\in[t+t_1,t+t_2]}({\rm min}(\rho^{\varphi_2}(\boldsymbol{y},t^{'}),$\\${\rm min}_{t^{''}\in[t+t_1,t^{'}]}(\rho^{\varphi_1}(\boldsymbol{y},t^{''}))))$} \\ 
 
\hline
\end{tabular}
\end{table}
\vspace{-0.15in}

\paragraph{Integrating locomotion robustness using STL.}
As illustrated in \ref{fig:specification}, the requirement that ``a CoM keyframe remains inside a Riemannian safety region'' can be encoded as a Signal Temporal Logic (STL) specification:
\begin{equation}
\varphi_{\text{riem}} = \bigwedge_{l=1}^8 \left( r_l(y) \geq 0 \right).
\end{equation}
This encoding bridges formal guarantees and quantitative robustness: the robustness degree $\rho^{\varphi_{\text{riem}}}$ associated with this specification equals the Riemannian safety margin $\rho_{\text{riem}}$. Incorporating this robustness degree into a trajectory optimization and maximizing the safety margin ensures locomotion stability against external perturbations.

\subsection{STL-MPC Formulation}

We formulate STL-guided push-recovery trajectory optimization as a nonlinear program:
\begin{align}
\min_{\boldsymbol{{X}}, \boldsymbol{{U}}, \boldsymbol{T}} \;\; & w \mathcal{L}(\boldsymbol{{U}}) - \Tilde{\rho}^{\varphi_{\rm loco}}(\boldsymbol{{X}}, \boldsymbol{{U}}) \label{eq:cost} \\
\textrm{s.t.} \quad & \boldsymbol{\bar{x}}_{i+1}^j = f(\boldsymbol{\bar{x}}_{i}^j, \boldsymbol{\bar{u}}_{i}^j, T^j), \qquad i \in \mathbb{H} \setminus \mathbb{S}, & j \in \mathbb{J}  \label{eq:continuous_dynamics}\\%A \boldsymbol{\bar{x}}_{i} + B \boldsymbol{\bar{u}}_i \\ 
& \boldsymbol{\bar{x}}^{+,j+1} = \bar{\Delta}_{j \rightarrow j+1} (\boldsymbol{\bar{x}}^{-,j}), & j \in \mathbb{J} \label{eq:reset_map}\\
%
% & \boldsymbol{y} = [\boldsymbol{\bar{x}}; \boldsymbol{\bar{u}}] \\
% & \Tilde{\rho}^{\varphi_{\rm loco}}(\boldsymbol{{X}}, \boldsymbol{{U}}) \ge 0 & \label{eq:robustness}\\
%
& g_{\rm collision}(\boldsymbol{\bar{x}}_i) \ge \epsilon, & i \in \mathbb{H} \label{eq:collision}\\ 
& g_{\rm duration}(T^j) \ge 0, & j \in \mathbb{J} \label{eq:duration}\\ 
& h_{\rm initial}(\boldsymbol{\bar{x}}_0) = 0, \label{eq:initial}\\ 
& h_{\rm transition}(\boldsymbol{\bar{x}}_i) = 0, & i \in \mathbb{S} \label{eq:transition}\\
& \boldsymbol{{X}} \in \mathcal{X}, \boldsymbol{{U}} \in \mathcal{U}, \label{eq:state_limit} 
\end{align}
% $$\mathcal{L}(\boldsymbol{\bar{x}},\boldsymbol{\bar{u}},\boldsymbol{T}) = ||\boldsymbol{T}^{des} - \boldsymbol{T}||^2 + \sum_{i}^{H}(||\boldsymbol{\bar{x}}^{des}_i - \boldsymbol{\bar{x}}_i||^2 + ||\boldsymbol{\bar{u}}^{des}_i - \boldsymbol{\bar{u}}_i||^2)$$ 
where $\mathbb{H}$ is a set of indices that includes all time steps in the horizon. We design $\mathbb{H}$ to span from the acquisition of the latest measured states to the end of the next $N$ walking steps, with a total of $M$ time steps. %\ref{fig:horizon} gives an example of a planning horizon with $N=2$.
$\mathbb{S}$ is the set of indices containing the time steps of all contact switch events, $\mathbb{S} \subset \mathbb{H}$. $\mathbb{J} = \{0, \ldots, N\}$ is the set of walking step indices. The decision variables include $\boldsymbol{{X}} = \{\boldsymbol{\bar{x}}_1, \ldots, \boldsymbol{\bar{x}}_M\}$, $\boldsymbol{{U}} = \{\boldsymbol{\bar{u}}_1, \ldots, \boldsymbol{\bar{u}}_M \}$, and $\boldsymbol{T} = \{T^0, \ldots, T^N\}$. $\boldsymbol{T}$ is a vector defining the individual step durations for all walking steps.

For the objective in \ref{eq:cost}, $\mathcal{L}(\boldsymbol{{U}}) = \sum_{i=1}^M ||\boldsymbol{\bar{u}}_i||^2$ is a cost function penalizing the control with a weight coefficient $w$. The robustness degree $\Tilde{\rho}^{\varphi_{\rm loco}}(\boldsymbol{{X}},\boldsymbol{{U}})$ represents the degree of satisfaction of the signal $(\boldsymbol{X},\boldsymbol{U})$ with respect to the locomotion specification $\varphi_{\rm loco}$.
$\Tilde{\rho}^{\varphi_{\rm loco}}$ is a smooth approximation of ${\rho}^{\varphi_{\rm loco}}$ using smooth operators \cite{Lin_Smooth}. Smooth operators are used because the exact, non-smooth version ${\rho}^{\varphi_{\rm loco}}$ has discontinuous gradients, which can cause the optimization problem to be ill-conditioned. 
Maximizing $\Tilde{\rho}^{\varphi_{\rm loco}}(\boldsymbol{{X}},\boldsymbol{{U}})$ encourages the keyframe towards the center of the Riemannian region. The selection of $w$ is a tradeoff between satisfying the STL specification and minimizing control effort. Specifically, a smaller $w$ allows more aggressive control for rapid recovery from disturbance. In contrast, a larger $w$ penalizes more on control effort but less on the CoM deviation.

To satisfy the augmented linear inverted pendulum model (LIPM) dynamics (\ref{sec:augmented_LIPM}) while adapting step durations $\boldsymbol{T}$, we use a second-order Taylor expansion to derive the approximated discrete dynamics (\ref{eq:continuous_dynamics}). %We use the numerical integration rather than the analytical solution of LIPM because the numerical integration between time steps allows us to enforce collision constraints on any continuous, non-contact-switching states.
\ref{eq:reset_map} defines the augmented reset map from the foot-ground contact switch. 
\ref{eq:collision} represents a set of self-collision avoidance constraints, which ensures a collision-free swing-foot trajectory. The threshold $\epsilon$ is the minimum allowable distance for collision avoidance. The $g_{\rm collision}$ is a set of pre-trained Multilayer Perceptrons (MLPs) representing the minimum distance between leg geometries. 
\ref{eq:duration} clamps step durations $\boldsymbol{T}$ within a feasible range. By allowing variations in step durations, we enhance the perturbation recovery capability of the bipedal system \cite{Ludovic_time_adapt}. \ref{eq:initial} denotes the initial state constraint. $h_{\rm transition}$ in \ref{eq:transition} is a guard function posing kinematic constraints between the swing foot height and the terrain height, ${p}_{{\rm swing},z} = h_{\rm terrain}$, for walking step transitions at contact-switching indices in $\mathbb{S}$. \ref{eq:state_limit} defines the state and control limits. %Notably, the CoM is constrained to a constant height for consistency with the LIPM dynamics, and the swing foot velocity $\boldsymbol{\bar{u}} = \boldsymbol{\dot{p}}_{\rm swing}$ is constrained within a bounding box.

% \subsection{Data-Driven Self-Collision Constraints}
% To calculate the collision status using only the reduced-order model, we adopt MLPs that learn the mapping from Cassie's LIPM state $[\boldsymbol{p}_{\rm CoM}; \boldsymbol{p}_{\rm swing}]$ to the shortest collision distances between leg geometry pairs that are under high risk of collision. These MLP constraints are particularly useful for planning crossed-leg motions. We encode the MLPs as constraints of the MPC to ensure collision-free trajectories. Multilayer perceptrons (MLPs) are trained to approximate inter-leg distances. The MLPs achieve an accurate prediction performance, yielding a maximum absolute error of $0.03$ m and an average absolute error of $0.002$ m. The evaluation speed of the MLPs exceeds $1$ MHz, as compared to less than $1$ kHz for a traditional method requiring iterative inverse kinematics. The accuracy and speed of the MLPs facilitate the online execution of our MPC.

\subsection{Real-Time Implementation}
The optimization is solved online in an MPC style at $50$~Hz using sequential quadratic programming, with speedup techniques such as warm-starting and analytical gradients. At each update, the optimizer outputs desired CoM and swing-foot trajectories, along with optimized step durations over the receding horizon. A whole-body passivity-based controller tracks the resulting trajectories at $2$~kHz. 

\section{Results}

To evaluate the STL-MPC framework, we perform extensive experiments both in high-fidelity simulation and on the Cassie bipedal robot in several perturbation environments, including CAREN~\cite{CAREN}, Bump'em~\cite{BumpEm}, and the \textit{Stiletto} vessel. 

\subsection{Simulation}

Simulation experiments are performed under diverse perturbation scenarios and benchmarked against multiple baseline approaches.

\begin{figure}[t]
\centerline{\includegraphics[width=.6\textwidth]{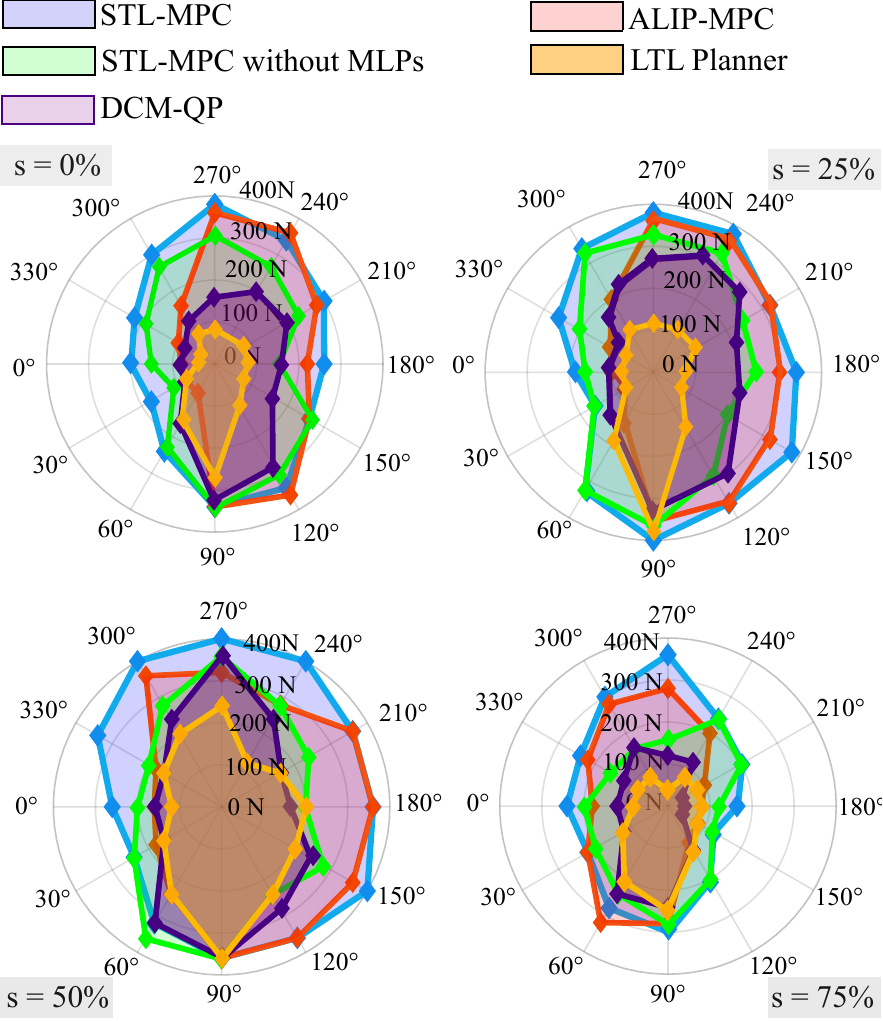}}
\caption{The maximum allowable force exerted on the pelvis from which the robot can safely recover within two steps in all 12 directions. The perturbations happen at different phases $s$ during a left leg stance. Values on the left half result in crossed-leg maneuvers, and values on the right half correspond to wide-step recoveries. 
We benchmark with an Angular-Momentum Linear Inverted Pendulum Model Predictive Control (ALIP-MPC) baseline, an LTL planner, a Divergent Component of Motion Quadratic Program (DCM-QP), and the STL-MPC without MLPs.
}
\label{fig:spider}
\vspace{-0.05in}
\end{figure}

\paragraph{Omnidirectional Perturbation Recovery.}
We evaluate STL-MPC under a comprehensive set of omnidirectional perturbations applied to Cassie’s pelvis. The simulation study systematically varies perturbation magnitudes ($80$--$400$ N), directions ($12$ evenly spaced orientations), and gait phases ($s = 0\%, 25\%, 50\%, 75\%$), yielding 432 distinct trials. Results in \ref{fig:spider} show that STL-MPC reliably recovers from omnidirectional perturbations and significantly outperforms baseline controllers. Notably, STL-MPC tolerates larger forces in lateral directions, where crossed-leg maneuvers are frequently employed. 

In contrast, LTL-based planners exhibit smaller feasible regions, failing whenever the perturbed state falls outside the Riemannian safety region. Similarly, the DCM-QP baseline produces overly conservative responses due to its reliance on periodic reference tracking and single-step duration optimization, which limits recovery from perturbations occurring late in the gait cycle. These results highlight STL-MPC’s superior resilience to omnidirectional disturbances compared to state-of-the-art baselines.

\begin{figure}[t]
\centerline{\includegraphics[width=.69\textwidth]{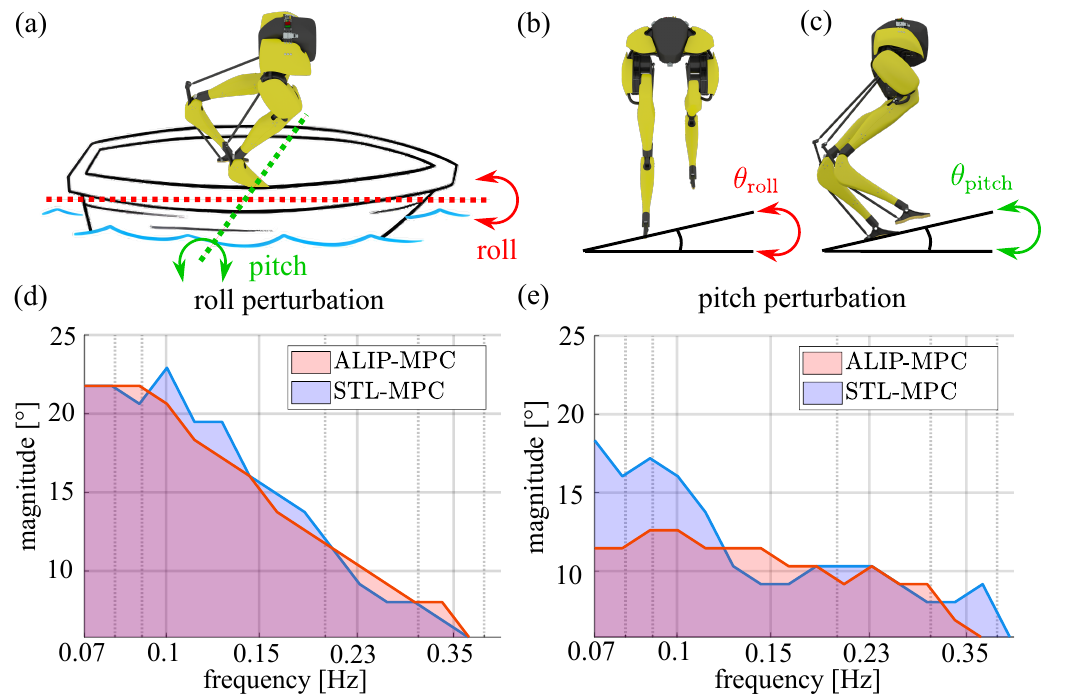}}
\caption{Recovery limits on a rigid surface undergoing sinusoidal ship-like motion. (a)--(c) illustrate the roll and pitch oscillations. (d) and (e) compare the maximum tolerable oscillation amplitudes of STL-MPC and ALIP-MPC across roll and pitch frequencies, respectively. The frequency axis is logarithmic, and failure cases are omitted.}
\label{fig:simubomo}
\vspace{-0.05in}
\end{figure}

\paragraph{Orientational Perturbations from Dynamic Rigid Surfaces.}
We simulate ship-like surfaces undergoing sinusoidal roll and pitch oscillations. For each axis, the oscillation amplitude ranges from $6^\circ$ to $30^\circ$, and the frequency ranges logarithmically from $0.07$ to $0.7$~Hz. The recovery limits in \ref{fig:simubomo} show that STL-MPC tolerates larger pitch oscillations than ALIP-MPC because it allows more CoM height variation and maintains firm ground contact. The two methods perform similarly under roll oscillations, where large surface inclinations primarily cause slipping.

\begin{figure}[t]
\centerline{\includegraphics[width=.75\textwidth]{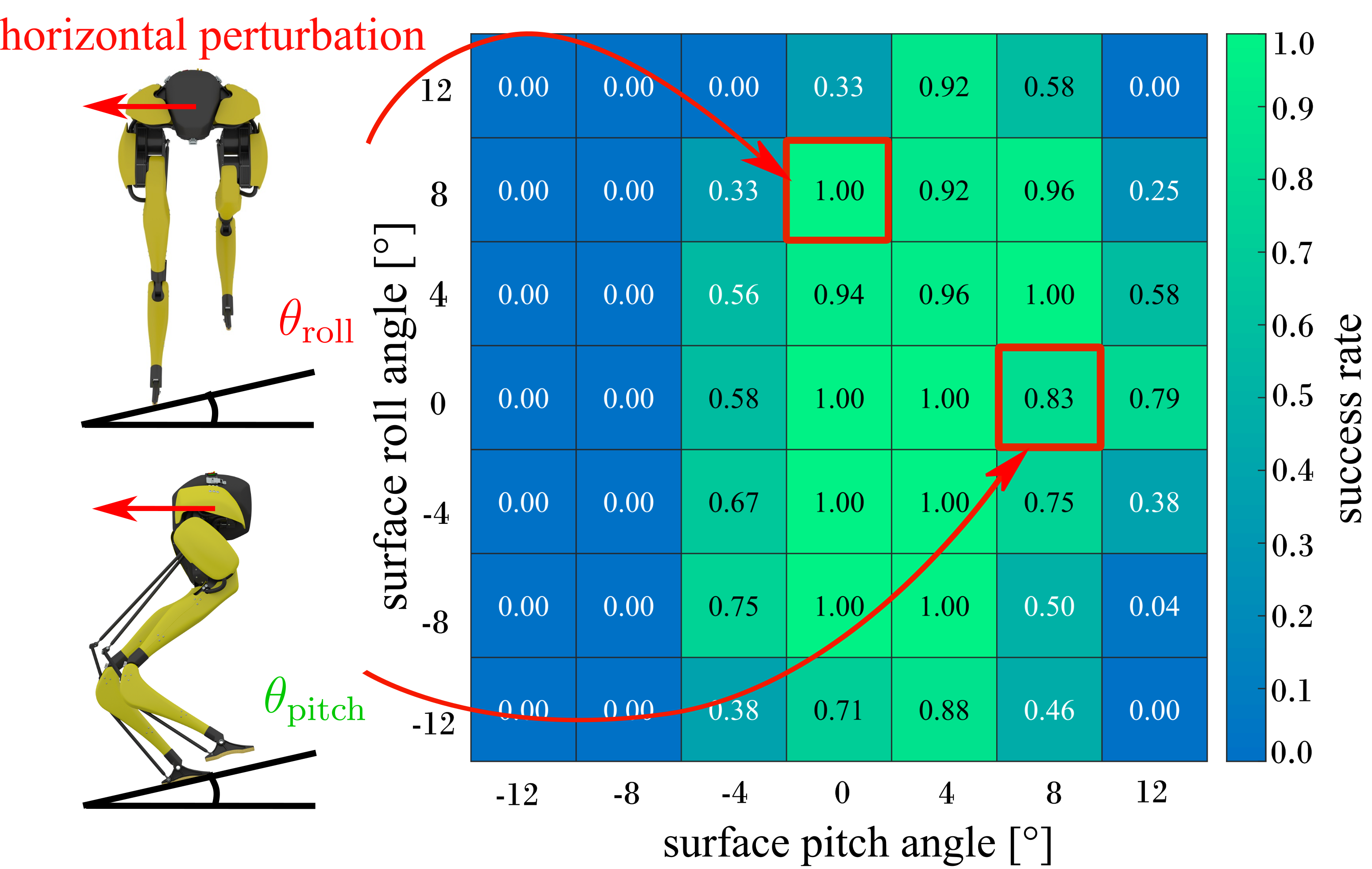}}
\caption{Push-recovery success rates across stationary surface inclinations. At each inclination, a $100$~N horizontal perturbation is applied from $12$ directions at locomotion phase $s=0\%$.}
\label{fig:cnstIncline}
\vspace{-0.05in}
\end{figure}

\paragraph{Horizontal Perturbations on Inclined Stationary Surfaces.}
We further evaluate push recovery on $49$ stationary inclinations formed by varying the platform pitch and roll from $-12^\circ$ to $12^\circ$ in $4^\circ$ increments. As shown in \ref{fig:cnstIncline}, the success rate generally decreases with inclination, with better recovery uphill than downhill. STL-MPC also remains robust across roll inclinations by using collision-free crossed-leg maneuvers for lateral recovery.

\begin{figure}[t]
    \centering
    \includegraphics[width=0.78\textwidth]{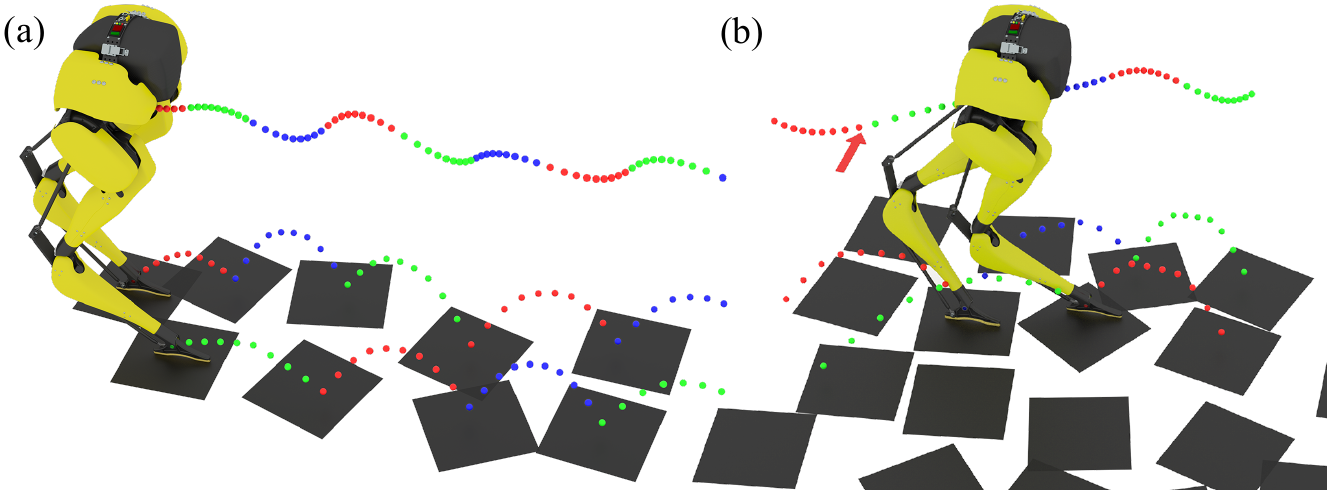}
\caption{Illustration of maneuvering over two stepping-stone scenarios. (a) STL-MPC solves dynamically feasible trajectories that satisfy an additional foot-on-stones specification. (b) STL-MPC successfully plans crossed-leg maneuvers to recover from a perturbation.}
    \label{fig:stepping_stone}
    \vspace{-0.05in}
\end{figure}
\paragraph{Stepping Stone Maneuvering.}
To further demonstrate STL-MPC’s flexibility in handling diverse task specifications, we evaluate its performance in stepping-stone locomotion scenarios. Unlike traditional approaches that require rule-based heuristics or mixed-integer formulations, STL encodes stepping-stone foothold constraints directly as logical specifications. Specifically, an additional constraint $\varphi_{\text{stones}}$ is introduced, requiring each foot placement to remain within the boundaries of convex stepping stones. The resulting compound specification $\varphi'_{\text{loco}} = \varphi_{\text{loco}} \wedge \varphi_{\text{stones}}$ enforces both locomotion stability and precise foot placement. 

Simulation trials in~\ref{fig:stepping_stone} demonstrate that STL-MPC successfully generates dynamically feasible trajectories across randomly placed and oriented stepping stones. Moreover, under lateral perturbations, STL-MPC plans stable crossed-leg maneuvers to recover balance while still respecting the stepping-stone constraints. This highlights our STL-MPC’s ability to simultaneously satisfy complex task-level requirements, such as terrain-constrained foot placement, while ensuring physical feasibility and robustness to disturbances. 

\begin{figure*}[t]
\centerline{\includegraphics[width=0.95\textwidth]{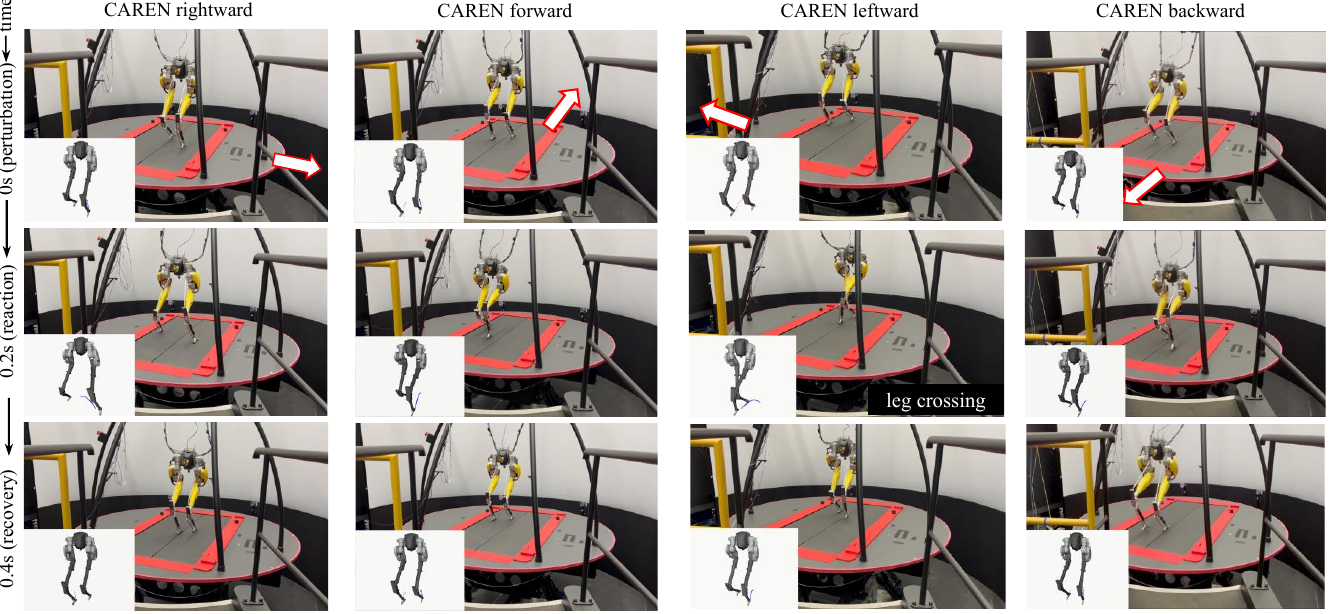}}
\caption{We test our STL-MPC framework on the CAREN system. The Stewart platform provides perturbations at a controlled timing with a specified direction and magnitude. The Cassie robot recovers from perturbations in all four directions using diverse recovery strategies. For example, the leftward CAREN perturbation (third column) induces a crossed-leg maneuver.}
\label{fig:CAREN}
\vspace{-0.05in}
\end{figure*}

\subsection{Hardware Experiments}
To validate the simulation results, we conduct hardware experiments on the Cassie robot in three settings: the CAREN and BumpEm perturbation platforms, which provide terrain and pelvis disturbances, respectively, and the \textit{Stiletto} vessel, which subjects the robot to real-world ocean-wave oscillations. Together, these experiments evaluate STL-MPC under omnidirectional, large-magnitude, and orientational perturbations in both controlled laboratory settings and an operational shipboard environment.

\begin{figure*}[t]
\centerline{\includegraphics[width=.95\textwidth]{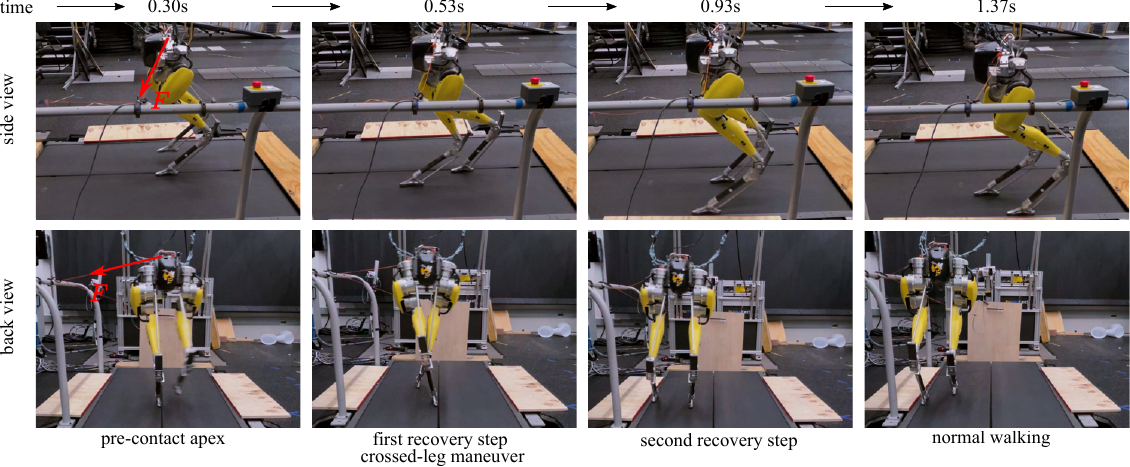}}
\caption{Cassie recovers from a CoM perturbation applied by the BumpEm pulling tower. The perturbation is of $130$ N magnitude at $0^\circ$ direction. The STL-MPC planner reacts with a crossed-leg maneuver, followed by a wide step, and finally recovers to stable normal walking.}
\label{fig:bumpem_recovery}
\vspace{-0.05in}
\end{figure*}

\paragraph{Omnidirectional Horizontal Perturbation on CAREN.}
We first test STL-MPC in the Computer-Aided Rehabilitation Environment (CAREN) \cite{CAREN}, which mounts a treadmill on a $6$-DoF Stewart platform capable of generating controlled translations and rotations. Perturbations of varying magnitudes ($10$--$20$ cm translations), directions ($12$ evenly spaced), and gait phases ($s = 0\%, 25\%, 50\%, 75\%$) are applied. 
STL-MPC consistently recovers from all cases, with the exception of one extreme instance ($20$ cm at $180^\circ$ during late stance), where the required recovery maneuver exceeds the treadmill’s spatial limits rather than the framework’s capability. 
Example trials in~\ref{fig:CAREN} show diverse recovery behaviors, including crossed-leg maneuvers for lateral pushes. 

\begin{figure}
\centerline{\includegraphics[width=.75\textwidth]{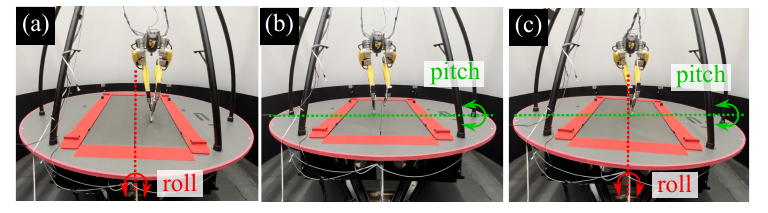}}
\caption{(a) The roll oscillation up to $5^\circ$. (b) The pitch oscillation to $5^\circ$. (c) Oscillation along both the roll and pitch axes up to $4^\circ$. }
\label{fig:caren_bomo}
\vspace{-0.05in}
\end{figure}

\paragraph{Large Horizontal Perturbation on a BumpEm System.}
To evaluate responses to high-magnitude impulses, we employ the BumpEm system, which delivers cable-pulling forces directly to Cassie’s pelvis. Perturbations of increasing magnitude are applied from eight directions until failure occurs. Results in~\ref{fig:bumpem_recovery} show that STL-MPC withstands forces comparable to simulation predictions, including crossed-leg recoveries for lateral pulls up to $130$ N. These experiments validate STL-MPC’s capacity to execute aggressive recovery strategies against large impulses beyond the limits of CAREN.

\paragraph{Orientational Perturbation on CAREN.}
We further challenge the framework with terrain orientation perturbations on the CAREN platform by oscillating the treadmill in pitch and roll axes up to $5^\circ$ (\ref{fig:caren_bomo}). STL-MPC maintains stable locomotion across all tested oscillations, adapting CoM trajectories to preserve ground contact and avoid slips. The hardware experiments reveal that the robot frequently
adopts crossed-leg recovery strategies to resist roll oscillations, underscoring the significance of this maneuver
for enhancing the overall stability. These results highlight the terrain-agnostic generalizability of our STL-MPC in handling rotational terrain disturbances.

\begin{figure}
\centerline{\includegraphics[width=.75\textwidth]{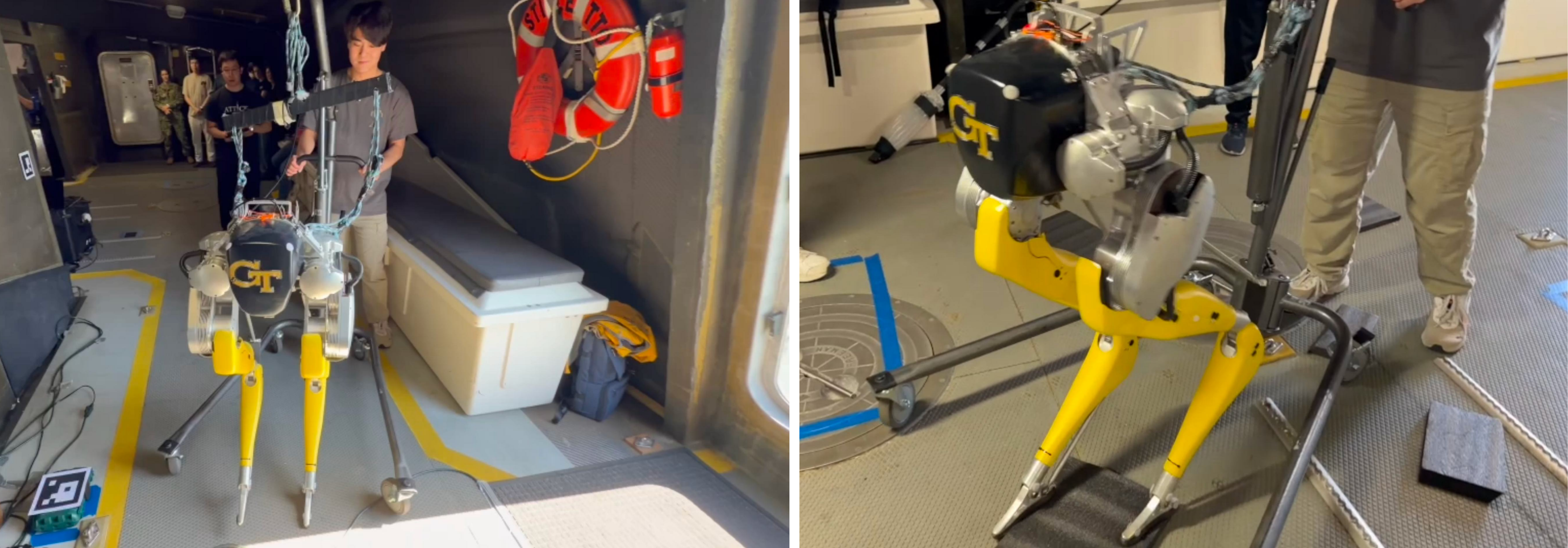}}
\caption{Cassie stabilizes under external pushes on the uneven deck of the Stiletto vessel.}
\label{fig:stiletto}
\vspace{-0.05in}
\end{figure}

\paragraph{Field Test on the \textit{Stiletto} Vessel.}
As a validation beyond laboratory perturbation platforms, we conducted a naval-research-sponsored field test aboard the \textit{Stiletto} vessel in Virginia Beach, VA. Cassie was deployed on the ship deck while the vessel traveled at speeds up to $20$ knots and executed an S-shaped maneuver. Compared with CAREN's orientational perturbations, the ship-induced perturbations had lower oscillation frequencies due to the vessel's large inertia, but larger perturbation magnitudes than CAREN could safely exert. As shown in ~\ref{fig:stiletto}, we further increased the task difficulty by applying external pushes to the robot and placing uneven blocks on the deck. Cassie achieved stable locomotion across these conditions, indicating that our STL-MPC can transfer from controlled laboratory perturbations to realistic field environments.

% \paragraph{Summary.}
% Across all three hardware setups, STL-MPC demonstrated robust push recovery, executing flexible strategies such as crossed-leg and wide-step maneuvers, while maintaining formal robustness margins. The experiments confirm that the framework’s advantages in simulation translate effectively to hardware, underscoring its potential for real-world deployment in unpredictable environments.

\section{Discussion}

\paragraph{STL-MPC compared to reinforcement learning for robust locomotion.}

Reinforcement learning (RL) can learn complex locomotion and recovery behaviors through trial-and-error in simulation. However, its robustness is typically established empirically through reward design, domain randomization, and large-scale training, rather than by providing a formal satisfaction guarantee for a specified recovery requirement. When accurate locomotion models and domain knowledge are available, a model-based approach with formal verification offers stronger assurances, particularly under severe perturbations where RL policies may encounter out-of-distribution scenarios.
Therefore, STL-MPC is most effective when recovery objectives can be explicitly specified, whereas RL is better suited to behaviors involving complex contact dynamics or whole-body coordination that cannot be accurately captured by compact analytical models.

% STL-MPC addresses this need by evaluating each planned trajectory against an explicit temporal-logic specification and computing a robustness degree equal to the Riemannian safety margin. This robustness degree quantifies both whether the planned keyframe sequence satisfies the recovery specification and how far it is from violating the safety region. 

\paragraph{Formal methods in the era of robot learning.}
More broadly, formal methods retain value in the era of AI because their specification and verification tools complement the learning approach at different stages. Before learning, formal synthesis or trajectory optimization can generate dynamically feasible trajectories with explicit guarantees, providing demonstrations or priors for data-driven policies. During learning or execution, formal methods can use specifications as online guidance or constraints, helping the robot remain consistent with dynamics and task requirements. After learning, formal verification can evaluate whether a learned policy satisfies explicit safety or recovery specifications, expose failure modes, and potentially constrain or improve the policy actions. Thus, formal methods need not be viewed as alternatives to learning; they can define what should be satisfied, guide how policies are trained or executed, and assess whether learned behaviors remain reliable beyond empirical test cases.

\section{Conclusion}

This chapter used STL to specify locomotion recovery requirements over continuous motion and to quantify their satisfaction through a Riemannian robustness degree. Incorporating this robustness degree into optimization, we unified stability specification and motion generation within an online MPC. Simulation and hardware experiments demonstrated omnidirectional recovery, constrained stepping, and responses to translational and orientational perturbations. The field test aboard the \textit{Stiletto} vessel demonstrated the framework's robustness and its potential for a real-world operational application.

Together, \ref{chapter:LTL-BT} and \ref{chapter:STL-MPC} show how formal specifications and dynamics models can quantify and enhance locomotion robustness. Their performance focused on mobility and built the foundation for humanoid autonomy. The next chapter turns to the complementary capability of whole-body manipulation, where human demonstrations, tactile sensing, and force feedback provide the information needed for contact-rich interaction.

%% file: chapters/tactile-humanoid.tex
\chapter{WT-UMI: Tactile-based Whole-Body Manipulation via Force-Supervised Contact-Aware Planning}
\label{chapter:tactile-humanoid}

\noindent\textbf{Dissertation Roadmap and Contribution of This Chapter.}
This chapter presents the second thrust of the dissertation: \emph{force-aware whole-body manipulation through large-area tactile sensing}. This thrust expands the dissertation from robust locomotion to tactile-based whole-body manipulation, with a focus on leveraging human demonstrations. Human data provides coordinated motion and natural contact-force interactions, but it lacks robot-executable action labels. To address this challenge, we design WT-UMI, a tactile interface that collects human tactile demonstrations at scale. WT-UMI is also a learning framework that converts human motion into robot-executable actions through a correction module trained on robot teleoperation data.
The corrected human data preserves the natural contact-force information and enables a planner to predict both target motion and contact force simultaneously. During deployment, the predicted force is regulated through tactile-based admittance control, allowing the humanoid to perform contact-rich whole-body manipulation with bulky, deformable objects.

\begin{figure}[t]
\centerline{\includegraphics[width=0.95\textwidth]{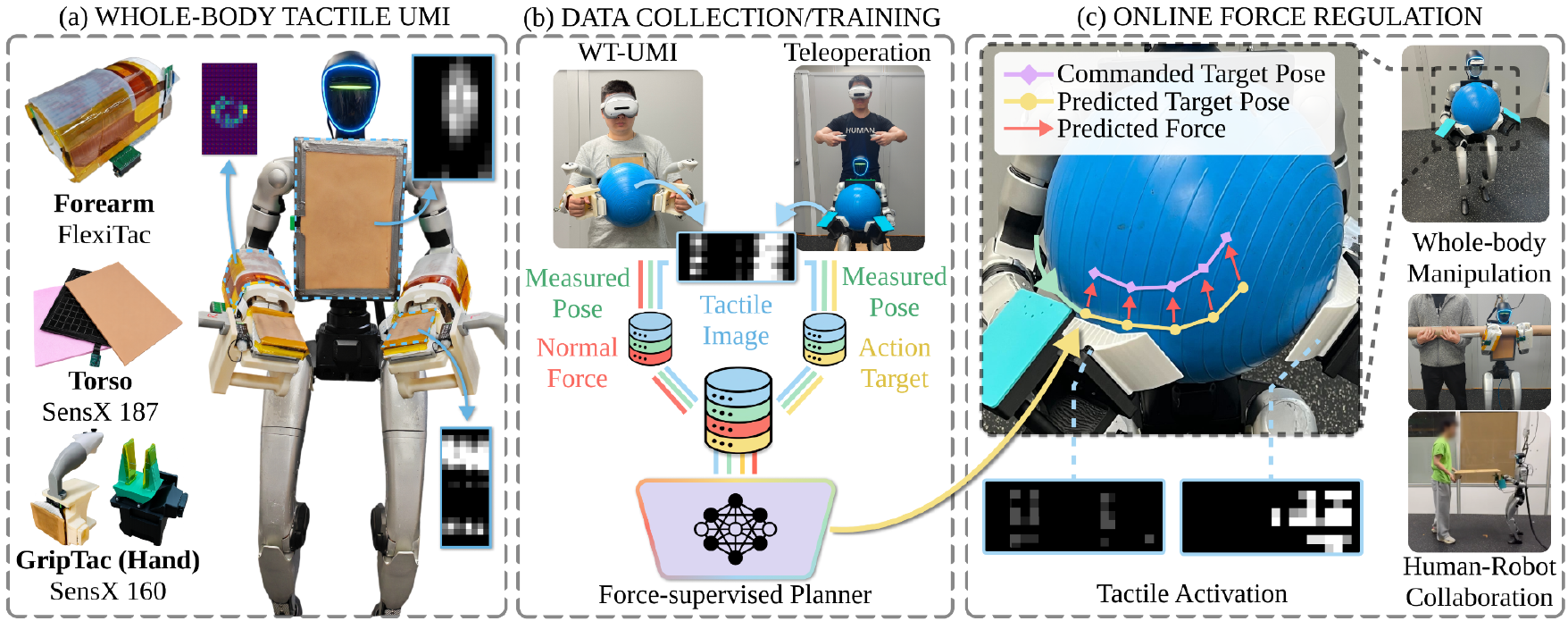}}
\caption{(a) \textbf{WT-UMI} is a shared interface between human demonstrators and humanoid robots for large-area tactile data collection. (b) A human demonstrator wears WT-UMI, or a humanoid is controlled via teleoperation using the same hardware. (c) A force-supervised planner trained from WT-UMI data executes contact-rich tactile-aware tasks, spanning whole-body manipulation of deformable and large rigid objects and human--humanoid collaborative transport.}
\label{fig:wtumi_teaser}
\vspace{-0.15in}
\end{figure}

\section{Introduction}

Human demonstrations naturally encode how people coordinate motion and distributed body contact when manipulating bulky, deformable, or shared-load objects. Transferring only the demonstrated kinematics, however, omits the interaction forces that make these behaviors physically successful. Tasks such as carrying a large box, reorienting a soft pillow, or transporting a beam with a human partner require coordinated contact across the torso, forearms, and hands~\cite{gu_humanoid_2025}; small errors in contact location or force allocation can cause slip, collision, or load loss. Reliable whole-body manipulation therefore requires demonstrations and policies that represent both motion and force, together with tactile feedback that regulates contact during execution.

Most learning-based manipulation systems rely on vision and proprioception~\cite{diffusion_policy}. However, vision is often occluded during contact and does not directly measure interaction forces, while proprioception cannot localize body-surface contact. Prior tactile sensing has focused primarily on fingertip arrays~\cite{niu2026learningversatilehumanoidmanipulation, hou2025adaptive, han2024learning, feats2025, helmut2025tactileconditioneddiffusionpolicyforceaware, zhang2026tacvlacontactawaretactilefusion}, which do not capture the distributed contact central to whole-body interaction. Existing body-mounted tactile systems~\cite{Mittendorfer_tactile2013, Punyo_SR} often rely on model-based controllers or per-task reward shaping, limiting their ability to learn diverse skills. Turning distributed whole-body tactile sensing into a contact-force plan for humanoid whole-body manipulation remains an open problem.

We address this problem with \textbf{WT-UMI}~\cite{jang2026wtumi} (\textbf{W}hole-Body \textbf{T}actile \textbf{U}niversal \textbf{M}anipulation \textbf{I}nterface), a wearable tactile interface paired with a force-aware learning framework for humanoid large-area upper-body manipulation (\ref{fig:wtumi_teaser}). Here, whole-body manipulation refers to distributed contact across multiple body regions; the implemented tactile coverage is limited to the palms, forearms, and chest. WT-UMI enables scalable human demonstration collection and reduces, but does not eliminate, the human--humanoid embodiment mismatch through shared sensing hardware. The human demonstrations collected through WT-UMI capture whole-body, contact-rich interactions with calibrated contact forces, enabling explicit contact-force prediction during training and force regulation during deployment. However, human demonstrations often lack robot-executable action labels. To convert these force-aware human demonstrations into robot-executable actions, we introduce a \textbf{force-conditioned target-pose correction} module that learns target-pose corrections from robot-in-the-loop teleoperation. Human and teleoperation trajectories are paired using force-inferred contact modes, with teleoperation commands supervising the correction of measured human poses. The resulting corrected target poses serve as action labels to supervise planner training. The action-labeled human data preserves accurate contact-force measurements, enabling a \textbf{force-supervised planner} to jointly predict contact forces and corrected actions (\ref{fig:wtumi_framework}). During deployment, a tactile-based admittance controller tracks the predicted force trajectory by modulating the predicted end-effector poses, achieving stable and force-regulated contact.

Overall, our framework offers four key contributions:
\begin{itemize}
\item We introduce \textbf{WT-UMI}, a wearable tactile interface that collects tactile images and force-supervised human demonstrations over large upper-body areas. The same sensing hardware supports both human demonstrations and humanoid teleoperation.
\item We design a \textbf{force-conditioned target-pose correction} module that converts human trajectories into contact-aware robot actions by learning pose corrections from teleoperation data.
\item We develop a \textbf{force-supervised planner} whose cross-attention head predicts the normal contact-force trajectory, with the predicted force serving as the reference for a tactile-based admittance controller.
\item We validate the framework on five large-area contact-rich manipulation tasks spanning deformable objects, bulky rigid objects, and human--humanoid collaboration, where it improves the success rate and reduces contact-position drift over four policy baselines.
\end{itemize}

\section{Related Work}

\subsection{Large-Area Tactile Sensing and Manipulation}
Whole-body tactile sensing is an emerging modality for contact-rich manipulation, yet its adoption remains challenging~\cite{gu_humanoid_2025}. Much prior work restricts contact to the hands, leaving the torso and forearms largely uncovered~\cite{cheng2026tacumimultimodaluniversalmanipulation}. Compliant-body designs such as Punyo~\cite{Punyo_SR} extend coverage with pressure-sensitive skins on the arms and chest, while discrete-cell skins such as HEX-o-SKIN~\cite{Mittendorfer_tactile2013} cover larger areas but require per-cell kinematic calibration. In contrast, our thin-film piezoresistive arrays produce dense 2D contact images with large-area coverage that integrate naturally with vision encoders and imitation-learning pipelines. Existing whole-body manipulation methods span model-based control~\cite{Armleder_Gordon_huamnoid_2025, Murooka_2024}, planning~\cite{Subburaman_plan_estimate_tactile_humanoid_2025}, reinforcement learning~\cite{zheng2025embracingbulkyobjectshumanoid, liu2024opt2skill, chip, HMC, softmimic, wu2025learn, REFINEDP}, and imitation learning~\cite{diffusion_policy, Murooka_TACT}. However, these methods typically rely on accurate contact models, object-state estimation, and task-specific reward design, rather than learning an explicit contact-force plan for whole-body humanoid manipulation. Murooka et al.~\cite{Murooka_2024} and ForceMimic~\cite{ForceMimic_2025} use six-dimensional contact wrenches for balance or fine-grained manipulation. Here, scalar normal-force prediction supports large-area contact regulation, with a simpler sensing representation that does not resolve shear forces or contact moments. This chapter develops that tactile sensing and control approach, which complements the vision-and-proprioception loco-manipulation policies studied later in \ref{chapter:DPFT}.

\subsection{Joint Motion--Force Prediction}
Model-based hybrid motion--force control requires explicit force sensing and careful contact modeling~\cite{Armleder_Gordon_huamnoid_2025, Khatib_IJRR_OSC, wijayarathne2023real}. Learning-based approaches such as ForceMimic and UMI-FT learn motion--force representations from force--motion capture or wrist/finger force sensing, but their measurements remain local~\cite{ForceMimic_2025, umift}. Recent tactile- and force-aware policies further demonstrate the value of contact-aware learning through tactile-conditioned force actions, tactile--force representation learning, force-feedback fusion, or force-token distillation~\cite{helmut2025tactileconditioneddiffusionpolicyforceaware, huang2026tafvlatactileforcealignmentvisionlanguageaction, NEURIPS2025_8633b46e, li2026forcevla2unleashinghybridforceposition, zhao2026fdvlaforcedistilledvisionlanguageactionmodel}. SoftAct~\cite{yoo2026softact} generates force-labeled demonstrations using simulated force sensors, while Contact-Grounded Policy~\cite{xu2026cgp} executes predicted contact states through compliant control for dexterous manipulation. These methods primarily focus on local fingertip, gripper, and wrist contacts for tabletop tasks. Humanoid Touch Dreaming~\cite{niu2026learningversatilehumanoidmanipulation} predicts future tactile signals in a learned latent space, but the predictions act as auxiliary regularization and contact is modeled only implicitly. Our method instead (i) predicts contact force explicitly from distributed tactile arrays spanning the robot body, and (ii) uses the predicted force as a reference for a tactile-based admittance controller, yielding explicit force-aware control during whole-body manipulation.

\subsection{Demonstration Interfaces for Behavior Cloning}
Teleoperation provides direct robot action labels, with Virtual Reality (VR), retargeting, and recent portable systems improving ergonomics and coverage~\cite{zhang2018deep, qin_anyteleop_2023, mobile_aloha, aloha, ze2026twist2}. However, it still requires robot access, task setup, and skilled operators. In contrast, UMI-style interfaces bypass robot-in-the-loop collection by recording in-the-wild human demonstrations with handheld devices and transferring them to robot policies~\cite{chi2024umi}. Adeniji et al.~\cite{adeniji2025feelforce} collect force-labeled human demonstrations with a tactile glove and transfer predicted forces to a tactile gripper. However, these interfaces do not provide distributed contact-force measurements across large surface areas for explicit force planning. When touch is included, human-to-robot transfer becomes more challenging since human and robot embodiments, as well as their tactile sensors, may differ. TactAlign tackles this issue by aligning human and robot tactile observations in a shared latent space using rectified flow, without paired data or manual labels~\cite{wi2026tactalignhumantorobotpolicytransfer}. WT-UMI takes a complementary route: it uses shared wearable/robot-mounted tactile hardware to reduce sensing mismatch, while combining human demonstrations with tactile force measurements and robot teleoperation labels for whole-body humanoid behavior cloning.

\section{Methods}
\label{sec:wtumi_methods}

\begin{figure}[t]
\centerline{\includegraphics[width=0.99\textwidth]{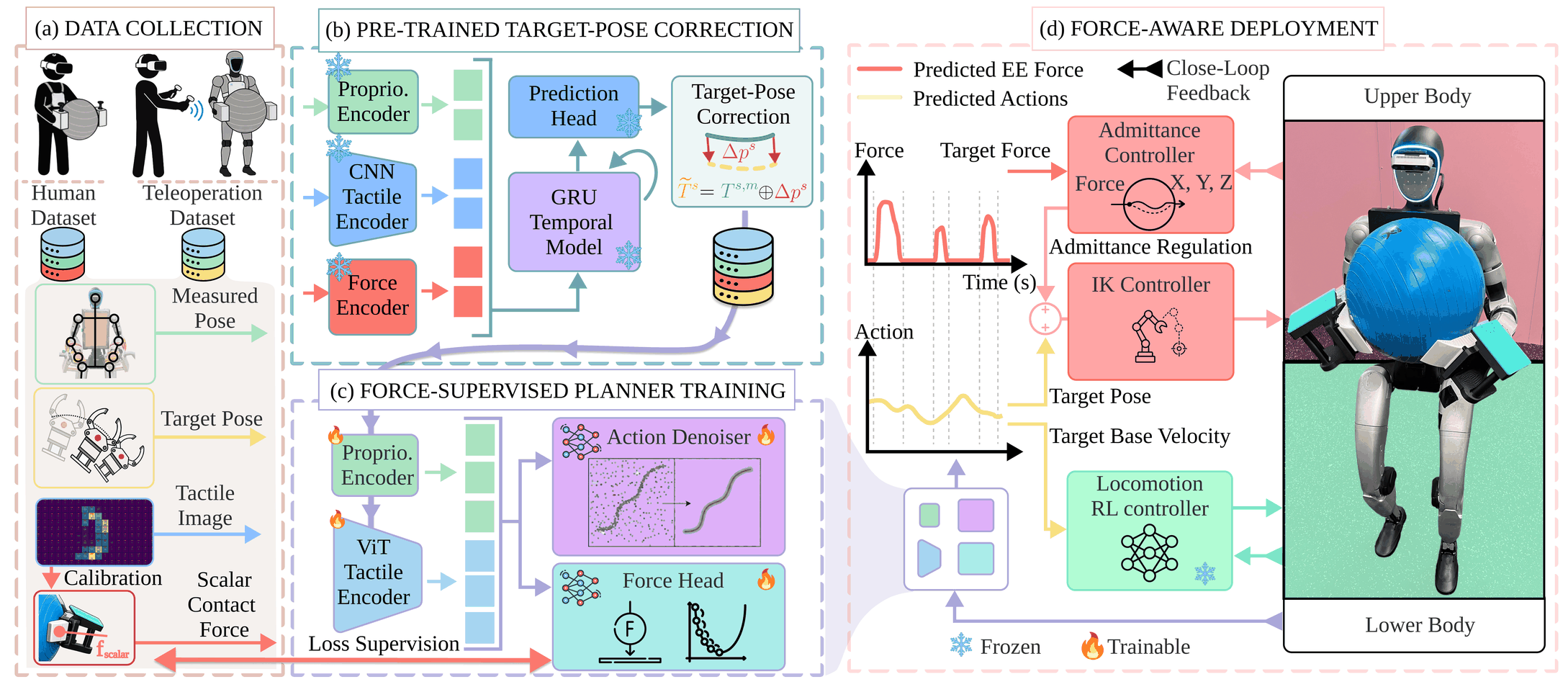}}
\caption{A force-conditioned target-pose correction module creates action labels for human data. A force-supervised planner produces a contact-force trajectory in addition to end-effector poses. The predicted forces are used for online force regulation via a tactile-based admittance controller.}
\label{fig:wtumi_framework}
\vspace{-0.15in}
\end{figure}

Our system architecture is outlined in \ref{fig:wtumi_framework}. In \emph{data collection}, WT-UMI records distributed tactile readings and calibrated contact forces from a human operator or via teleoperation on a humanoid robot. In \emph{pre-trained target-pose correction}, a force-conditioned model learns to correct human target poses into robot-executable actions. In \emph{planner training}, an action denoiser predicts corrected bimanual pose chunks, while a force head predicts contact-force trajectories. During \emph{force-aware deployment}, the predicted forces drive a tactile-based admittance controller that regulates target poses. A pre-trained RL policy simultaneously controls lower-body locomotion.

\subsection{Tactile Sensing and Force Calibration}
\label{sec:wtumi_sensing}
WT-UMI uses a TouchTronix SensX~187 sensor on the chest, SensX~160 sensors on the palm end-effectors~\cite{touchtronix_sensx}, and custom $16 \times 26$ FlexiTac-style sensors on the forearms~\cite{vitac}. These thin-film piezoresistive sensors are composed of flexible pressure-sensitive layers and a row--column electrode grid that enables scalable multiplexed sensing: local pressure compresses the conductive material and changes electrical resistance, and a row--column scan reconstructs a dense 2D tactile activation map. An $M \times N$ array uses only $M+N$ signal lines, reducing wiring complexity, and the resulting map serves as a distributed representation of local contact intensity.

\begin{figure}[t]
\centerline{\includegraphics[width=0.6\textwidth]{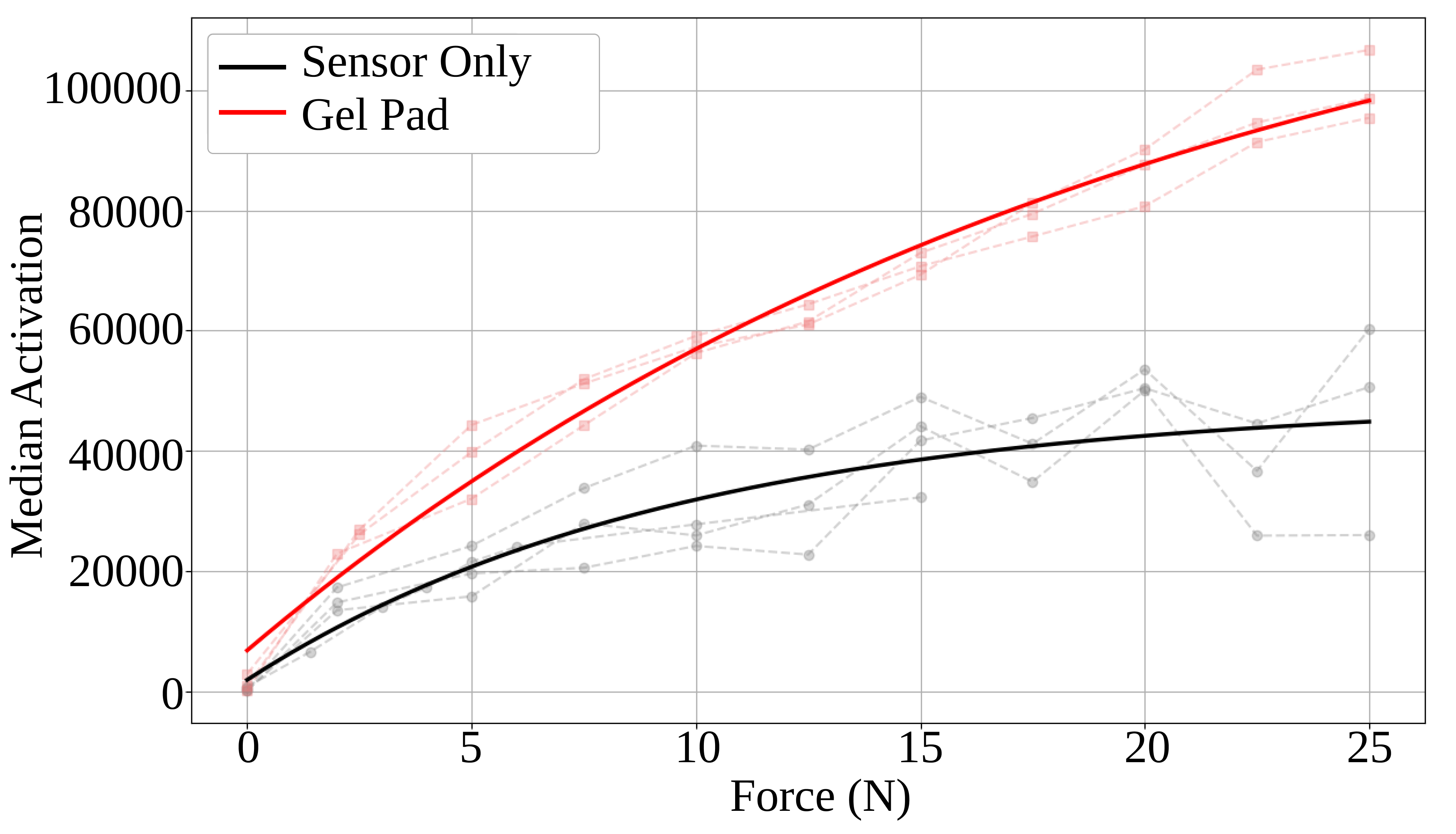}}
\caption{Force calibration of the SensX~160 palm sensor with and without a gel pad. The bare sensor response rises quickly and saturates near $20$\,N. The gel-pad configuration distributes contact pressure more evenly, yielding a smoother and less rapidly saturating response while extending the usable calibration range.}
\label{fig:wtumi_calib}
\vspace{-0.15in}
\end{figure}

Both the forearm FlexiTac and palm SensX~160 sensors are calibrated. The palm calibration is characterized under controlled quasi-static normal loading from $0$ to $25$\,N, comparing direct loading on the bare sensor against loading through a compliant gel pad ($5$\,mm thickness, Shore-A-20 hardness). As shown in \ref{fig:wtumi_calib}, the bare sensor rises quickly and saturates near $20$\,N, whereas the gel pad spreads the contact load across the sensing surface, yielding a smoother and less rapidly saturating response that extends the usable calibration range. The compliant gel surface additionally increases contact friction, improving grasp stability while preserving a consistent force scale between WT-UMI demonstrations and robot deployment. We adopt the gel-pad configuration in all manipulation experiments. The calibration experiment reported here characterizes the palm sensors; the forearm and chest arrays also provide distributed contact localization and force information. Forearm force prediction is evaluated separately in the collaborative beam-transport task (\ref{sec:wtumi_results_collab}). The piezoresistive arrays measure scalar normal force and do not resolve shear forces or full six-dimensional contact wrenches. But it enables swappable sensor for future versions with lower build profile or richer quality that can detect shear forces and full 6D wrenches.

The contact normal force is calibrated from the tactile activation map. The zero-load reading is subtracted from each tactile frame to form a calibrated activation map $X(t) \in \mathbb{R}^{H \times W}$ over the taxel grid $\Omega=\{1,\ldots,H\}\times\{1,\ldots,W\}$. Active taxels above a fixed threshold $\tau_a$ form the index set $\mathcal{I}(t)$, whose values sum to an aggregate activation score $A(t)$:
\begin{equation}
\mathcal{I}(t) = \{(i,j)\in\Omega \mid X_{ij}(t) > \tau_a\},
\qquad
A(t) = \sum_{(i,j)\in\mathcal{I}(t)} X_{ij}(t).
\label{eq:wtumi_activation}
\end{equation}
The threshold $\tau_a$ is selected empirically from no-contact recordings to suppress sensor noise and kept fixed across calibration and deployment. The calibration fits an inverse-exponential response model between $A(t)$ and the applied normal force $f(t)$, capturing the rapidly saturating response of piezoresistive sensors:
\begin{equation}
A(t) = c_1\left(1 - \exp(-c_2 f(t))\right) + c_3,
\label{eq:wtumi_calib_model}
\end{equation}
where $c_1, c_2, c_3 \in \mathbb{R}$ are scalar parameters fitted from the calibration data. Inverting Eq.~\eqref{eq:wtumi_calib_model} gives an intermediate force estimate $\tilde f(t)$, and a linear post-calibration scaling and offset $(\alpha,\beta)$ followed by clipping to the operating range $[0, f_{\max}]$ yields the final estimate
\begin{equation}
\tilde f(t) = -\frac{1}{c_2}\log\!\left(1 - \operatorname{clip}\!\left(\tfrac{A(t)-c_3}{c_1},\,0,\,1-\epsilon\right)\right),
\qquad
f^{\mathrm{m}}(t) = \operatorname{clip}\!\left(\alpha\,\tilde f(t) + \beta,\,0,\,f_{\max}\right),
\label{eq:wtumi_calib_inverse}
\end{equation}
where $\epsilon > 0$ keeps the logarithm argument strictly positive. The same calibration is applied on the WT-UMI hardware and on the robot, so a fixed force value carries the same physical meaning during training and inference.

\subsection{Whole-Body Tactile UMI for Data Collection}
\label{sec:wtumi_data}
\begin{figure}[t]
\centerline{\includegraphics[width=0.9\textwidth]{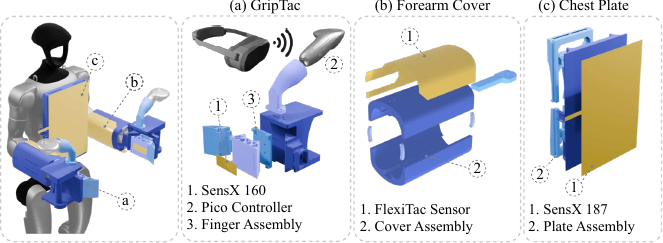}}
\caption{\textbf{WT-UMI} includes GripTac end-effectors, forearm covers, and a chest plate, each equipped with a thin-film tactile sensor.}
\label{fig:wtumi_sensors}
\vspace{-0.15in}
\end{figure}

WT-UMI is an extensible wearable system for capturing operator motion and distributed tactile feedback. As shown in \ref{fig:wtumi_sensors}, it integrates tactile sensors on hardware modules shared between humans and humanoids: a chest plate, forearm covers, and handheld GripTacs, which are tactile-instrumented handheld interfaces with interchangeable end-effectors. We use a palm-like end-effector with a soft, deformable surface to enlarge contact area and increase friction. By using the same sensing modules for human data collection, robot teleoperation, and robot execution, WT-UMI reduces the domain gap between demonstration and deployment.

WT-UMI supports demonstration collection in two modes. In \emph{teleoperation mode}, WT-UMI is mounted on the robot, and an operator streams bimanual pose commands through a PICO VR headset~\cite{pico4ultra2023} and two handheld controllers via XRoboToolkit~\cite{xrobotoolkit2025}. In \emph{human mode}, a human operator wears the chest plate, forearm covers, and GripTacs, with a PICO controller on each GripTac to track the bimanual pose, requiring no robot in the loop. Both modes stream hand poses, tactile images, and calibrated force measurements; teleoperation additionally records robot proprioception and VR commands.

Before each collection session, kinematic tracking and contact constraints calibrate the offset between the operator's head and hands. Transformations among the PICO headset, the handheld GripTacs, and the chest plate express demonstrations from different operators in a consistent frame before mapping them to robot poses. During calibration, the operator wears the headset and chest plate while holding the GripTacs, and presses each GripTac tip against the chest tactile sensor, activating exactly one cell. The resulting tactile image localizes the GripTac contact on the chest plate; combined with the known GripTac and chest-plate geometry, this estimates the transformation from the chest-plate center to the handheld GripTac, $T^s_{\mathrm{Chest}\rightarrow\mathrm{Hand}}$, with $s \in \{l, r\}$. Using the known transformation from the robot base to the mounted chest-plate center, $T_{\mathrm{Base}\rightarrow\mathrm{Chest}}$, the corresponding robot end-effector pose is
\begin{equation}
T^s_{\mathrm{Base}\rightarrow\mathrm{Hand}} = T_{\mathrm{Base}\rightarrow\mathrm{Chest}} \cdot T^s_{\mathrm{Chest}\rightarrow\mathrm{Hand}}.
\label{eq:wtumi_retarget}
\end{equation}
During data inspection, the resulting bimanual targets are passed to an inverse-kinematics solver, and poses that exceed the feasibility tolerance or risk self-collision are rejected; retained demonstrations are then verified in simulation or on robot hardware. Training uses these retargeted robot data rather than raw human motion, reducing dependence on operator-specific geometry.

\subsection{Force-Conditioned Target-Pose Correction}
\label{sec:wtumi_correction}
We introduce a force-conditioned target-pose correction module that converts the measured human hand pose $T^{s,\mathrm{m}}$, contact force $f^{s,\mathrm{m}}$, and tactile observation $\mathbf{I}^{s,\mathrm{m}}$ into a robot target pose, where $s\in\{l,r\}$ indexes the left and right hands and $\mathrm{m}$ denotes measured quantities. After pre-training, this module is applied to human demonstrations to generate robot-executable action labels that supervise planner training.

The correction module is pre-trained offline using paired robot teleoperation and human demonstration trajectories. For both data sources, we use contact forces to infer contact modes, such as right-contact, both-contact, and left-contact. We then match contact modes between teleoperation and human data using Dynamic Time Warping (DTW) over the end-effector motion. DTW is sufficient for the simple, rhythmic motions considered here; more complex tasks may require alternative sequence-alignment methods. Given the aligned contact modes, the target-pose offset from the robot teleoperation data supervises the corresponding offset in the human data, gated by inferred contact state: only hands that are in contact for that segment receive a nonzero target-pose offset. For each hand $s$, a lightweight CNN--GRU correction network $g_\theta$ predicts the hand-specific translation offset from the measured tactile image, pose, and force,
\begin{equation}
\Delta\mathbf{p}^{s}=g_\theta(\mathbf{I}^{s,\mathrm{m}},T^{s,\mathrm{m}},f^{s,\mathrm{m}})\in\mathbb{R}^3.
\label{eq:wtumi_offset}
\end{equation}
After pre-training, the learned correction module is applied to human demonstration data to generate target-pose action labels. The corrected target pose in SE(3) is
\begin{equation}
\tilde{T}^{s}=T^{s,\mathrm{m}}\oplus\mathbf{1}[f^{s,\mathrm{m}}\ge f_{\text{threshold}}]\,\Delta\mathbf{p}^{s},
\label{eq:wtumi_corrected_pose}
\end{equation}
where $\oplus$ shifts the translation component of the pose by the offset. The resulting corrected poses $\tilde{T}^{s}$ serve as robot-executable action labels for training the action denoiser.

\begin{figure}[t]
\centerline{\includegraphics[width=0.95\textwidth]{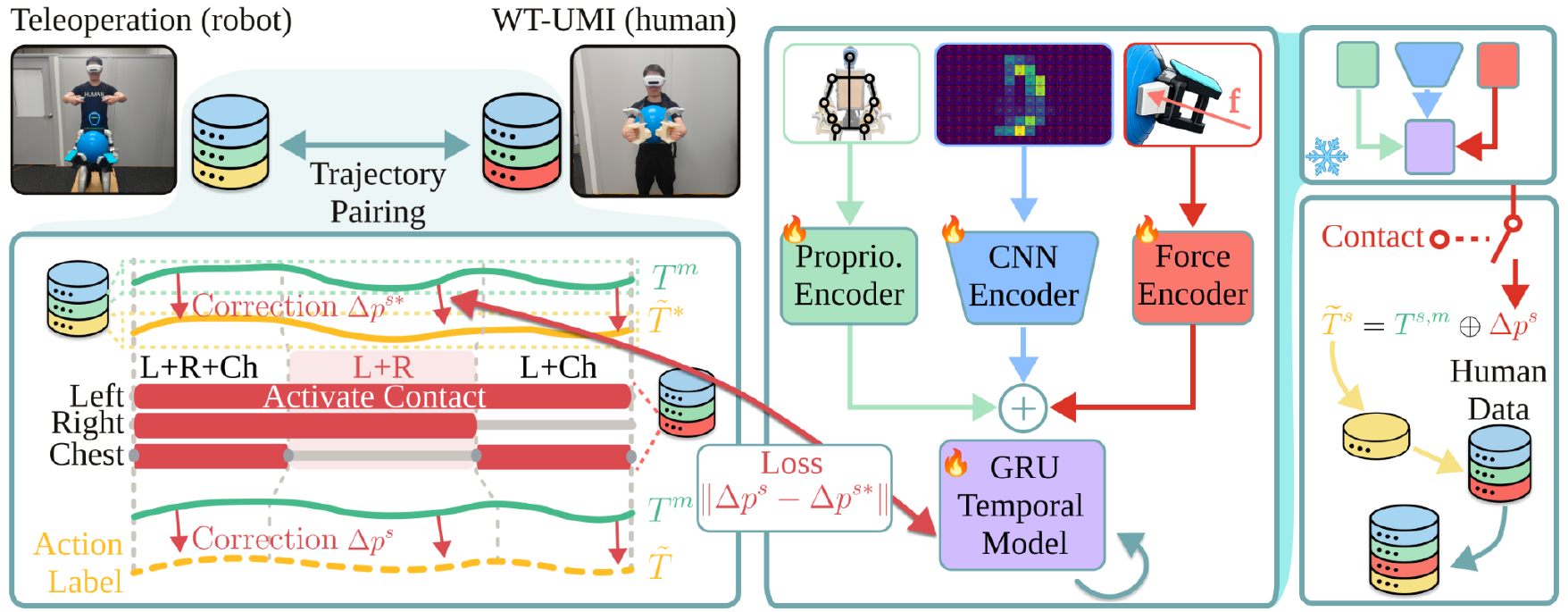}}
\caption{Target-pose correction training. Contact-mode-aligned teleoperation and human segments supervise contact-gated offsets that convert human poses into robot target poses.}
\label{fig:wtumi_correction}
\vspace{-0.15in}
\end{figure}

The module is trained separately for each task on paired teleoperation and human trajectories (\ref{fig:wtumi_correction}). It assumes that human contact strategies can be transferred through translation-only pose corrections, with rotational adjustments supplied by the admittance controller. This assumption is appropriate for our palm- and forearm-based tasks with simple, discrete contact modes; \ref{sec:wtumi_results_se3} evaluates full SE(3) correction. Trajectories are paired by matching their contact-mode sequences (e.g., left hand $+$ right hand $+$ chest $\rightarrow$ left hand $+$ right hand $\rightarrow$ left hand $+$ chest $\rightarrow \cdots$), so teleoperation and human segments share the same ordered contact structure before offset supervision. Each teleoperation trajectory supplies a residual target-pose offset $\Delta\mathbf{p}^{\star}$ in the local end-effector frame, which supervises the predicted correction. The training data are resampled to $50$\,Hz; the model observes a $0.2$\,s history of hand pose, tactile images, contact force, and binary contact state for both hands, and predicts corrections over a $0.1$\,s future target-pose chunk. Because the correction is applied offline to recorded demonstrations, the human hand poses in the upcoming prediction chunk are also available and provided as input. The network combines a per-block Convolutional Neural Network (CNN) tactile encoder with a Gated Recurrent Unit (GRU) temporal model: a two-layer GRU over the history and a two-layer bidirectional GRU over the future window, both with hidden dimension $256$ and dropout $0.3$. An auxiliary head predicts the contact-segment class and per-hand contact direction, while the output head regresses the translational offset $\Delta\mathbf{p}=[\Delta\mathbf{p}^l;\Delta\mathbf{p}^r]\in\mathbb{R}^{6}$. Training minimizes a contact-masked Smooth-$L_1$ offset loss, a Smooth-$L_1$ temporal-smoothness loss on consecutive offset differences, and the auxiliary contact losses, optimized with AdamW using cosine learning-rate decay, warmup, gradient clipping, and early stopping on validation offset error and auxiliary contact accuracy.

\subsection{Force-Supervised Planner}
\label{sec:wtumi_planner}
Existing generative planners often do not explicitly model contact forces, leaving force regulation to position-based feedback alone. WT-UMI addresses this limitation by providing calibrated normal force from tactile measurements. We use these calibrated forces as supervision to augment the planner with a force head that predicts future contact-force trajectories; during deployment, the predicted forces serve as explicit references for a tactile-based admittance controller.

The planner uses $400$\,ms observation and prediction horizons ($H_o = H_a = 20$), where $H_o$ and $H_a$ denote the numbers of observation and prediction time steps. Each predicted pose and force chunk contains $20$ time steps sampled at $50$\,Hz; this sampling rate is distinct from the asynchronous inference rate. At time step $t$, the model consumes an observation history $\mathbf{o}_{t-H_o+1:t}$, where each $\mathbf{o}_t = (\mathbf{I}^{l}_t,\, \mathbf{I}^{r}_t,\, \mathbf{I}^{ch}_t,\, \mathbf{T}_t)$ consists of normalized tactile images from the left/right contact sites and the chest, alongside the measured bimanual end-effector pose $\mathbf{T}_t = [\mathrm{vec}(T^{l}_t);\, \mathrm{vec}(T^{r}_t)]\in \mathbb{R}^{18}$, where $\mathrm{vec}(\cdot)$ maps each SE(3) pose $T^{s}_t \in \mathbb{R}^{4\times4}$, $s\in\{l,r\}$, to its 3D translation and 6D continuous rotation representation~\cite{zhou2019continuity} in $\mathbb{R}^{9}$. A Vision Transformer (ViT)~\cite{VIT} encodes the channel-stacked tactile images and a two-layer Multilayer Perceptron (MLP) encodes the hand poses; per time step, their outputs are concatenated into one observation token of width $D = 320$. The hand pose is additionally linearly projected to a separate pose token of the same width, yielding a conditioning sequence $\mathbf{h}\in\mathbb{R}^{2H_o\times D}$. Conditioned on $\mathbf{h}$, a transformer-based action denoising policy (either flow matching~\cite{lipman2023flow} or diffusion~\cite{diffusion_policy}) generates a predicted action chunk $\mathbf{a} = [\mathbf{T}_{t+1},\dots,\mathbf{T}_{t+H_a}] \in \mathbb{R}^{H_a \times 18}$, supervised by the target-pose-corrected labels $\mathbf{a}^{\star}=[\tilde{\mathbf{T}}_{t+1},\dots,\tilde{\mathbf{T}}_{t+H_a}]\in \mathbb{R}^{H_a \times 18}$ from \ref{sec:wtumi_correction}.

In this formulation, tactile images are policy observations rather than planner outputs. We choose calibrated normal force as the predicted tactile quantity because it has physical units, is directly comparable across human demonstrations and robot deployment, and can be consumed immediately by the admittance controller. Predicting full tactile images remains a natural extension for anticipatory contact planning, especially when future contact location, pressure distribution, or slip precursors are more informative than aggregate normal force.

Unlike the action denoising policy, a force head performs direct regression without a diffusion process. It is a cross-attention decoder that takes $H_a$ learnable positional queries $\mathbf{Q} \in \mathbb{R}^{H_a \times D}$, whose $k$-th row is a positional query embedding for predicting the force at the $k$-th future action step, and uses the shared observation embeddings $\mathbf{h}$ as keys and values. By performing self-attention across the queries followed by cross-attention to $\mathbf{h}$, the decoder models the temporal force dynamics before a linear projection maps the output to the predicted force chunk of both hands $\mathbf{F} \in \mathbb{R}^{H_a \times 2}$. Supervision is provided by the calibrated ground-truth forces $\mathbf{F}^{\star} \in \mathbb{R}^{H_a \times 2}$, with a softplus projection enforcing physical non-negativity. Because the force head directly outputs a force trajectory, its gradients flow back through $\mathbf{h}$, allowing the shared encoder to be jointly optimized by both the action and force objectives. To suppress gradient spikes from contact onset, the force head is supervised via an element-wise Huber loss ($\mathrm{SmoothL1}$), yielding the combined objective
\begin{equation}
\mathcal{L} = \mathcal{L}_{\text{gen}}(\mathbf{a}, \mathbf{a}^{\star}) + \lambda_F\,\mathrm{SmoothL1}\!\left(\mathbf{F},\mathbf{F}^{\star}\right),
\label{eq:wtumi_loss}
\end{equation}
where $\mathcal{L}_{\text{gen}}$ is a flow-matching or diffusion denoising MSE and $\lambda_F$ weights the action denoiser against the force head.

The force head uses a transformer decoder with two layers and four attention heads at hidden dimension $D=320$, adding roughly $0.5$M parameters. Crucially, the action denoiser and the force head are split into independent decoders that share only the encoder: the action denoiser takes the noisy action chunk $\mathbf{a}_k$ and the denoising step $k$ as input, while the force head reads only the clean observation embedding $\mathbf{h}$. Keeping the decoders independent keeps the force head's input free of the noise schedule while still letting force supervision shape the shared encoder. During training, both the flow-matching and diffusion backbones undergo the same number of gradient steps using AdamW, cosine learning-rate decay, and an exponential moving average over weights. Image augmentations include additive noise, channel dropout, and random patch masking. The sensor cap $f_{\max}{=}20$\,N is fitted to the calibrated palm range, and the calibration-aware force term is enabled throughout training with $\lambda_F = 1.0$. At inference, the force-conditioned target-pose correction of \ref{sec:wtumi_correction} is gated by the anticipated per-hand force using $f_{\text{threshold}} = 0.5$\,N.

\subsection{Tactile-based Admittance Controller}
\label{sec:wtumi_lowlevel}
A similar hierarchical upper--lower decomposition is later used for autonomous loco-manipulation in \ref{chapter:DPFT}. Here, the low-level controller is split into a lower body ($12$ leg joints and $3$ waist joints) and an upper body ($14$ arm joints). The lower body tracks pelvis-frame velocity $\mathbf{v} = [v_x, v_y, \omega_z]^\top$ with a pre-trained RL locomotion policy~\cite{gr00tn1_2025} that maintains balance under arbitrary upper-body motion. In contrast to \ref{chapter:DPFT}, where a vision-conditioned diffusion planner issues base and hand commands, here the upper body is regulated by a tactile-based admittance controller driven by the force-supervised planner. At each control step $t$, we track the predicted end-effector pose target $\tilde{T}^{s}_{t}$ and a normal-force reference $f^{s}_{t}$ for hand $s$. A proportional admittance controller refines the planner-generated pose using both normal-force and contact-centroid feedback,
\begin{equation}
T^{s}_{\text{cmd},t} = \tilde{T}^{s}_{t}\,\Delta T^{s}_t,
\label{eq:wtumi_adm_law}
\end{equation}
where $T^{s}_{\text{cmd},t}$ is the commanded target pose and $\Delta T^{s}_t \in \mathbb{R}^{4\times4}$ is the corrective SE(3) increment for palm $s$ expressed in the palm frame. The increment contains a rotation $\Delta R^{s}_{xy,t}$ that re-centers the contact centroid and a translation $\Delta\mathbf{p}^{s}_{z,t}$ that tracks the reference normal force, both following proportional control laws,
\begin{equation}
\Delta R^{s}_{xy,t}=\operatorname{Exp}\!\big([\,K_R(\mathbf{c}^{s}_{t}-\mathbf{c}^{s,\mathrm{m}}_{t});\,0\,]_{\times}\big),
\qquad
\Delta \mathbf{p}^{s}_{z,t}=K_F(f^{s}_{t}-f^{s,\mathrm{m}}_{t})\,[0,0,1]^\top,
\label{eq:wtumi_adm_gains}
\end{equation}
where $\mathbf{c}^{s}_{t}, \mathbf{c}^{s,\mathrm{m}}_{t} \in \mathbb{R}^2$ are the desired and measured contact centroids in the sensor frame, $f^{s}_{t}$ and $f^{s,\mathrm{m}}_{t}$ are the reference (planner-predicted) and measured normal forces, $K_R\in\mathbb{R}^{2\times2}$ is the contact-centering gain matrix, and $K_F$ is a scalar normal-force gain. The rotation re-centers contact on the palm, while the translation tracks the normal-force reference. We use $K_R=\mathrm{diag}(0.5,0.5)$ and $K_F=0.001$, with maximum corrective translation $\Delta p_{\max}=0.04$\,m and rotation $\Delta\theta_{\max}=0.5$\,rad. When contact is lost under an appropriate force prediction, the controller shifts the target along the sensor normal until contact is restored or the translation bound is reached. Incorrect force predictions can therefore still cause contact loss or excessive force. The commanded poses are passed to an optimization-based inverse-kinematics solver~\cite{pink}, $q_{\text{cmd},t} = \mathrm{IK}(T^{l}_{\text{cmd},t},\,T^{r}_{\text{cmd},t},\,q_t,\,\dot{q}_t)$, and joint-level tracking uses a PD controller with gravity compensation, $\tau_t = K_p(q_{\text{cmd},t} - q_t) + K_d(-\dot{q}_t) + G(q_t)$, where $q_t, \dot{q}_t$ are measured joint positions and velocities, $\tau_t$ is the commanded torque, $K_p, K_d$ are joint-PD gains, and $G(q_t)$ is the gravity-compensation torque.

\begin{figure}[t]
\centerline{\includegraphics[width=0.45\textwidth]{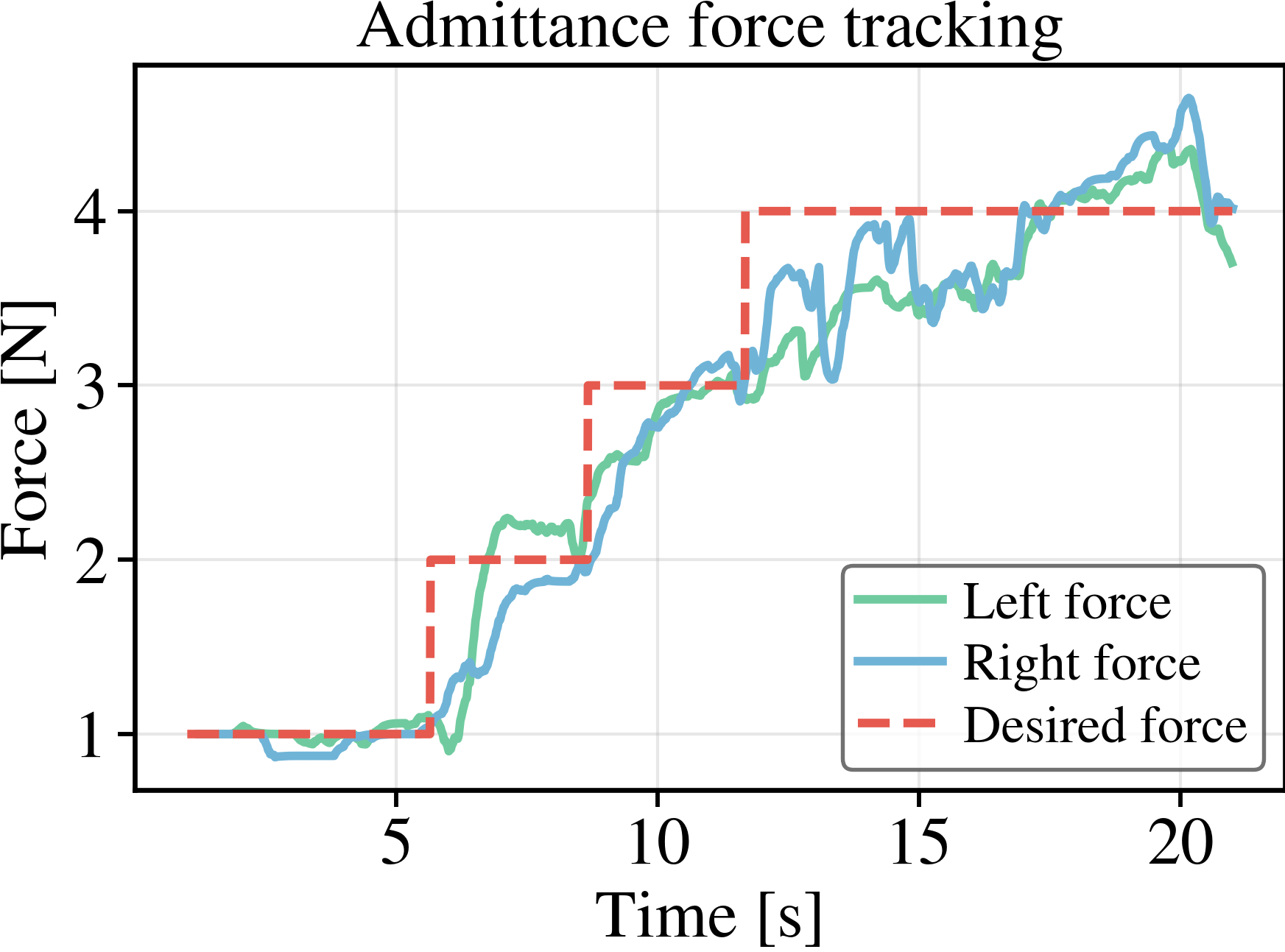}}
\caption{Admittance controller force-tracking.}
\label{fig:wtumi_adm_tracking}
\vspace{-0.15in}
\end{figure}

To evaluate the closed-loop force regulation in isolation, the robot holds a yoga ball between its chest and palms under a fixed pose target, with both palms initially contacting the ball at $1$\,N, and then tracks step changes in the desired contact force from $1$\,N to $4$\,N. As shown in \ref{fig:wtumi_adm_tracking}, both palms reach each setpoint with steady-state errors within $3.5\%$, confirming that the proportional admittance law tracks the planner-predicted force reference accurately.

\section{Experiment Setup}
\label{sec:wtumi_setup}

\paragraph{Tasks.}
\label{sec:wtumi_tasks}
We demonstrate five contact-rich tasks in three categories, with representative deployments shown in \ref{fig:wtumi_deployment}. For \textbf{deformable object manipulation}, \emph{yoga ball manipulation} (\textbf{T1}) stabilizes and repositions an inflated ball, and \emph{pillow reorientation} (\textbf{T2}) reorients a soft pillow. For \textbf{bulky rigid object manipulation}, \emph{bucket manipulation} (\textbf{T3}) repositions a cone-shaped container with diverse loads. For \textbf{human--humanoid collaborative manipulation}, \emph{beam transport} (\textbf{T4}) and \emph{table transport} (\textbf{T5}) require the robot and a human partner to jointly carry a beam and a table, respectively, while inferring human physical intent from tactile feedback and following the partner's motion. Tasks T1--T3 use palm and chest sensors, T4 uses forearm and chest sensors, and T5 uses palm sensors mounted on a custom end-effector. Each task uses a separate policy and target-pose correction module trained on task-specific data; cross-task generalization is not evaluated.

\paragraph{Policy Baselines.}
We train four widely used behavior-cloning policies as baselines: \emph{ViT-FMT}~\cite{lipman2023flow}, \emph{ViT-DiT}~\cite{diffusion_policy}, $\pi_{0.5}$~\cite{pi05}, and $\Psi_0$~\cite{wei2026psi0openfoundationmodel}. \emph{ViT-FMT} and \emph{ViT-DiT} share the same vision transformer encoder but differ in the generative process: flow matching for \emph{ViT-FMT} and denoising diffusion for \emph{ViT-DiT}; their ViT tactile encoder is trained end-to-end with both heads. $\pi_{0.5}$ and $\Psi_0$ are fine-tuned foundation policies: $\pi_{0.5}$ is a vision--language--action model conditioned on tactile images and a fixed per-task language instruction, adopting the OpenPI-based implementation in FASTER~\cite{lu2026fasterrethinkingrealtimeflow}; $\Psi_0$ is a humanoid foundation policy conditioned on tactile images and proprioception. For both foundation policies, encoders are frozen during fine-tuning.

\paragraph{Hardware Setup.}
We use a Unitree G1 humanoid for teleoperation data collection and policy deployment. Raw tactile data streams at $100$\,Hz and proprioception at $500$\,Hz; both are resampled to a synchronized $50$\,Hz stream during training and deployment. The ViT-FMT and ViT-DiT policies run on an RTX~4500 and perform asynchronous inference with a $100$\,ms chunk latency. $\pi_{0.5}$ and $\Psi_0$ similarly run at $12$\,Hz with $83$\,ms per chunk on an RTX~6000. The low-level tactile-based admittance controller runs at $200$\,Hz, with a full force-regulation-loop latency of $20$\,ms. Controller gains, loss weights, and tactile calibration parameters remain fixed across tasks and trials. For ViT-DiT, we use Denoising Diffusion Implicit Models (DDIM)~\cite{song2021denoising} for faster inference. All policies adopt training-time real-time chunking (RTC)~\cite{black2025trainingtimeactionconditioningefficient} to compensate for inference delay.

\section{Results}
\label{sec:wtumi_results}
\begin{figure}[t]
\centerline{\includegraphics[width=0.99\textwidth]{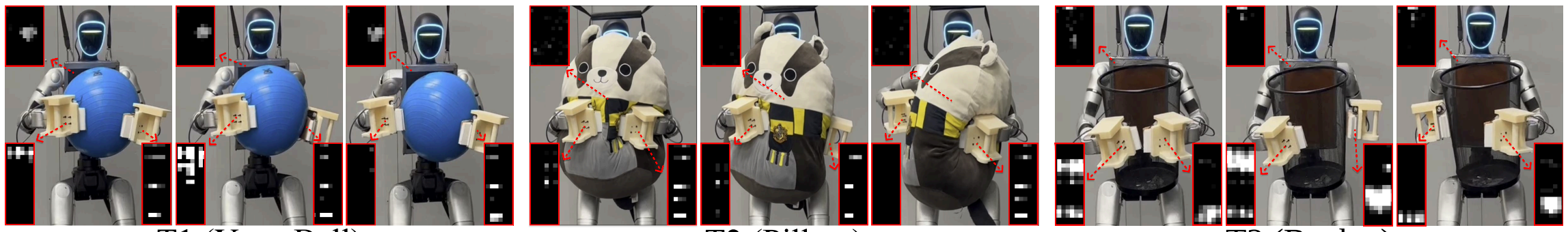}}
\caption{Deployment on three large-area contact-rich manipulation tasks: yoga-ball manipulation, pillow reorientation, and bucket manipulation.}
\label{fig:wtumi_deployment}
\vspace{-0.15in}
\end{figure}

We evaluate WT-UMI across the five tasks introduced in \ref{sec:wtumi_tasks}. Representative deployments for T1--T3 are shown in \ref{fig:wtumi_deployment} and used for quantitative evaluation throughout this section. The collaborative tasks T4 and T5 are detailed in \ref{sec:wtumi_results_collab}.

\subsection{Force Head Evaluation}
\label{sec:wtumi_results_force}
This section investigates whether the force head produces accurate, smooth, and temporally aligned force predictions. We evaluate the force head (\ref{sec:wtumi_planner}) within a \emph{ViT-FMT} policy, using a 90/10 train--validation split across the teleoperation and human datasets from task T1, and evaluate the trained planner on $10$ held-out demonstrations. \ref{tab:wtumi_force_quality} reports force-prediction accuracy via RMSE, smoothness via the force-rate $|\mathrm{d}F/\mathrm{d}t|$ RMS averaged across episodes, and temporal alignment via the cross-correlation peak offset between predicted and ground-truth forces. Overall, the force head tracks the ground-truth force trajectory with low prediction error and small temporal lag. Notably, the force head trained on human demonstrations achieves an RMSE of $1.05$\,N out of a $5.56$\,N force RMS and a $2.2\times$ lower temporal lag, with the force-rate RMS decreasing to $3.74$\,N/s compared to the $19.80$\,N/s from the force head trained on teleoperation data.

\begin{table}[t]
\centering
\footnotesize
\setlength{\tabcolsep}{3pt}
\caption{Force signal quality. Human demonstrations exhibit more accurate and smoother force profiles than teleoperation, improving the quality of supervision for learning.}
\label{tab:wtumi_force_quality}
\begin{tabular}{l|cccc}
\toprule
\textbf{Source} & \textbf{Force RMSE (N)} & \textbf{Lag (ms)} & \multicolumn{2}{c}{\textbf{Force Rate RMS (N/s)}}\\
\cmidrule(lr){4-5}
 &     &   & \textbf{Meas.} & \textbf{Pred.} \\
\midrule
Human  & \textbf{1.05} & \textbf{68}  & \textbf{5.86}  & \textbf{3.74}  \\
Teleop & 2.07          & 151          & 30.62          & 19.80          \\
\bottomrule
\end{tabular}
\end{table}

\subsection{Effect of Training Data Sources and Target-Pose Correction}
\label{sec:wtumi_results_data}
This section evaluates the complementary roles of teleoperated robot data and human demonstrations collected with WT-UMI. We evaluate ViT-FMT, the best-performing baseline (\ref{sec:wtumi_results_backbone}), and adopt two data-source settings: pure teleoperation data (\textbf{Tel.}) and a combined dataset (\textbf{Comb.}) that augments teleoperation with target-pose-corrected human demonstrations. Across T1--T3, \textbf{Tel.} contains $2.2$, $2.2$, and $1.4$\,min, while \textbf{Comb.} contains $13.2$, $15.7$, and $8.8$\,min; thus teleoperation accounts for only $16.7\%$, $14.0\%$, and $15.9\%$ of the corresponding \textbf{Comb.} demonstrations. Each setting is evaluated with $N=25$ trials per task. In each trial the object is rotated once, starting and ending in the same hugging pose. We report (i) success rate; (ii) contact off-center drift, the distance between the measured contact centroid and the geometric center of the tactile sensor; (iii) mean contact force during required contact phases; and (iv) translational and rotational accelerations of end-effector poses, which assess motion smoothness.

\begin{table}[t]
\centering
\footnotesize
\setlength{\tabcolsep}{2pt}
\renewcommand{\arraystretch}{0.92}
\caption{Effect of data source over tasks \textbf{T1}--\textbf{T3}. Best values are highlighted in bold; parentheses in the \textbf{Comb.} columns report one-sided Mann--Whitney $p$-values. Smaller $p$-values provide stronger evidence that \textbf{Comb.} outperforms \textbf{Tel.} Values are rounded as in the source study; $.00$ does not denote an exact zero.}
\label{tab:wtumi_data_source}
\begin{tabular*}{\textwidth}{@{\extracolsep{\fill}} l | cc cc cc cc cc @{}}
\toprule
\textbf{Task}
& \multicolumn{2}{c}{\textbf{Succ. (\%)}}
& \multicolumn{2}{c}{\textbf{Cont. Drift (mm)}}
& \multicolumn{2}{c}{\textbf{Cont. Force (N)}}
& \multicolumn{2}{c}{\makecell{\textbf{Trans. Accel.}\\\textbf{($\text{m/s}^2$)}}}
& \multicolumn{2}{c}{\makecell{\textbf{Rot. Accel.}\\\textbf{($\text{rad/s}^2$)}}}\\
\cmidrule(lr){2-3}\cmidrule(lr){4-5}\cmidrule(lr){6-7}\cmidrule(lr){8-9}\cmidrule(lr){10-11}
\multicolumn{1}{l}{\textbf{Data~$\rightarrow$}\hspace{-0.6em}} & Tel. & Comb. & Tel. & Comb. & Tel. & Comb. & Tel. & Comb. & Tel. & Comb. \\
\midrule
T1 & \textbf{100.0} & \textbf{100.0} & 20.41 & \textbf{18.12}{\scriptsize\,(.00)} & 3.27 & \textbf{4.77}{\scriptsize\,(.00)} & 4.08 & \textbf{3.14}{\scriptsize\,(.00)} & 22.93 & \textbf{20.29}{\scriptsize\,(.02)} \\
T2 & \textbf{100.0} & \textbf{100.0} & 24.61 & \textbf{21.04}{\scriptsize\,(.00)} & 0.17 & \textbf{0.52}{\scriptsize\,(.00)} & 6.03 & \textbf{1.87}{\scriptsize\,(.00)} & 34.54 & \textbf{12.63}{\scriptsize\,(.00)} \\
T3 & 60.0  & \textbf{80.0}        & 26.01 & \textbf{25.00}{\scriptsize\,(.00)} & 0.66 & \textbf{0.95}{\scriptsize\,(.00)} & 3.49 & \textbf{1.85}{\scriptsize\,(.00)} & 27.82 & \textbf{14.05}{\scriptsize\,(.00)} \\
\bottomrule
\end{tabular*}
\end{table}

\ref{tab:wtumi_data_source} presents the quantitative results. The \textbf{Tel.} policy alone achieves a strong $86.7\%$ average success rate, because teleoperation provides robot-feasible action labels. However, teleoperation lacks accurate force feedback during data collection, reflected in occasional missed contacts or over-pressing, which result in larger contact-region drift of about $11\%$ on average compared to \textbf{Comb.} Human demonstrations offer valuable contact information because the demonstrator directly perceives interaction forces and naturally regulates contact, but pure human data is not directly robot-executable due to missing action labels and the human--humanoid kinematics gap. Our target-pose correction addresses this by converting human data into robot-feasible pose targets. In the \textbf{Comb.} policy, the corrected human data supplements teleoperation by reducing contact center drift and increasing mean contact force by approximately $2\times$ on average. Motion smoothness also improves consistently: translational and rotational accelerations are reduced by $46.3\%$ and $41.5\%$ on average, respectively, and the success rate increases by $20\%$ for T3, where the most common failure mode is loss of contact and motion freeze due to out-of-distribution observations. Notably, the correction module requires only a small amount of teleoperation data to convert a much larger set of human demonstrations into robot-executable data, demonstrating its teleoperation-data efficiency.

\subsection{Effect of Force-Conditioned Admittance Control}
\label{sec:wtumi_results_admittance}
This section evaluates whether tactile-based admittance control improves contact quality and motion stability. All policies are trained on the combined human and teleoperation dataset. The results with and without admittance control are reported in \ref{tab:wtumi_backbones}. For ViT-FMT, enabling admittance control improves motion smoothness on average, reducing translational acceleration by $10.4\%$ and rotational acceleration by $9.0\%$ on average. Contact quality also improves: contact off-center drift decreases by $10.9\%$ and mean contact force increases by approximately $3\%$ on average. The exception in \textbf{T2} is due to the noisy signal, as the pillow is soft, and the tactile measurement is close to minimum contact force.

\subsection{Effect of Policy Backbone}
\label{sec:wtumi_results_backbone}
We compare the baseline policies $\pi_{0.5}$, $\Psi_0$, ViT-DiT, and ViT-FMT in \ref{tab:wtumi_backbones}. All policies are trained on the same combined dataset with force prediction, deployed with and without the admittance controller. Each setting is evaluated with $N=25$ trials per task.

Overall, ViT-FMT achieves the highest average success rate $97.3\%$ and the best motion smoothness, with the lowest translational and rotational accelerations ($2.07~\text{m/s}^2$ and $14.18~\text{rad/s}^2$, respectively), and most closely reproduces the motion patterns in the dataset.
$\pi_{0.5}$ ($4.62~\text{m/s}^2$, $28.45~\text{rad/s}^2$) shows more jittery motion, often with rapid end-effector swings. The ViT-DiT ($2.90~\text{m/s}^2$, $18.35~\text{rad/s}^2$) and $\Psi_0$ ($3.38~\text{m/s}^2$, $22.83~\text{rad/s}^2$) tend to get stuck in static poses or fail to continue the motion, leading to a lower success rate ($62.7\%$), with $\Psi_0$ failing \textbf{T3} entirely.
We also tested variants with ResNet-based~\cite{he2016resnet} tactile encoders, but all variants including TACT~\cite{Murooka_TACT} trained poorly and failed on the tasks.

Among these experiments, the dominant failure mode is stalling, in which the robot ceases motion and fails to complete the task within the allotted time. Loss of contact is also a common failure mode. Another failure mode is that, over repeated trials, the object gradually drifts from its nominal holding configuration, eventually resulting in dropping the object. Degraded or noisy tactile measurements occasionally contribute to failures and primarily result from sensor-connector issues.

\begin{table}[t]
\centering
\footnotesize
\setlength{\tabcolsep}{2pt}
\renewcommand{\arraystretch}{0.92}
\caption{Performance across four policy backbones over tasks \textbf{T1}, \textbf{T2}, and \textbf{T3}. Each metric is evaluated without (w/o) and with (w/) our admittance control. The better or tied value is bolded. Values in the parentheses in {w/} cells show the one-sided Mann-Whitney $p$, lower meaning stronger evidence that our admittance control improves performance.}
\label{tab:wtumi_backbones}
\begin{tabular}{ll | cc  cc  cc  cc  cc @{}}
\toprule
\textbf{Policy}
& \textbf{Task}
& \multicolumn{2}{c|}{\begin{tabular}{c}\textbf{Succ. Rate}\\\textbf{(\%)}\end{tabular}}
& \multicolumn{2}{c|}{\begin{tabular}{c}\textbf{Cont. Drift}\\\textbf{(mm)}\end{tabular}}
& \multicolumn{2}{c|}{\begin{tabular}{c}\textbf{Cont. Force}\\\textbf{(N)}\end{tabular}}
& \multicolumn{2}{c|}{\begin{tabular}{c}\textbf{Smooth.-Trans.}\\\textbf{($\mathrm{m/s}^2$)}\end{tabular}}
& \multicolumn{2}{c}{\begin{tabular}{c}\textbf{Smooth.-Rot.}\\\textbf{($\mathrm{rad/s}^2$)}\end{tabular}} \\
\cmidrule(lr){3-4}\cmidrule(lr){5-6}\cmidrule(lr){7-8}\cmidrule(lr){9-10}\cmidrule(lr){11-12}
\multicolumn{2}{c}{\textbf{Admi. (Ours) $\rightarrow$}}
& w/o & w/
& w/o & w/
& w/o & w/
& w/o & w/
& w/o & w/ \\
\midrule
\multirow{3}{*}{ViT-FMT}
& \textbf{T1}  & \textbf{100} & \textbf{100} & 18.12 & \textbf{15.67}{\scriptsize\,(.00)} & 4.77 & \textbf{5.50}{\scriptsize\,(.00)} & 3.14 & \textbf{2.98}{\scriptsize\,(.00)} & 20.29 & \textbf{18.56}{\scriptsize\,(.00)} \\
& \textbf{T2}  & \textbf{100} & \textbf{100} & 21.04 & \textbf{19.44}{\scriptsize\,(.56)} & \textbf{0.52} & 0.13{\scriptsize\,(1.00)} & \textbf{1.87} & 1.95{\scriptsize\,(1.00)} & \textbf{12.63} & 13.61{\scriptsize\,(1.00)} \\
& \textbf{T3}  & 80 & \textbf{92} & 25.00 & \textbf{22.08}{\scriptsize\,(.00)} & 0.95 & \textbf{1.61}{\scriptsize\,(.00)} & 1.85 & \textbf{1.29}{\scriptsize\,(.00)} & 14.05 & \textbf{10.38}{\scriptsize\,(.00)} \\
\midrule
\multirow{3}{*}{ViT-DiT}
& \textbf{T1} & \textbf{52} & \textbf{52} & 21.79 & \textbf{18.00}{\scriptsize\,(.00)} & 2.21 & \textbf{2.57}{\scriptsize\,(.02)} & 2.92 & \textbf{2.54}{\scriptsize\,(.00)} & 18.73 & \textbf{14.97}{\scriptsize\,(.00)} \\
& \textbf{T2} & \textbf{100} & \textbf{100} & 21.22 & \textbf{19.61}{\scriptsize\,(.00)} & 0.19 & \textbf{0.76}{\scriptsize\,(.00)} & 4.67 & \textbf{2.41}{\scriptsize\,(.00)} & 25.93 & \textbf{15.17}{\scriptsize\,(.00)} \\
& \textbf{T3} & \textbf{52} & \textbf{52} & 26.22 & \textbf{22.20}{\scriptsize\,(.00)} & \textbf{1.74} & 0.93{\scriptsize\,(1.00)} & 2.33 & \textbf{2.05}{\scriptsize\,(.00)} & 16.47 & \textbf{14.79}{\scriptsize\,(.00)} \\
\midrule
\multirow{3}{*}{$\pi_{0.5}$}
& \textbf{T1} & 88 & \textbf{92} & 21.51 & \textbf{11.78}{\scriptsize\,(.00)} & 2.53 & \textbf{2.80}{\scriptsize\,(.00)} & \textbf{4.69} & 4.82{\scriptsize\,(1.00)} & \textbf{29.33} & 29.79{\scriptsize\,(1.00)} \\
& \textbf{T2} & 68 & \textbf{76} & 19.78 & \textbf{15.97}{\scriptsize\,(.00)} & \textbf{0.50} & \textbf{0.50}{\scriptsize\,(1.00)} & 5.06 & \textbf{5.03}{\scriptsize\,(.39)} & 31.86 & \textbf{31.22}{\scriptsize\,(.00)} \\
& \textbf{T3} & \textbf{84} & 76 & 20.18 & \textbf{19.40}{\scriptsize\,(.14)} & 2.57 & \textbf{3.38}{\scriptsize\,(.00)} & 4.11 & \textbf{3.53}{\scriptsize\,(.00)} & 27.26 & \textbf{23.56}{\scriptsize\,(.06)} \\
\midrule
\multirow{3}{*}{$\Psi_0$}
& \textbf{T1} & 88 & \textbf{92} & 15.80 & \textbf{13.56}{\scriptsize\,(.00)} & \textbf{3.14} & 3.03{\scriptsize\,(.63)} & 4.20 & \textbf{3.27}{\scriptsize\,(.00)} & 27.06 & \textbf{23.01}{\scriptsize\,(.00)} \\
& \textbf{T2} & 88 & \textbf{96} & 20.48 & \textbf{18.69}{\scriptsize\,(.00)} & \textbf{2.50} & \textbf{2.50}{\scriptsize\,(1.00)} & \textbf{4.94} & 5.44{\scriptsize\,(.99)} & \textbf{32.12} & 33.04{\scriptsize\,(.28)} \\
& \textbf{T3} & 0 & 0 & - & - & - & - & - & - & - & - \\
\bottomrule
\end{tabular}
\end{table}

\subsection{Module Effectiveness on Recorded Data}
\label{sec:wtumi_results_replay}
To isolate the effects of target-pose correction and admittance control from the confounding influence of policy backbones and training setups, we replay the collected yoga-ball trajectories on the robot hardware under four configurations: raw human data, raw teleoperation data, target-pose-corrected human data, and corrected human data with admittance control enabled (\ref{tab:wtumi_replay}). The \textbf{Raw Human} configuration fails to complete the task due to loss of contact, confirming that human motion is not directly robot-executable because it lacks action labels. \textbf{Raw Teleoperation} is executable and completes the task at $85.35\%$. Our target-pose \textbf{Correction} improves the feasibility of human trajectories and raises the success rate to $89.29\%$. Adding admittance control (\textbf{Correction + Admittance}) further raises the success rate to $96.15\%$ and reduces contact center drift by $35.9\%$, at a moderate cost in motion smoothness, which is expected because admittance control introduces reactive adjustments based on contact feedback. This confirms that target-pose correction is the key component enabling the feasibility of human data, while admittance control improves contact quality.

\begin{table}[t]
\centering
\footnotesize
\setlength{\tabcolsep}{3pt}
\renewcommand{\arraystretch}{0.92}
\caption{Target-pose correction and admittance ablation via data replay on the yoga-ball task across four configurations. ``Failed'' indicates the configuration cannot complete the task. The best value is shown in bold for success rate, contact drift, and motion smoothness.}
\label{tab:wtumi_replay}
\begin{tabular}{l|ccccc}
\toprule
\textbf{Configuration}
& \makecell{\textbf{Success}\\\textbf{Rate (\%)}}
& \makecell{\textbf{Contact Center}\\\textbf{Drift (mm)}}
& \makecell{\textbf{Mean Contact}\\\textbf{Force (N)}}
& \makecell{\textbf{Smooth.-Trans.}\\\textbf{($\text{m/s}^2$)}}
& \makecell{\textbf{Smooth.-Rot.}\\\textbf{($\text{rad/s}^2$)}} \\
\midrule
Raw Human     & Failed         & --             & --            & --            & --             \\
Raw Teleoperation   & 85.35          & 15.59          & \textbf{3.30} & 3.93          & 26.42          \\
Correction      & 89.29          & 17.88          & 3.23          & \textbf{3.57} & \textbf{25.51} \\
Correction + Admittance & \textbf{96.15} & \textbf{11.46} & 3.02          & 4.06          & 27.15          \\
\bottomrule
\end{tabular}
\end{table}

\subsection{Translation-Only versus Full SE(3) Target-Pose Correction}
\label{sec:wtumi_results_se3}
Our pose correction module corrects only positional offsets and leaves the rotational offsets to the admittance controller. To analyze the effect of a full SE(3) correction, we also make SE(3) corrections to human data and train the ViT-FMT backbone on \textbf{T1}-\textbf{T3} (\ref{tab:wtumi_se3}).

\begin{table}[t]
\centering
\footnotesize
\setlength{\tabcolsep}{2pt}
\renewcommand{\arraystretch}{0.95}
\caption{Ablation study of SE(3) action representation and admittance control on the \textbf{ViT-FMT} backbone over tasks \textbf{T1} (yogaball), \textbf{T2} (pillow) and \textbf{T3} (bucket).
\textbf{Trans.} is translation-only correction without admittance, \textbf{SE(3)} is full SE(3) correction without admittance, and \textbf{Trans. + Admit.} is translation-only correction with admittance control (ours).
Lower contact drift and smoothness are better; higher contact force is better. Best per task/metric in bold.}
\label{tab:wtumi_se3}
\begin{tabular}{ll | c c c c @{}}
\toprule
\textbf{Task} & \textbf{Config}
& \makecell{\textbf{Cont. Drift}\\\textbf{(mm)}}
& \makecell{\textbf{Cont. Force}\\ \textbf{(N)}}
& \makecell{\textbf{Smooth.-Trans.}\\ \textbf{($\text{m/s}^2$)}}
& \makecell{\textbf{Smooth.-Rot.}\\ \textbf{($\text{rad/s}^2$)}} \\
\midrule
\multirow{3}{*}{\textbf{T1}}
& Trans.         & 18.12 & 4.77 & 3.14 & 20.29 \\
& SE(3) & \textbf{13.50} & 1.31 & 4.01 & 23.55 \\
& Trans. + Admit.  (Ours)         & 15.67 & \textbf{5.50} & \textbf{2.98} & \textbf{18.56} \\
\midrule
\multirow{3}{*}{\textbf{T2}}
& Trans.         & 21.04 & \textbf{0.52} & \textbf{1.87} & \textbf{12.63} \\
& SE(3) & 20.50 & 0.17 & 2.36 & 13.36 \\
& Trans. + Admit.  (Ours)          & \textbf{19.44} & 0.13 & 1.95 & 13.61 \\
\midrule
\multirow{3}{*}{\textbf{T3}}
& Trans.         & 25.00 & 0.95 & 1.85 & 14.05 \\
& SE(3) & 22.58 & 0.71 & \textbf{0.87} & \textbf{4.99} \\
& Trans. + Admit.  (Ours)          & \textbf{22.08} & \textbf{1.61} & 1.29 & 10.38 \\
\bottomrule
\end{tabular}
\end{table}

In most cases, our admittance controller correction approach outperforms the learned full SE(3) correction, particularly in \textbf{T1}. Learning full SE(3) pose correction provides little benefit over our admittance controller's rotational correction.

We attribute this to two factors. First, our framework is unable to align torsional phases between teleoperation and human data, since we only measure normal force.
 Second, the tasks tested in WT-UMI are dominated by normal force contacts instead of fine-orientation-based manipulation. Extending beyond the scope of this chapter, learning a full SE(3) correction may be useful for tasks where an accurate predicted orientation is important, such as cloth folding or tool manipulation.

\subsection{Force Supervision and Predicted Force Targets}
\label{sec:wtumi_results_force_ablation}
To isolate the contributions of force supervision and admittance control, we compare four ViT-FMT variants: (i) neither force loss ($\lambda_F=0$) nor admittance; (ii) no force loss, with admittance tracking a fixed $5$\,N force; (iii) force supervision, with admittance tracking a fixed $5$\,N force; and (iv) our full method.

\begin{table}[t]
\centering
\footnotesize
\setlength{\tabcolsep}{3pt}
\renewcommand{\arraystretch}{0.95}
\caption{Ablation of force prediction and admittance controller across tasks \textbf{T1}--\textbf{T3}. Each row adds one component over the row above. \textbf{Force Pred}: whether the policy is trained with force supervision loss. \textbf{Admit.}: \emph{Static} means the admittance controller tracks a constant $5$\,N input, \emph{Predicted} means the admittance controller tracks the force head prediction. The \emph{Predicted} rows are taken from the ViT-FMT w/ results in \ref{tab:wtumi_backbones}. ``--'' means the trial has no successful samples. Best value of each metric is bolded.}
\label{tab:wtumi_force_ablation}
\begin{tabular}{l cc | c c c c @{} }
\toprule
\textbf{Task} & \multicolumn{2}{c|}{\textbf{Config}}
& \textbf{Cont. Drift} & \textbf{Cont. Force} & \textbf{Smooth.-Trans.} & \textbf{Smooth.-Rot.} \\
& \textbf{Force Pred} & \textbf{Admit.}
& \textbf{(mm)} & \textbf{(N)} & \textbf{($\text{m/s}^2$)} & \textbf{($\text{rad/s}^2$)} \\
\midrule
\multirow{4}{*}{\textbf{T1}}
& No  & No      & -- & -- & -- & -- \\
& No  & Static  & 29.92 & -- & 0.87 & 5.80 \\
& Yes & Static  & 16.93 & 0.99 & 2.44 & 16.94 \\
& Yes & Predicted & \textbf{15.67} & \textbf{5.50} & 2.98 & 18.56 \\
\midrule
\multirow{4}{*}{\textbf{T2}}
& No  & No      & -- & -- & -- & -- \\
& No  & Static  & 27.79 & 0.53 & 4.59 & 26.72 \\
& Yes & Static  & 24.62 & \textbf{0.55} & 3.91 & 23.47 \\
& Yes & Predicted & \textbf{19.44} & 0.13 & \textbf{1.95} & \textbf{13.61} \\
\midrule
\multirow{4}{*}{\textbf{T3}}
& No  & No      & -- & -- & -- & -- \\
& No  & Static  & 25.45 & 0.77 & 1.76 & 13.12 \\
& Yes & Static  & 23.23 & 1.38 & 2.00 & 14.31 \\
& Yes & Predicted & \textbf{22.08} & \textbf{1.61} & \textbf{1.29} & \textbf{10.38} \\
\bottomrule
\end{tabular}
\end{table}

Across \textbf{T1}--\textbf{T3}, tracking the policy-predicted force (iv) achieves better force regulation and smaller contact drift than tracking a constant force (ii, iii). Removing admittance (i) often results in task failure.
Inaccurate force predictions can cause contact loss or excessive contact force, thereby degrading contact quality and preventing task completion.
Our admittance controller helps improve the contact quality: when a contact is lost under correct force prediction, the admittance controller shifts the target pose along the sensor normal direction to re-establish contact, until a contact is detected or the shift reaches its maximum allowable offset.

\subsection{Effect of the Teleoperation-to-Human Training Data Ratio}
\label{sec:wtumi_results_ratio}
To study the effect of training data composition, we vary the teleoperation-to-human training data ratio while holding the total data length fixed. We consider three settings: teleoperation only ($10-0$), an equal mix ($5-5$), and $90\%$ human data ($1-9$). We evaluate contact drift, contact force, translational smoothness, and rotational smoothness on tasks \textbf{T1}--\textbf{T3}.

\begin{table}[t]
\centering
\footnotesize
\setlength{\tabcolsep}{3pt}
\renewcommand{\arraystretch}{0.95}
\caption{Effect of the teleoperation-to-human training data ratio on tasks \textbf{T1}--\textbf{T3}, with total data length held fixed. \textit{Tel:Hum} denotes the teleoperation-to-human ratio: \textbf{10-0}, \textbf{5-5}, and \textbf{1-9} correspond to 100\%:0\%, 50\%:50\%, and 10\%:90\%, respectively.
Lower drift, smaller smoothness values, and higher contact force indicated better performance. The best value for each metric is bolded.}
\label{tab:wtumi_data_ratio}
\begin{tabular}{ll | c c c c @{}}
\toprule
\textbf{Task} & \textbf{Tel:Hum}
& \textbf{Cont. Drift (mm)}
& \textbf{Cont. Force (N)}
& \textbf{Smooth.-Trans. ($\text{m/s}^2$)}
& \textbf{Smooth.-Rot. ($\text{rad/s}^2$)} \\
\midrule
\multirow{3}{*}{\textbf{T1}}
& 10-0 & 10.78 & 1.21 & \textbf{3.62} & \textbf{22.25} \\
& 5-5  & \textbf{10.37} & \textbf{2.19} & 3.72 & 22.84 \\
& 1-9  & -- & -- & -- & -- \\
\midrule
\multirow{3}{*}{\textbf{T2}}
& 10-0 & 23.61 & 0.54 & 5.08 & 27.25 \\
& 5-5  & 22.43 & 0.63 & 4.22 & 23.66 \\
& 1-9  & \textbf{20.19} & \textbf{0.68} & \textbf{2.52} & \textbf{17.50} \\
\midrule
\multirow{3}{*}{\textbf{T3}}
& 10-0 & \textbf{20.49} & 1.59 & 2.77 & 19.28 \\
& 5-5  & 24.21 & \textbf{2.50} & 3.07 & 20.09 \\
& 1-9  & 31.85 & 1.21 & \textbf{1.10} & \textbf{6.92} \\
\bottomrule
\end{tabular}
\end{table}

Across data ratios, policies trained with corrected human data perform similarly to those trained with teleoperation data alone (\ref{tab:wtumi_data_ratio}), indicating that the target-pose correction effectively converts human demonstrations into usable robot data. The 5-5 split gives the most balanced results, improving contact force on all three tasks without the larger drift or task failure observed at 1-9. The 1-9 split fails in \textbf{T1}, showing that a minimum amount of teleoperation data is still required.

\subsection{Human--Humanoid Collaborative Manipulation}
\label{sec:wtumi_results_collab}
\begin{figure}[t]
\centering
\includegraphics[width=0.95\textwidth]{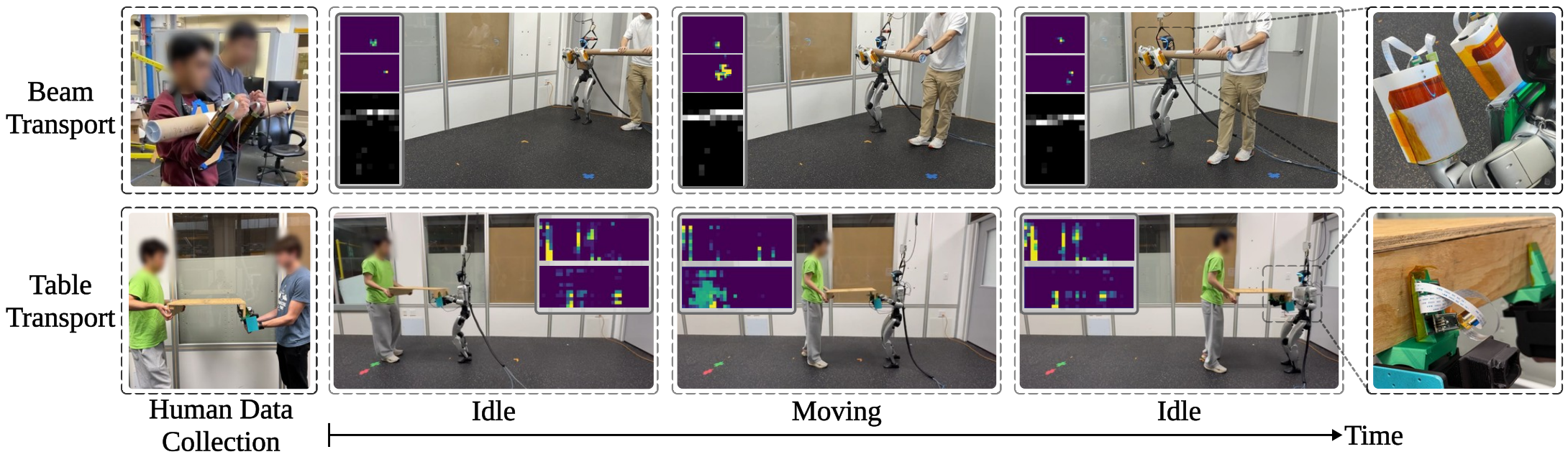}
\caption{Human--humanoid collaborative manipulation tasks. Top images show the beam transport task, and bottom images show the table transport task. For each task, a depiction of the human-human data collection is included.}
\label{fig:wtumi_collaborative}
\vspace{-0.15in}
\end{figure}

We further evaluate WT-UMI and our planning framework on two human--humanoid collaborative manipulation tasks: beam transport (\textbf{T4}) and table transport (\textbf{T5}), shown in \ref{fig:wtumi_collaborative}. Unlike the single-agent tasks (\textbf{T1}--\textbf{T3}), these tasks require the humanoid to maintain distributed contact while reading its partner's intent from tactile feedback: changes in force distribution and contact activation across the sensing surfaces signal whether the human is pushing or pulling, and the humanoid adapts its motion to accomplish coordinated transport.

\paragraph{Task Setup.}
Both tasks require continuous contact regulation and shared-load coordination. In the beam transport task, the humanoid stabilizes a cardboard beam (length $1.33$ m, diameter $0.09$ m, weight $8.72$ N) using distributed contact across the forearms and chest, while a human partner applies pushing or pulling forces to guide the motion. The robot's forearms support most of the load, while its chest provides additional stabilization and prevents slip. The tactile signals in these regions contain visible patterns for intent inference   (\ref{fig:wtumi_collaborative}): during forward motion, left forearm activation typically increases while right forearm activation decreases, and the chest activation distribution shifts with the interaction-force direction.

In the table-transport task, the humanoid supports a plywood table with both palms while adapting to translational motions initiated by a human. The table measures $609\,\mathrm{mm}\times609\,\mathrm{mm}\times125\,\mathrm{mm}$, weighting $39\,\mathrm{N}$. We use FlexiTac sensors mounted on a custom-designed gripper, as in \ref{fig:wtumi_teaser}.

\paragraph{Data Collection.}
For both tasks, demonstrations are collected through human-human collaborative transport, where one participant wears WT-UMI and physically mimics the robot role, while the other acts as the human partner. These demonstrations capture natural shared-load coordination behaviors and tactile contact patterns without requiring robot teleoperation.  For the beam transport task, we further collect an additional robot-in-the-loop dataset to cover the embodiment mismatch between human and robot. In this data collection, WT-UMI is mounted directly on the humanoid robot. Overall, we collected $56$ beam transport trajectories and $45$ table transport trajectories containing pushing, pulling, and idle interaction behaviors, with each trajectory lasting approximately $15$ seconds on average. The action chunk in transport tasks consists of base velocities in SE(2): $\mathbf{a} = [\mathbf{v}_{t+1},\dots,\mathbf{v}_{t+H_a}] \in \mathbb{R}^{H_a \times 3}$.

\paragraph{Deployment and Evaluation.}
During deployment, the humanoid supports forward and backward locomotion in the beam and table transport tasks and returns to idle when the contact distribution relaxes to the nominal stationary-holding pattern. For \textbf{T4} and \textbf{T5}, we evaluate $22$ and $20$ trials respectively, each involving one unique human partner and one stand--walk--stand cycle, with the human partner being consistent across all trials for each task. We define the success rate as the fraction of time that the policy-selected locomotion mode (walk or stand) matches the human partner's intent. The \textbf{T4} policy response latency is approximately $0.789$s and the \textbf{T5} policy response latency is approximately $0.865s$; therefore, we use a latency-aware evaluation in which, during the latency window around each ground-truth intent change, a prediction is considered correct if it matches either the ground-truth intent immediately before the change or the ground-truth intent immediately after the change. Under this evaluation, \textbf{T4} achieves a success rate of $78.91 \pm 14.51\%$, while \textbf{T5} obtains $87.80 \pm 11.46\%$. For \textbf{T4} forearm contacts, the force predictor achieves an RMSE of $0.23$ N, providing the force reference for contact regulation.

\section{Conclusion}
\label{sec:wtumi_conclusion}
In this chapter, we presented WT-UMI, a large-area tactile-based humanoid manipulation system built around a force-conditioned target-pose correction module and a force-supervised planner that leverages the natural tactile forces in human demonstrations. The correction module converts human hand poses into contact-aware robot target poses by applying learned corrections from teleoperated robot data, yielding action labels for planner training. The force-supervised planner uses a cross-attention force head to predict a contact-force trajectory via direct regression, and at deployment a tactile-based admittance controller consumes the predicted force as the normal-force reference to maintain stable contact. Together, the target-pose correction, force-supervised planner, human--robot co-training data, and admittance controller support large-area contact-rich manipulation across four policy backbones, including manipulation of deformable and large rigid objects and collaborative transport.

The system has three main sensing limitations that point to future work. First, tactile coverage is constrained by the available sensor configurations, so only the palms, forearms, and chest are instrumented; extending coverage to dexterous hands, legs, and back would broaden the set of tactile-driven tasks. Second, the policy does not yet consume RGB vision, because third-person views introduce substantial differences between human and humanoid observations and a consistently unoccluded camera angle is difficult to maintain during transport; fusing tactile sensing with vision is an important next step toward anticipating future contact and re-establishing lost contact. Third, the force head predicts only a scalar normal force. Future models could instead predict multi-axis contact wrenches, distributed force maps, or future tactile images, which would support finer contact centering, slip anticipation, and whole-body contact redistribution in more complex contact configurations.

Different from Thrust 1 (LTL and STL-MPC), Thrust 2 (WT-UMI) in this section leverages human demonstrations as a source of contact-rich manipulation. This shift is motivated by the complexity of whole-body physical interaction, where specifications and models alone cannot fully capture the richness of contact. Together, Thrust 1 and Thrust 2 establish two foundational capabilities of the dissertation: robust locomotion and force-aware whole-body manipulation. Thrust 3 builds on these capabilities by integrating locomotion and manipulation toward autonomous humanoid loco-manipulation.

%% file: chapters/PPF.tex
\chapter{PPF: Reinforcement Learning Fine-Tuning for Robust Locomotion}
\label{chapter:PPF}

\noindent\textbf{Dissertation Roadmap and Contribution of This Chapter.}
\ref{chapter:PPF} and \ref{chapter:DPFT} together constitute the third thrust of this dissertation: \emph{improving pre-trained humanoid capabilities through reinforcement learning fine-tuning}. Although Behavior Cloning (BC) is effective in learning policies from demonstrations, its performance is inherently constrained by the capabilities and coverage of the demonstrator. Consequently, the learned policy may deteriorate substantially when deployed in scenarios that deviate even modestly from the training distribution. To address this limitation, we adopt reinforcement learning fine-tuning. Rather than expanding the dataset to cover every possible scenario, reinforcement learning fine-tuning adapts to unseen scenarios while preserving the prior behaviors acquired during pre-training.

\ref{chapter:PPF} demonstrates how reinforcement learning fine-tuning can improve pre-trained locomotion policies. In particular, it addresses the challenge of catastrophic forgetting, in which fine-tuning degrades the natural gait acquired during pre-training. To mitigate this issue, we introduce Model-Assumption-based Regularization (MAR), which dynamically balances reinforcement learning exploration with imitation of expert demonstrations. We validate the improved locomotion robustness on slippery and uneven terrain.

This ``pre-training then fine-tuning" paradigm developed in this chapter lays the foundation for \ref{chapter:DPFT}, which extends this idea to autonomous loco-manipulation. Together, these two chapters show that reinforcement learning fine-tuning can improve pre-trained humanoid capabilities beyond the limitations of their initial demonstrations.
%half of the third thrust by treating a model-based locomotion controller as a demonstrator for policy pre-training. PPF distills controller-generated trajectories into a neural locomotion policy and then fine-tunes the policy with reinforcement learning on terrain conditions that violate the controller's assumptions. 

\begin{figure}[!htbp]
  \centering
  \includegraphics[width=\linewidth]{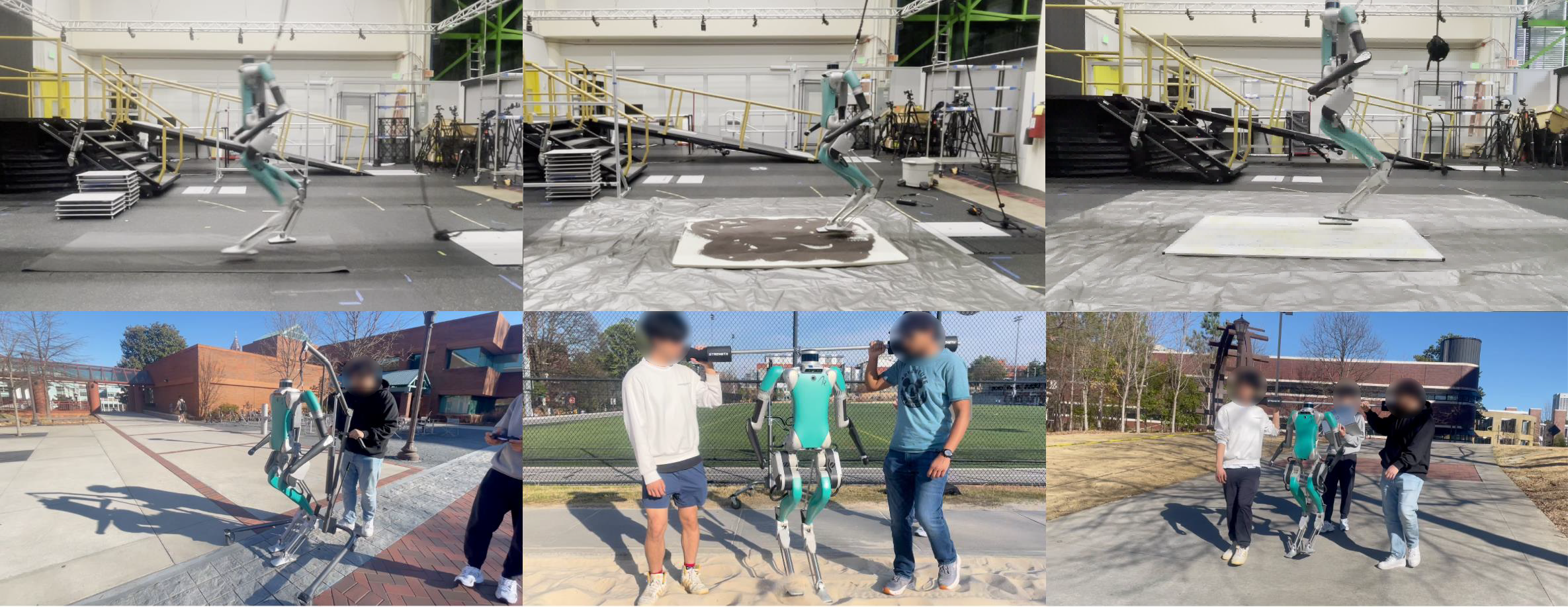}
  \caption{Our Pre-training and Preservative Fine-tuning (PPF) framework
  achieves a forward velocity of $1.5$ m/s while successfully traversing a
  whiteboard covered with poppy seeds or olive oil, as well as diverse outdoor
  terrains, including hills, uneven surfaces, and sand.}
  \label{fig:ppf_overview}
\end{figure}

\section{Introduction}
The first thrust in \ref{chapter:STL-MPC} encodes locomotion robustness as a formal specification and enhances robustness through model-based optimization, synthesizing motions with interpretable stability guarantees. This model-based capability, however, remains bounded by reduced-order models such as the Linear Inverted Pendulum Model (LIPM)~\cite{Kajita2001,pratt2006capture}. These models capture essential locomotion dynamics but simplify whole-body motion, terrain geometry, and contact dynamics by assuming constrained center-of-mass motion, flat terrain, and non-slippery contact. Consequently, Model Predictive Control (MPC) methods~\cite{scianca2020mpc,gibson2022terrain} based on these models can struggle on slippery, deformable, or uneven terrain that violates these assumptions. Higher-fidelity models can reduce modeling error~\cite{Dai_centroidal_kinematics,khazoom2024tailoring}, but they often increase computational cost or require predefined contact sequences. This chapter therefore builds on model-based control by using it as a source of behavioral priors rather than as the final controller.

In recent years, learning-based approaches have emerged as an effective alternative for bipedal locomotion control. Reinforcement learning (RL), in particular, has enabled bipedal and quadrupedal robots to achieve agile and robust locomotion on challenging terrain \cite{transformer_digit_RL, Ma_RL_MPC_quad, Rudin2022, Kumar2021}. By optimizing control policies directly through trial and error, RL can discover behaviors, such as heel-to-toe contact and slip recovery, that are difficult to capture with simplified models. Successful examples include blind bipedal locomotion \cite{all_gait_cassie, siekmann2021blind} and quadrupeds traversing outdoor environments \cite{Miki2022}. However, these methods typically require extensive reward engineering, and their low data efficiency leads to substantial computational cost and long training times. The learned policies may also be brittle outside the training distribution and lack safety guarantees. In practice, deploying an RL controller on humanoid hardware requires careful domain randomization and sim-to-real transfer techniques to bridge the gap between simulation and real hardware \cite{Rudin2022}. Moreover, without prior knowledge, learning a walking gait from scratch can result in unnatural or suboptimal motions unless guided by reference motions or expert policies.

Hybrid model-based and learning-based methods offer a practical way to combine structured locomotion priors with data-driven adaptation. Some approaches incorporate a model-based controller during inference to supervise or guide an RL policy \cite{Xie2021, Da2020, yu2024learning, Lee_2024}, but this strategy can retain the runtime cost and tuning complexity of the model-based component. An alternative is to use the Model-Based Controller (MBC) only during training: the controller first provides demonstrations for BC, after which RL further improves the learned policy beyond the controller's original capabilities. Prior work has shown the effectiveness of this two-stage strategy in quadrupedal locomotion \cite{IFM_quad}. Motivated by these results, this chapter uses an MBC as a source of stable locomotion demonstrations and then applies reinforcement learning fine-tuning to improve speed, terrain robustness, and disturbance rejection without requiring the controller during deployment.

\begin{figure}[t]
\centering
\begin{tabular}{cc}
    \includegraphics[width=0.3\linewidth]{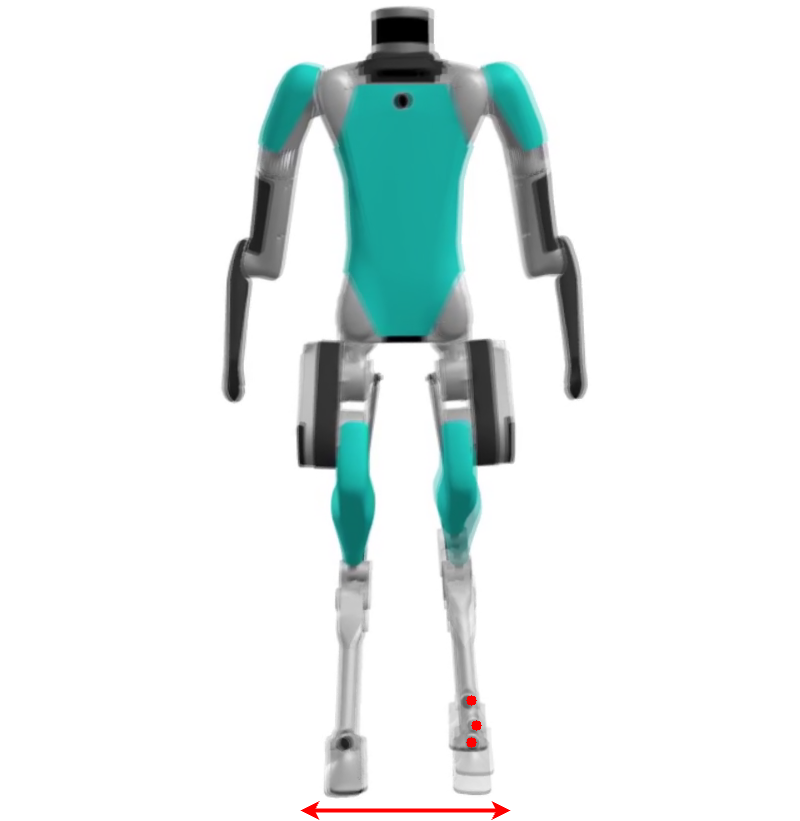} &
    \includegraphics[width=0.3\linewidth]{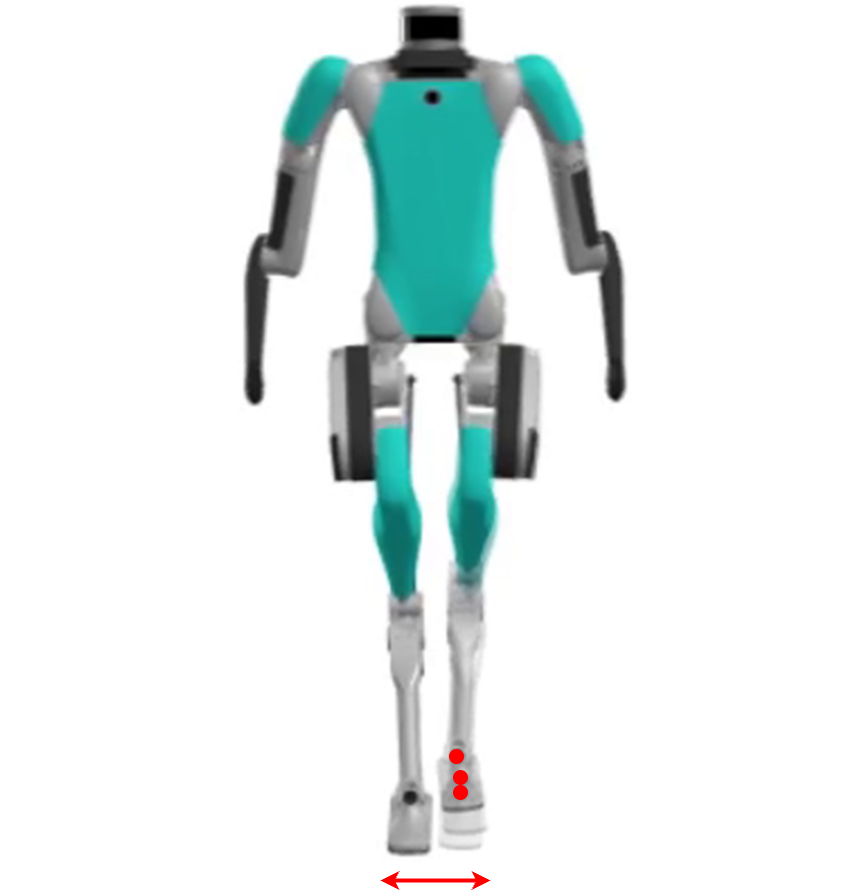} \\
    \scriptsize (a) MBC motion & \scriptsize (b) IFM motion \\
\end{tabular}
\caption{Catastrophic forgetting of motion style. Unlike the MBC (a), imitation-and-fine-tuning
without regularization (IFM, b) drives the policy toward a narrower foot placement and an
altered gait that optimizes lateral tracking in simulation, sacrificing the stable,
periodic gait learned from the controller and raising the risk of self-collision and
tripping on hardware.}
\label{fig:forgetting_example}
\vspace{-0.05in}
\end{figure}

%Motivated by this idea, we apply a two-stage learning framework to humanoid locomotion, building on our model-based humanoid locomotion controller.
% Humanoids, however, introduce additional challenges compared with quadrupeds. They are inherently more unstable because of their smaller support polygon and higher center of mass, and subtle modeling errors or policy deviations can lead to falls. 
A major challenge in fine-tuning a pre-trained policy is \emph{catastrophic forgetting} of the original walking behavior. The policy may overfit to the new task dynamics and lose useful gait characteristics learned from the model-based controller, such as symmetric foot placement and periodic gait. As an example, \ref{fig:forgetting_example} shows that the fine-tuned policy tends to swing its feet inward and adopt a narrower stance than the MBC, an adaptation that improves lateral velocity tracking in the training simulator but increases the chance of self-collision and tripping once the policy is transferred to a different simulator or to hardware. This observation aligns with general issues in continual learning \cite{Wang2024CL} and motivates the need to preserve model-based gait priors during RL fine-tuning. To address this issue, we develop Pre-training and Preservative Fine-tuning (PPF), a framework that incorporates Model-Assumption-based Regularization (MAR) into RL fine-tuning to preserve pre-trained gait styles and achieve robust humanoid locomotion.

\section{Methods}
The PPF framework trains a humanoid locomotion policy that imitates a Model-Based Controller (MBC) and then enhances its robustness through reinforcement learning without forgetting the gait and motion style of the original controller. For the MBC, we adopt our humanoid locomotion controller in Shamsah et al.~\cite{shamsah2023TRO}, which consists of a foot placement controller based on the Angular-Momentum Linear Inverted Pendulum (ALIP) model~\cite{gong2022jdsmc} and a passivity-based whole-body inverse dynamics controller~\cite{PassivityControl}.

\subsection{Model-Based Controller}
The foot-placement controller models the bipedal robot as the ALIP~\cite{gong2022jdsmc}, a Center of Mass (CoM) connected to the contact point by a massless telescopic leg. Assuming a constant CoM height $z$, the ALIP dynamics are
\begin{equation}\label{eq:alip}
\dot{\mathbf{x}}_c = \frac{\mathbf{L}}{mz}, \qquad \dot{\mathbf{L}} = mg\,\mathbf{x}_c + \mathbf{u}_a,
\end{equation}
where $\mathbf{x}_c$ is the horizontal CoM position relative to the contact point, $\mathbf{L}$ is the angular momentum about the contact point, $\mathbf{u}_a$ is the ankle torque, $m$ is the robot mass, and $g$ is gravitational acceleration. The model assumes a constant height and no vertical motion,
\begin{equation}\label{eq:model_assumption}
z = \bar z, \qquad \dot z = 0, \qquad \ddot z = 0,
\end{equation}
with $\bar z = 1.01$~m for the Digit humanoid robot~\cite{agility}. Under~\eqref{eq:model_assumption}, the desired foot placement relative to the CoM follows from a one-step-ahead prediction,
\begin{equation}\label{eq:foot_placement}
\mathbf{x}^{\text{des}} = \frac{\mathbf{L}\cosh(\omega T)/mz - \mathbf{v}_c^{\text{des}}}{\omega \sinh(\omega T)},
\end{equation}
where $\mathbf{x}^{\text{des}}$ is the calculated horizontal foot placement, $T$ is a pre-defined step duration, $\omega = \sqrt{g/\bar z}$ is the natural frequency, and $\mathbf{v}_c^{\text{des}}$ is the desired horizontal CoM velocity. The foot-placement controller computes $\mathbf{x}^{\text{des}}$ as the input to a passivity-based whole-body inverse-dynamics controller~\cite{shamsah2023TRO, PassivityControl}, which solves for the joint torques needed to produce a smooth, symmetric, and periodic walking gait.

\subsection{Pre-training and Preservative Fine-tuning}
PPF distills this MBC into a learnable policy and then improves it with reinforcement learning while preserving the MBC's gait. It comprises three components: imitation pre-training, RL fine-tuning, and Model-Assumption-based Regularization (MAR) during RL fine-tuning.

\begin{itemize}
\item \textbf{Pre-training via imitation of the MBC:} In the first stage, we distill the expert
Model-Based Controller (MBC) into a learnable neural network policy using Dataset Aggregation (DAgger) \cite{DAgger}, a BC
approach. Demonstration trajectories
$(s, a^E)$ are collected from the MBC across nominal scenarios, where $s$ is the robot
state and $a^E$ is the corresponding expert action. The policy parameters $\theta$ are
trained by minimizing the squared Euclidean imitation loss:
\[
L_{\text{DAgger}}(\theta) = \mathbb{E}_{(s,a^E)\sim \mathcal{D}}\left[\|a^E - \mu_\theta(s)\|^2\right],
\]
where $\mathcal{D}$ is the aggregated dataset and $\mu_\theta(s)$ is the policy output, which is the mean of the Gaussian policy. During this
stage, the network learns to reproduce the stable gait of the MBC. This procedure yields initial policy parameters $\theta_0$ and the corresponding policy $\pi_{\theta_0}$, which closely matches the MBC’s behavior while remaining learnable for
subsequent fine-tuning.

\item \textbf{Reinforcement learning fine-tuning:} After pre-training the policy, we further
improve its robustness through reinforcement learning using Proximal Policy Optimization
(PPO) \cite{PPO}. Training proceeds with a task reward that balances commanded velocity tracking, posture regularization, and stability, while penalizing
undesirable behaviors such as high-frequency steps or falling.
To facilitate learning, we employ a \emph{curriculum} both in velocity command and in terrain difficulty (flat ground, uphill, downhill, uneven terrain).
In addition, we randomize dynamics parameters and inject observation noise during fine-tuning
to improve sim-to-real transfer. 
Nevertheless, even with carefully designed rewards and randomization, fine-tuning may suffer from catastrophic forgetting of the pre-trained gait. To mitigate this, PPF augments the RL objective with MAR.

\begin{figure}[t]
\centering
\includegraphics[width=0.8\columnwidth]{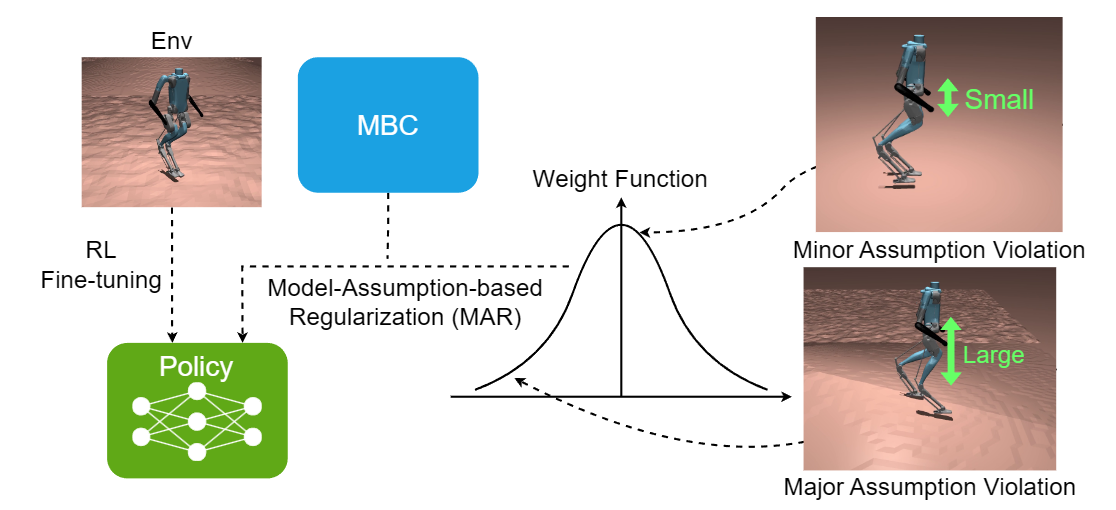}
\caption{Overview of Model-Assumption-based Regularization (MAR). Our framework automatically adjusts the supervised loss regularization based on the assumption violation of the MBC.}
\label{fig:ppf_diagram}
\vspace{-0.05in}
\end{figure}

\item \textbf{Model-Assumption-based Regularization (MAR):} The key contribution of PPF is MAR, which prevents catastrophic forgetting during RL fine-tuning by
selectively regularizing the policy toward the model-based controller’s actions. We augment the PPO loss with a per-sample imitation term whose weight depends on the state:
\begin{equation}\label{eq:ppf_loss}
L_{\text{PPF}}(\theta,\sigma) = L_{\text{PPO}}(\theta,\sigma) + \mathbb{E}_{(s,a^E)\sim\mathcal{D}}\!\left[\,w(s)\,\|a^E-\mu_\theta(s)\|^2\,\right],
\end{equation}
where $L_{\text{PPF}}$ is the regularized fine-tuning objective, $L_{\text{PPO}}$ is the PPO loss, $w(s)$ is a state-dependent weight applied per sample, and the policy is modeled as a Gaussian distribution with mean $\mu_\theta(s)$ and learnable standard deviation $\sigma$. Specifically, we define
\begin{equation}\label{eq:mar_weight}
w(s) = w_0 \exp\!\left(-\frac{\dot z(s)^2}{\delta}\right),
\end{equation}
where $\dot z(s)$ is the vertical CoM velocity from~\eqref{eq:model_assumption}, and $w_0 = 5$ and $\delta = 0.0159$ set the strength and falloff of the regularization. MAR therefore uses violation of the vertical-motion assumption as a state-dependent reliability signal for expert supervision. As shown in \ref{fig:ppf_diagram}, when the ALIP assumption ($\dot z$ close to zero) holds, MAR strongly enforces similarity to the MBC, preserving its stable periodic gait.
When the assumption is violated (e.g., on slopes or uneven ground), $w(s)$ decays toward
zero, allowing the policy to deviate and adapt through RL. This adaptive regularization
balances gait style preservation with the flexibility to learn new behaviors.

\end{itemize}

The policy $\pi_{\theta_0}$ first obtains a stable gait on easy terrain and is then fine-tuned with PPF on more challenging, randomized terrain to maximize reward while the regularized objective~\eqref{eq:ppf_loss} preserves the MBC gait through MAR. The result is a single policy that retains the MBC's stabilizing periodic gait when the model assumption is reliable, yet remains free to deviate and exploit the RL objective when the model assumption breaks.

\subsection{Implementation Details}
\noindent\textbf{Observation and action space.} The policy observations comprise the root linear and angular velocities, projected gravity, the user command, lower-body joint positions and velocities, the gait phase and domain indicator, the previous action, and the history of lower-body joint positions and velocities. The action space consists of target joint positions and velocities for the lower body.

\noindent\textbf{Reward.} The reward tracks the commanded linear and angular velocities and regularizes excessive motion by penalizing joint torques and base angular velocity, as listed in \ref{tab:ppf_reward}.

\begin{table}[h]
\centering
\caption{PPF reward terms and weights. $\mathbf{v}_{xy}$ and $\omega_z$ are the measured planar linear and yaw velocities, $\mathbf{v}_{xy}^{\text{cmd}}$ and $\omega_z^{\text{cmd}}$ are the corresponding commands, $\boldsymbol{\tau}$ denotes the joint torques, $\boldsymbol{\omega}_{xy}$ denotes the base roll/pitch rates, and $\sigma_{xy},\sigma_{\omega}$ are the tracking temperatures.}
\label{tab:ppf_reward}
\small
\begin{tabular}{l l r}
\hline
\textbf{Term} & \textbf{Expression} & \textbf{Weight} \\
\hline
Linear velocity tracking & $\exp(-\|\mathbf{v}_{xy} - \mathbf{v}_{xy}^{\text{cmd}}\|_2^2 / \sigma_{xy})$ & $1.2$ \\
Angular velocity tracking & $\exp(-(\omega_z - \omega_z^{\text{cmd}})^2 / \sigma_{\omega})$ & $1.1$ \\
Torque penalty & $\|\boldsymbol{\tau}\|_2^2$ & $-4\times10^{-6}$ \\
Base motion penalty & $\|\boldsymbol{\omega}_{xy}\|_2^2$ & $-0.6$ \\
\hline
\end{tabular}
\vspace{-0.15in}
\end{table}

\noindent\textbf{Training.} Policies are trained in MuJoCo~\cite{todorov2012mujoco} on a single RTX 4090 GPU. The DAgger pre-training stage runs for $380$ iterations ($9.12$M samples, roughly $5.5$~hours), and the RL fine-tuning stage runs for $10{,}000$ iterations (roughly $400$M samples, $31.5$~hours).

\section{Results}
We evaluate the PPF-trained policy both in a high-fidelity simulation and on a real humanoid robot (Agility Robotics’ Digit~\cite{agility}) across a variety of tasks. The results demonstrate that PPF achieves stronger robustness than the original model-based controller and other baseline learning approaches, validating the effectiveness of the RL fine-tuning with regularization.

\subsection{Experimental Setup}
We compare PPF against four baselines that span model-based control, pure reinforcement learning, and intermediate regularization choices. All learned policies use the same Multilayer Perceptron (MLP) with hidden layers of size $512$, $256$, and $64$.
\begin{itemize}
\item \textbf{MBC:} the model-based controller~\cite{shamsah2023TRO} serves as the expert that teaches the pre-training policies.
\item \textbf{IFM~\cite{IFM_quad}:} imitation pre-training followed by RL fine-tuning with no regularization toward the MBC, i.e., the objective~\eqref{eq:ppf_loss} with $w(s)=0$.
\item \textbf{FullReg:} the same pre-training and fine-tuning as PPF, but with a constant regularization weight ($w(s)=w_0$ for all states), so the policy is pulled toward the MBC even where the model assumptions fail.
\item \textbf{PureRL:} a policy trained from scratch with RL alone, without imitation pre-training or MBC regularization.
\end{itemize}
PPF is the full framework, combining imitation pre-training, RL fine-tuning, and MAR. Policies are trained and tested in MuJoCo~\cite{todorov2012mujoco} and additionally evaluated in AR-Sim, a higher-fidelity simulator from the robot manufacturer whose dynamics closely match the physical robot, to assess sim-to-sim transfer. Deployment to AR-Sim and to hardware is zero-shot, without additional tuning or training.

\subsection{Simulation Experiments}
\begin{figure}[t]
\centering
\includegraphics[width=0.6\linewidth]{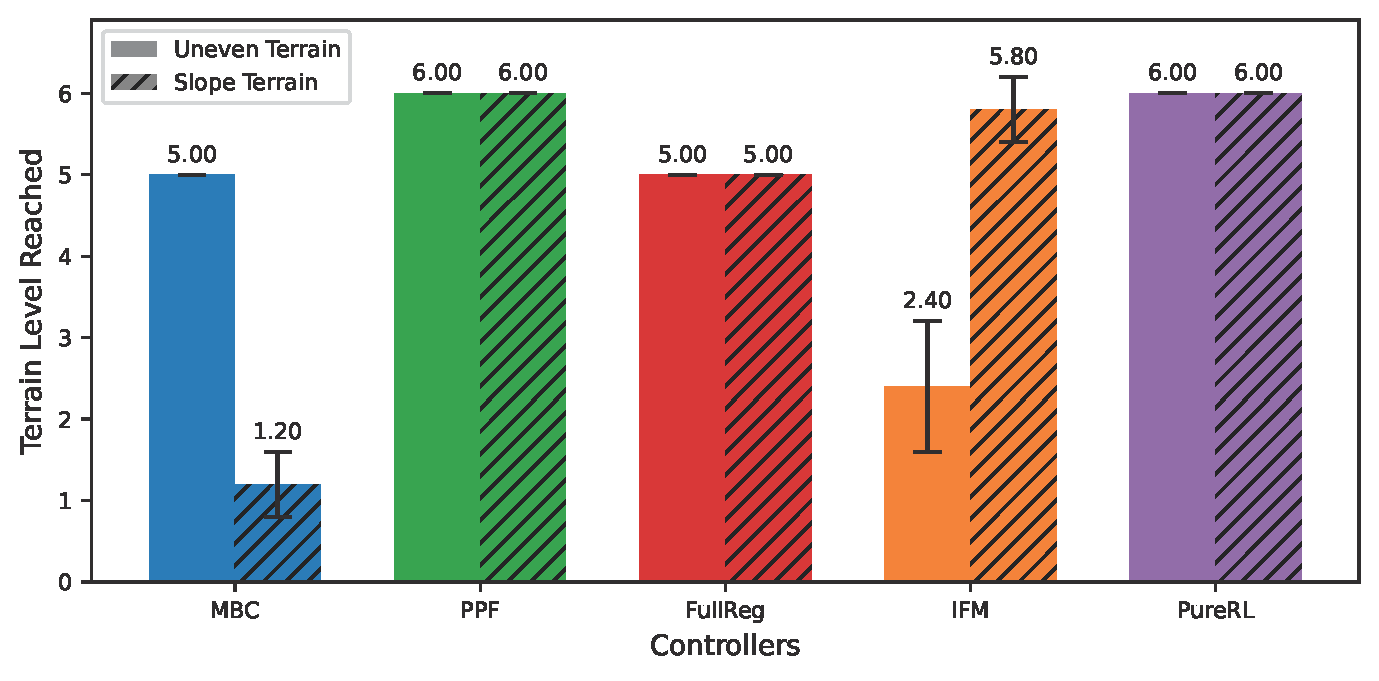}
\caption{Terrain level reached in the MuJoCo robustness test (higher is better),
averaged over five seeds for a forward command of $0.6$~m/s on courses of
increasingly difficult slope and uneven segments. PPF clears the final level on
both courses; IFM nears the final level on slopes but trips on uneven terrain;
FullReg times out before the goal under degraded tracking; PureRL reaches the goal
but overfits to the training simulator, which later degrades its transfer.}
\label{fig:ppf_mujoco_terrain}
\vspace{-0.05in}
\end{figure}
In simulation, we evaluate robustness on courses of increasingly difficult terrain segments, spanning sloped and uneven terrains. The robot is commanded forward at $0.6$~m/s while adjusting its heading to stay on course, and each policy is run with five random seeds until it falls, reaches the $130$~s time limit, or clears the final segment. \ref{fig:ppf_mujoco_terrain} reports the terrain level reached by each policy. PPF reaches the final level on both the slope and uneven courses. MBC fails to cope with sloped terrain. FullReg remains robust on the easier segments, but its tracking degrades under model-assumption violations, so it times out before reaching the goal. IFM~\cite{IFM_quad} nears the final level on slopes, yet on uneven ground unexpected foot contacts cause disturbances that make it trip over its own foot. PureRL also reaches the goal within the time limit, but it overfits to the training simulator, which degrades its transfer to AR-Sim and hardware (\ref{tab:ppf_ar_sim_slope} and \ref{tab:ppf_hw}). These results show that PPF preserves the MBC gait well to traverse the full course, while MAR lets it deviate where the conservative gait would otherwise stall; the tracking-accuracy advantage that accompanies this robustness is quantified in the sim-to-sim transfer test below (\ref{tab:ppf_ar_sim_slope}).

\begin{table}[t]
\centering
\caption{AR-Sim sim-to-sim robustness. Success rate and linear-velocity tracking
error (as a percentage of commanded speed, lower is better) on uphill ramps of
increasing incline. PPF is the only learned policy that combines a $100\%$ success
rate with the lowest tracking error at every slope.}
\label{tab:ppf_ar_sim_slope}
\vspace{-0.5em}
\small
\begin{tabular}{l cc cc cc}
\hline
\multirow{2}{*}{\textbf{Method}}
 & \multicolumn{2}{c}{\textbf{$10^\circ$ Uphill}}
 & \multicolumn{2}{c}{\textbf{$12^\circ$ Uphill}}
 & \multicolumn{2}{c}{\textbf{$14^\circ$ Uphill}} \\
\cline{2-7}
 & Succ.\ (\%) & Err.\ (\%) & Succ.\ (\%) & Err.\ (\%) & Succ.\ (\%) & Err.\ (\%) \\
\hline
MBC          & 0   & 34.5 & 0   & 33.2 & 0   & 34.4 \\
PureRL       & 85  & 10.9 & 60  & 15.4 & 0   & 54.1 \\
IFM          & 100 & 15.5 & 100 & 24.9 & 100 & 30.7 \\
FullReg      & 100 & 12.1 & 100 & 13.1 & 100 & 28.7 \\
\textbf{PPF} & \textbf{100} & \textbf{8.4} & \textbf{100} & \textbf{9.1} & \textbf{100} & \textbf{18.9} \\
\hline
\end{tabular}
\vspace{-0.15in}
\end{table}

We also conduct a sim-to-sim transfer test to evaluate how well policies trained in MuJoCo perform in a higher-fidelity simulator (AR-Sim) provided by the robot manufacturer, assessing generalization to different dynamics without additional training. The humanoid is commanded to walk up $5$~m ramps of increasing incline ($10^\circ$, $12^\circ$, and $14^\circ$); a trial succeeds if the robot reaches the top without falling. \ref{tab:ppf_ar_sim_slope} reports the success rate and the linear-velocity tracking error as a percentage of commanded speed. PPF is the only policy to combine a $100$\% success rate at every incline with the lowest tracking error, and its error grows the least as the slope steepens, rising from $8.4$\% at $10^\circ$ to $18.9$\% at $14^\circ$. FullReg and IFM also reach the top at all angles but track less accurately ($28.7$\% and $30.7$\% at $14^\circ$), indicating that fully imitating the MBC degrades tracking once the model assumptions break. PureRL transfers poorly, dropping from $85$\% success at $10^\circ$ to $0$\% at $14^\circ$, consistent with overfitting to the training simulator. The model-based controller fails at every incline, reflecting its model limitations. These results show that imitating the MBC gait through pre-training and partial regularization improves robustness and generalization, while MAR preserves tracking accuracy under model-assumption violations.

\subsection{Hardware Experiments}

We deploy the PPF policy on the Digit humanoid robot to validate its performance in real-world scenarios. PPF enables the robot to walk at a peak speed of $1.5$~m/s on flat ground, which is three times faster than the maximum speed achievable by the original model-based controller (approximately $0.5$~m/s in our tests). This represents a substantial gain in agility obtained through RL fine-tuning.

We also test the policy's robustness on a challenging slippery surface: a flat board coated with poppy seeds or olive oil. The PPF policy achieves a $100$\% success rate (no falls in $5/5$ trials), adapting its gait on the fly to prevent slipping. In contrast, the model-based controller and the FullReg policy both fail on every attempt. In those experiments, as soon as the robot’s leading foot hits the slippery board, it slides, and the robot cannot recover. We observe that PPF can quickly adjust its center of mass and foot placement when a slip occurs, effectively responding with a corrective step or reducing push-off force, whereas the MBC and FullReg continue with gaits that assume no slip, resulting in falls. This demonstrates how fine-tuning with RL and regularizing via MAR enables the policy to cope with low-friction contact dynamics that are not explicitly handled by the MBC.

\begin{table}[t]
\centering
\caption{Hardware experiments on the Digit humanoid. Indoor: maximum forward
velocity on flat ground and success rate on the step-and-slip board. Outdoor:
traversal of previously unseen sloped, uneven, and sandy terrain ($\checkmark$
success, $\times$ failure, N/A not attempted for safety). PPF is the only policy
that reaches $1.5$~m/s and clears every terrain.}
\label{tab:ppf_hw}
\vspace{-0.5em}
\small
\begin{tabular}{l cc ccc}
\hline
\multirow{2}{*}{\textbf{Method}}
 & \multicolumn{2}{c}{\textbf{Indoor}}
 & \multicolumn{3}{c}{\textbf{Outdoor}} \\
\cline{2-6}
 & Max.\ Vel.\ (m/s) & Step \& Slip (\%) & Slope & Uneven & Sand \\
\hline
MBC          & 0.5 & 0   & $\times$ & $\times$ & N/A \\
PureRL       & N/A & 0   & N/A & N/A & N/A \\
IFM          & $>$1.0 (unsafe) & 40  & N/A & N/A & N/A \\
FullReg      & 1.1 & 0   & \checkmark & \checkmark & \checkmark \\
\textbf{PPF} & \textbf{1.5} & \textbf{100} & \checkmark & \checkmark & \checkmark \\
\hline
\end{tabular}
\vspace{-0.15in}
\end{table}

Finally, we conduct outdoor field tests in which Digit walks over unstructured terrain that is \textit{not seen during training}, including sloped, uneven, and sandy terrain, as shown in~\ref{fig:ppf_overview}. We compare PPF against the MBC and FullReg; IFM and PureRL are excluded because of their unsafe behavior in the indoor tests. Both PPF and FullReg successfully traverse all terrain types, including deformable surfaces such as sand, which are not included during training. In contrast, MBC shows limited robustness, struggling even on mildly challenging terrain.
Notably, PPF handles these novel terrains in zero-shot deployment, without any additional training or tuning for the new environments, which underscores the benefit of preserving the core locomotion strategy while also learning to adapt to new conditions.

\subsection{Effect of Model-Assumption-based Regularization}
The experiments above show that PPF outperforms FullReg whenever the model assumptions are violated, for example at high speed or on steep slopes. We now examine MAR directly to confirm that it adjusts the regularization weight in response to assumption violations and that this adjustment delivers the primary improvement over FullReg.

\begin{figure}[t]
\centering
\includegraphics[width=0.9\linewidth]{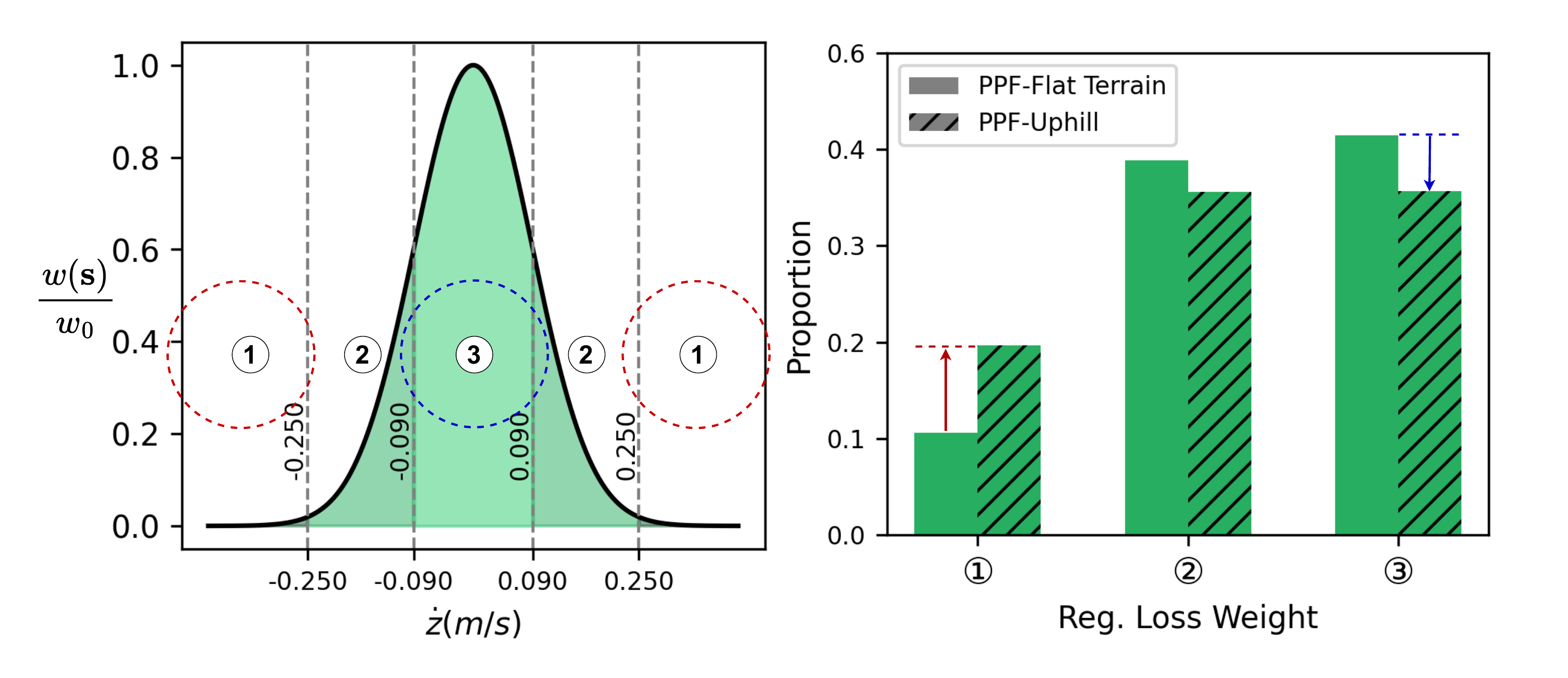}
\caption{\textbf{Left:} The adaptive regularization weight as a function of the model-assumption violation, measured by the vertical CoM velocity $\dot z$. \textbf{Right:} Histogram of regularization weights during PPF training on flat and uphill terrain. Uphill walking produces more low-weight, unreliable samples (region 1) and fewer high-weight, reliable samples (region 3), showing that MAR downweights expert supervision where the ALIP assumption breaks.}
\label{fig:ppf_mar_train}
\vspace{-0.05in}
\end{figure}

\begin{figure}[t]
\centering
\includegraphics[width=0.65\linewidth]{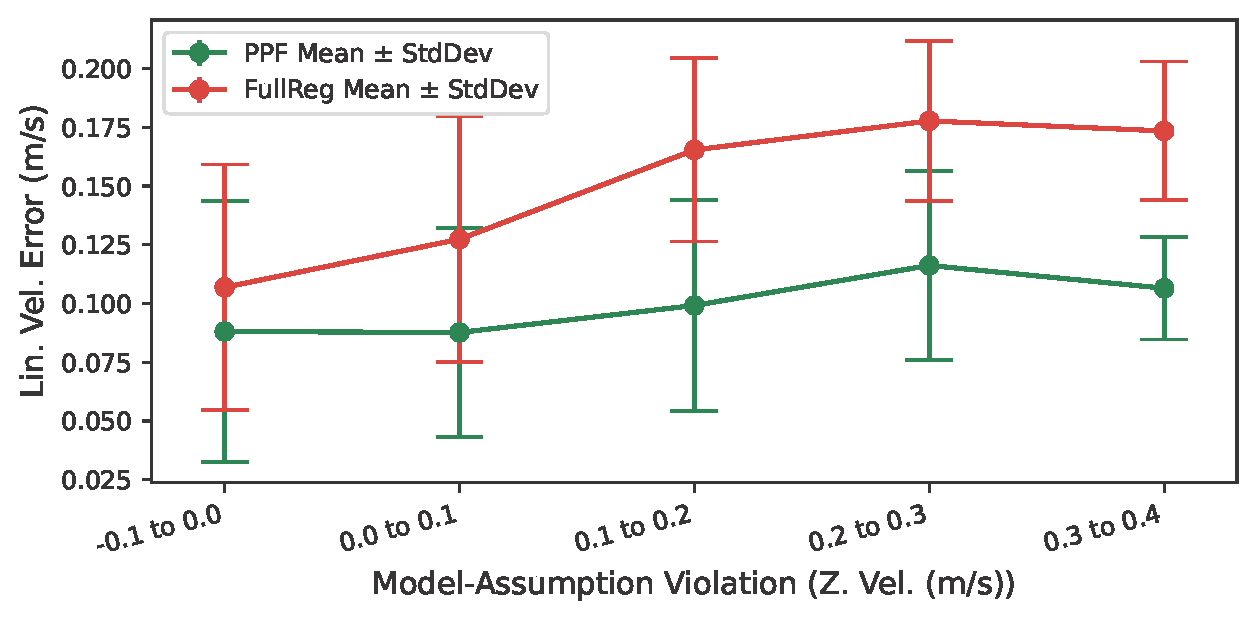}
\caption{Model-assumption violation (vertical CoM velocity $\dot z$) versus linear-velocity tracking error on a $14^\circ$ uphill ramp. FullReg's tracking error grows as violations become more pronounced, while PPF maintains a consistently low error, isolating MAR as the source of PPF's tracking advantage.}
\label{fig:ppf_mar_test}
\vspace{-0.05in}
\end{figure}

\noindent\textbf{During training}, MAR reduces the weight of the imitation term as the vertical CoM velocity $\dot z$ grows, allowing the policy to deviate from the MBC where the ALIP model is unreliable. \ref{fig:ppf_mar_train} plots the distribution of regularization weights on flat and uphill terrain. On flat ground, the assumption rarely breaks, so most samples retain a high weight (region 3) and the policy stays close to the MBC gait. On uphill terrain, violations are frequent, causing a shift toward low weights (region 1) and allowing the RL objective to dominate. MAR thus filters out unreliable expert supervision automatically, without hand-tuning a separate weight per terrain.

\noindent\textbf{During testing}, this selective trust of expert guidance translates into more accurate tracking under violations. \ref{fig:ppf_mar_test} relates the assumption violation $\dot z$ to the linear-velocity tracking error on a $14^\circ$ uphill ramp. FullReg, which imitates the MBC in every state, accumulates larger tracking error as violations grow because it keeps following an expert whose performance degrades on such terrain. PPF maintains consistently low error across the same range of violations. This result explains the tracking-error gap in \ref{tab:ppf_ar_sim_slope}: MAR allows PPF to preserve the MBC's gait style while still exceeding the MBC's capability on terrain that the model does not account for.

\section{Conclusion}

%In summary, the PPF approach successfully imitated the motion style of a model-based locomotion controller and then enhanced it via reinforcement learning. The adaptive regularization (MAR) proved essential to attain high performance: it allowed the policy to surpass the model-based controller’s capabilities (achieving higher speeds and tackling unseen terrains) without suffering the catastrophic forgetting that might come from unrestricted policy updates. 
%Across our tests, PPF outperformed baseline methods in terms of terrain robustness, maximum achievable walking speed, and sim-to-real generalization. These results validate our central hypothesis that RL fine-tuning, when guided by model-based regularization, can produce a humanoid locomotion controller with greater robustness. This represents a significant step towards reliable humanoid robots that can operate in the real world, combining the trustworthiness of model-driven control with the flexibility of learned behaviors.
In summary, the PPF approach imitates the motion style of a model-based
locomotion controller and then enhances it via reinforcement learning. MAR is essential for attaining high performance: it allows the policy
to surpass the model-based controller’s capabilities (achieving higher speeds and tackling
more challenging terrains) without suffering the catastrophic forgetting that might come from
unrestricted policy updates.

Across our tests, PPF consistently outperforms baseline methods in terms of terrain robustness, maximum achievable walking speed,
and sim-to-real generalization. These results validate our central hypothesis that RL
fine-tuning, when guided by model-based regularization, can produce a humanoid
locomotion controller with greater robustness. This represents a step toward reliable humanoid locomotion that combines the style of model-driven control with the robustness of learned behavior.

Several limitations bound these results and motivate future work. MAR relies on a single scalar, the vertical CoM velocity $\dot z$ of the ALIP model, to measure the assumption violation, so it cannot detect failure modes that leave $\dot z$ unaffected, such as foot slip or angular-momentum modeling error.
A natural extension is to derive the regularization signal from additional model assumptions, for example estimated foot slipperiness through stance-foot velocity, since most higher-fidelity models assume zero contact velocity. Pairing MAR with a more capable model-based expert would test whether a stronger prior improves training efficiency and final performance.

PPF demonstrates one instantiation of the third thrust: RL fine-tuning for improving pre-trained capabilities beyond those shown in the demonstrations. The next chapter applies the same pre-training and fine-tuning principle to autonomous loco-manipulation, where reinforcement learning fine-tuning adapts a more complex hierarchical framework involving a diffusion planner and a loco-manipulation controller.

%% file: chapters/REFINE-DP.tex
% NOTE: This chapter requires the following packages in thesis.tex preamble:
%   \usepackage[ruled,vlined]{algorithm2e}
%   \usepackage{booktabs}
%   \usepackage{tabularx}

\chapter{REFINE-DP: Diffusion Policy Fine-Tuning for Humanoid Loco-Manipulation}
\label{chapter:DPFT}

\noindent\textbf{Dissertation Roadmap and Contribution of This Chapter.}
This chapter extends the ``pre-training and fine-tuning" method introduced in \ref{chapter:PPF} from locomotion to autonomous loco-manipulation. Unlike pure locomotion, loco-manipulation presents unique challenges, as humanoids must coordinate whole-body motions under complex robot-environment dynamics over long horizons. To address these challenges, we develop \textbf{RE}inforcement learning \textbf{FINE}-tuning of \textbf{D}iffusion \textbf{P}olicy (REFINE-DP), a hierarchical planner--controller framework for autonomous loco-manipulation.
The hierarchical structure reduces the planner’s action-space complexity and lowers its data requirements by delegating whole-body tracking and contact-rich execution to a separately-trained loco-manipulation controller. We further introduce a fine-tuning approach that jointly optimizes both the planner and controller, reducing the distribution shift between the planner’s output actions and the controller’s input goals. By mitigating this planner--controller mismatch, REFINE-DP improves task success and tracking accuracy on long-horizon loco-manipulation tasks with the Booster T1 humanoid.

% Our method, \textbf{RE}inforcement learning \textbf{FINE}-tuning of \textbf{D}iffusion \textbf{P}olicy (REFINE-DP), \emph{improves pre-trained loco-manipulation policies through reinforcement learning fine-tuning}. 
% Our method, \textbf{RE}inforcement learning \textbf{FINE}-tuning of \textbf{D}iffusion \textbf{P}olicy (REFINE-DP), first pre-trains a high-level diffusion planner using demonstrations from teleoperation and scripted planners. 
%To address this challenge, we develop a joint optimization approach that fine-tunes both the diffusion planner and the RL-based loco-manipulation controller simultaneously.

\begin{figure}[t]
\centerline{\includegraphics[width=0.95\textwidth]{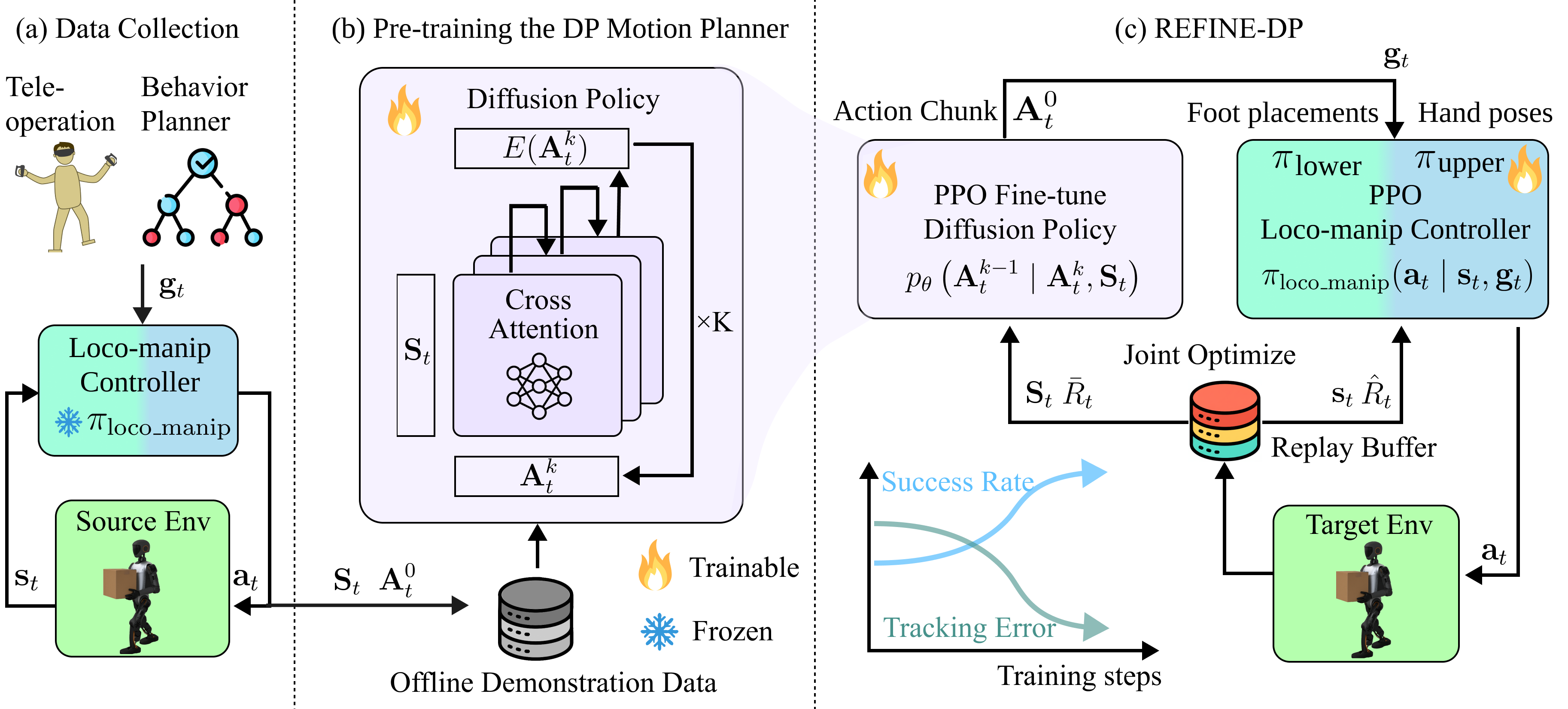}}
\caption{The REFINE-DP pipeline consists of three stages: (a) Data Collection, where expert demonstrations are collected in a source environment using a frozen RL-based loco-manipulation controller via VR teleoperation or heuristic planners; (b) Pre-Training, where a diffusion policy is trained on the expert dataset; and (c) Joint Optimization, where the pre-trained diffusion policy is jointly fine-tuned with the loco-manipulation controller via RL, achieving a higher success rate and better motion tracking quality.}
\label{fig:dpft_framework}
\vspace{-0.05in}
\end{figure}

\section{Introduction}

Humanoid robots are increasingly capable in their physical behaviors, yet they still lack the task-level autonomy required to perform human-like tasks, such as coordinating locomotion and manipulation for door traversal. Recent studies have demonstrated agile whole-body control with both model-based methods~\cite{khazoom2024tailoring,jang2026seec} and learning-based motion imitation~\cite{he2024omnih2o,humanplus}. Despite these advances, reliable execution often still depends on human commands or heuristic planners. Autonomous loco-manipulation requires robots to interact with the physical world through contact, manipulate objects, exert forces on the environment, and recover when the system drifts away from demonstrated states. %This chapter addresses this gap by treating demonstration-derived behavior as a prior and improve the prior through interaction.

These requirements motivate task-level policies that can generate actions from demonstrations while adapting beyond the demonstrated state distribution. In particular, recent advances in Behavior Cloning (BC) have introduced Diffusion Policies (DPs)~\cite{diffusion_policy} for learning action-generative models from expert demonstrations. These models are effective at capturing multimodal expert behaviors. However, their success often relies on offline datasets, which can lead to poor out-of-distribution robustness due to distribution shift and compounding errors. To mitigate this issue, DPs are typically trained on large datasets that require high-capacity models to improve coverage and generalization. However, for high-dimensional humanoid systems, this scaling strategy incurs prohibitive costs in both data collection and policy training computation, while still failing to achieve robust performance in loco-manipulation, where long horizons, high dimensionality, and compounding execution errors exacerbate distribution mismatch.

Rather than scaling data, this chapter addresses the distribution-shift issue through Reinforcement Learning Fine-Tuning (RLFT)~\cite{ren_diffusion_2024}, enabling adaptation beyond the offline pre-training dataset. Specifically, we fine-tune a DP by exploring interactions in a simulator, collecting trials, and directly updating the DP via policy gradient. By exploring unseen state-action pairs, RLFT improves task success with only a sparse reward, allowing a pre-trained DP on a small dataset to outperform policies trained on substantially larger offline datasets.

Deploying a DP on a humanoid robot remains challenging because loco-manipulation is high-dimensional and contact-rich. To simplify the action space of the DP, we adopt a hierarchical framework. The DP serves as a motion planner that outputs Cartesian action chunks, including base-velocity commands and $SE(3)$ hand-pose trajectories, while an RL-based loco-manipulation controller converts these commands into joint-position references and maintains locomotion stability and manipulation accuracy. This hierarchical design keeps the command space low-dimensional and intuitive while delegating the complexity of loco-manipulation control to the dedicated policy described in \ref{sec:RL_train}. \ref{fig:dpft_framework} illustrates the resulting framework.

Based on this hierarchical framework, we introduce REFINE-DP, a fine-tuning approach that jointly optimizes both the DP planner and the loco-manipulation RL controller. Prior work on diffusion policy fine-tuning, such as Diffusion Policy Policy Optimization (DPPO)~\cite{ren_diffusion_2024}, optimizes only the DP planner, while residual RL approaches~\cite{ankile2024imitationrefinementresidual} learn corrective actions on top of a frozen DP. Neither setting addresses the distributional mismatch that arises when the planner's evolving command distribution is executed by a loco-manipulation controller trained on a fixed command distribution. REFINE-DP updates both components during joint optimization, maintaining distributional consistency between the planner's outputs and the controller's inputs and improving command-tracking performance and task success.

The main contributions of this chapter are as follows:

\begin{itemize}
\item \textbf{Hierarchical humanoid framework:} A DP serves as the autonomous planner and an RL policy acts as the loco-manipulation controller. Instead of operating directly in full-body configuration spaces, the DP outputs low-dimensional Cartesian action chunks, including base velocities and $SE(3)$ hand-pose trajectories, tailored to humanoid loco-manipulation. These action chunks substantially simplify the DP action space, support intuitive VR teleoperation, and can be converted by the RL policy into joint-position references.

\item \textbf{Joint optimization of planner and controller:} REFINE-DP simultaneously fine-tunes the diffusion planner through a PPO-based diffusion policy gradient and the RL-based loco-manipulation controller to track the planner's evolving command distribution. This joint optimization reduces the distributional mismatch that degrades motion quality and improves task success and tracking accuracy compared with fine-tuning either component alone.

\item \textbf{Hardware validation:} The framework is validated on the Booster T1 humanoid robot across tasks including door traversal, long-horizon object transport, and stair-assisted object retrieval. We further demonstrate the pipeline using onboard RGB information for object pose estimation, reducing reliance on privileged motion-capture state.
\end{itemize}

\section{Related Work}

\subsection{Sim-to-Real RL for Humanoid Loco-Manipulation}

Loco-manipulation skills involve simultaneous locomotion and manipulation, posing a high-dimensional control problem for humanoid robots~\cite{gu_humanoid_2025}. Prior work has addressed this challenge using sim-to-real RL for whole-body control frameworks, while other approaches decompose the problem into upper- and lower-body controllers~\cite{zhang2025falcon, lu2024MobileTV}. Many of these RL methods~\cite{humanplus, he2024omnih2o, ze2026twist2, ben2025homie} support teleoperation, enabling demonstration data collection for training a motion planner.

Among those RL methods, one category is learning from scratch, where the desired behavior is specified through a complex reward function. However, learning loco-manipulation without demonstrations is challenging due to the high dimensionality of humanoid systems and the diversity and compositional complexity of loco-manipulation tasks. Several studies have demonstrated successful sim-to-real transfer for autonomous object manipulation~\cite{zhang2024wococo, Curiosity-schwarke23a, SR_deepmind_soccer, xue2025openingsimtorealdoorhumanoid}, though these policies are typically tuned for each specific task, making it difficult to acquire long-horizon, autonomous loco-manipulation skills.

The second category uses RL for motion tracking of reference trajectories, either manually designed~\cite{dao2023simtoreal, liu2024opt2skill} or derived from human data~\cite{xie_box_locomanip, he2024omnih2o, humanplus}. While this paradigm has achieved versatile motion tracking capability, it fundamentally relies on externally provided references and therefore lacks the task- and environment-level awareness required for true autonomy. In this chapter, our loco-manipulation controller belongs to the second category. It can walk and manipulate objects with simple commands, including desired base velocity and hand poses, and is more physically interpretable than approaches using latent vector inputs or full-body joint angle commands.

\subsection{Autonomous Humanoid Loco-Manipulation}

To achieve autonomous task execution on humanoid robots, many studies adopt modern behavior cloning techniques, such as DPs~\cite{diffusion_policy} or action-chunking transformers~\cite{aloha}, to learn contact-rich manipulation directly from multimodal demonstration data. However, while most autonomous policies focus on relatively stable manipulator platforms, extending these approaches to autonomous humanoid loco-manipulation remains challenging due to underactuated and unstable dynamics, and the need to coordinate whole-body motion over long horizons.

Among the few prior works on autonomous humanoid loco-manipulation, some approaches adopt hierarchical frameworks that decouple task planning from loco-manipulation control, thereby keeping the action space simple and interpretable~\cite{ze2026twist2, humanplus, yin2025visualmimic}. Other approaches maintain a unified end-to-end policy but rely on large-scale domain randomization to integrate perception and improve sim-to-real robustness~\cite{xue2025openingsimtorealdoorhumanoid, weng2025hdmi}. REFINE-DP's hierarchical framework falls into the first category, with a diffusion planner offering the advantage of learning a diverse range of tasks.

\subsection{Fine-Tuning Pre-Trained Imitation Learning Policies}

To address the inherent issue of compounding errors and domain shift in DPs, several studies have explored integrating RL with DP by learning a residual policy on top of a pre-trained DP to refine actions and enhance performance. These residual policies provide corrective compensation for imitation errors. Existing studies have explored on-policy RL~\cite{ankile2024imitationrefinementresidual, su2026rfsreinforcementlearningresidual} and off-policy RL~\cite{ankile2025residualoffpolicyrlfinetuning} for learning residual refinement to improve precision and sample efficiency.

Another line of work focuses on fine-tuning diffusion policies with RL. In Diffusion Policy Policy Optimization (DPPO)~\cite{ren_diffusion_2024}, the authors adapt Proximal Policy Optimization (PPO)~\cite{PPO} to fine-tune a pre-trained diffusion policy for a higher success rate. A related approach~\cite{ma2025efficientonlinereinforcementlearning} introduces Diffusion Policy Mirror Descent (DPMD) and Soft Diffusion Actor Critic (SDAC) for online training of diffusion policies from scratch. More broadly, $\pi_\text{RL}$~\cite{chen2026piRL} adopts RL for fine-tuning vision-language-action models. While prior work optimizes only the planner, few studies jointly optimize a diffusion planner and a loco-manipulation controller within a unified planning and control framework.

\section{Methods}

The REFINE-DP pipeline consists of three stages illustrated in \ref{fig:dpft_framework}: (i) expert data collection using the loco-manipulation controller, (ii) diffusion policy pre-training, and (iii) joint optimization via RL fine-tuning.

\subsection{Loco-Manipulation Controller for Data Collection}
\label{sec:RL_train}

\begin{figure}[t]
  \centering
  \includegraphics[width=0.6\columnwidth]{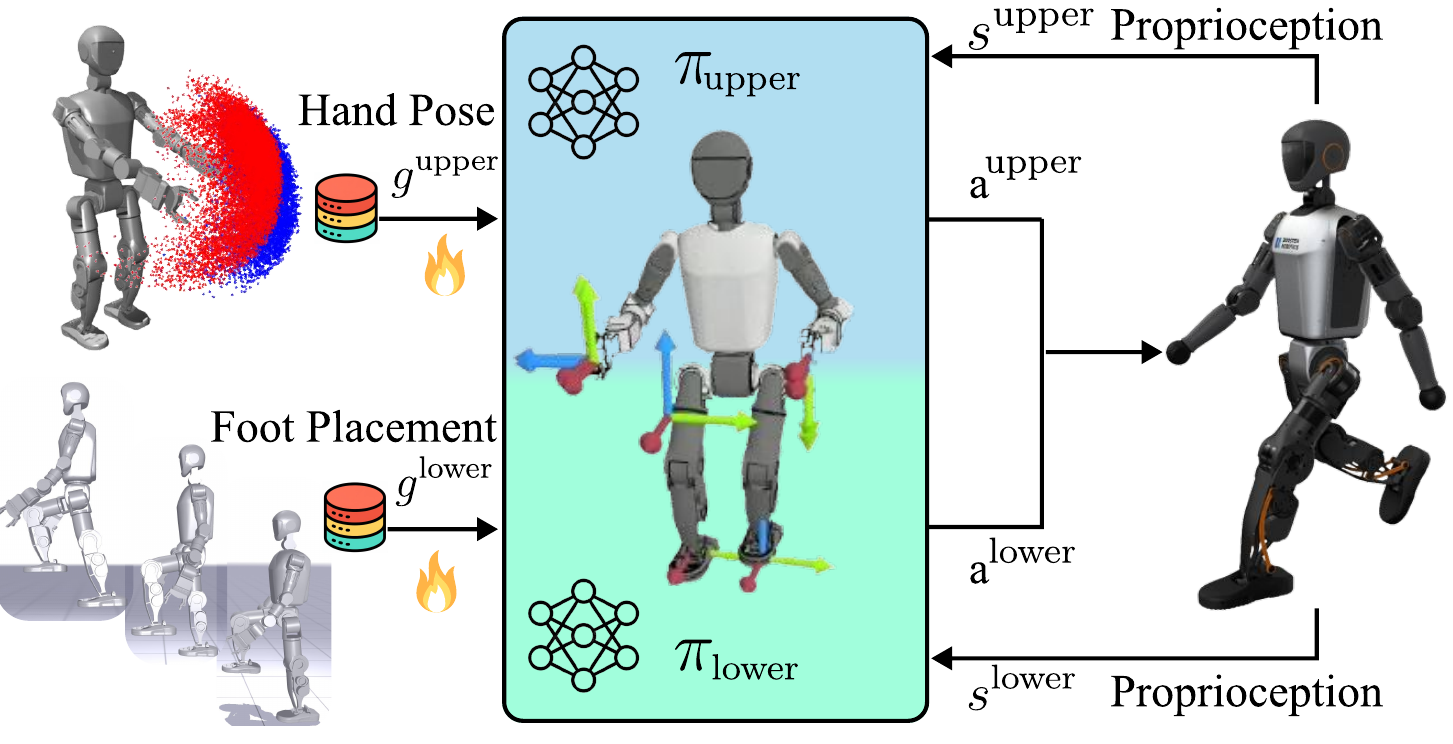}
  \caption{The sim-to-real RL training pipeline for the loco-manipulation control policy.}
  \label{fig:RL_flow}
\vspace{-0.05in}
\end{figure}

We develop an RL policy $\pi_{\text{loco\_manip}}$ capable of simultaneous locomotion and manipulation, illustrated in \ref{fig:RL_flow}. Following prior humanoid loco-manipulation works~\cite{zhang2025falcon, lu2024MobileTV}, $\pi_{\text{loco\_manip}}$ adopts a decoupled architecture that separates upper-body and lower-body control: an upper-body policy $\pi_{\rm upper}$ tracks desired hand poses to execute manipulation, while a lower-body policy $\pi_{\rm lower}$ achieves precise foot placements while maintaining dynamic balance. The two policies are trained independently, each handling the disturbances induced by the other through domain randomization. We describe the upper-body policy first, then the lower-body policy, and finally combine them into $\pi_{\text{loco\_manip}}$ for data collection.

\subsubsection{Upper-Body Policy $\pi_{\rm upper}$}
The upper-body policy $\pi_{\rm upper}(\mathbf{a}^{\rm upper}_t \mid \mathbf{s}^{\rm upper}_t, \mathbf{g}^{\rm upper}_t)$ outputs an offset $\mathbf{a}^{\rm upper}_t$ from a default upper-body joint configuration while tracking a hand-pose command $\mathbf{g}^{\rm upper}_t$ that specifies SE(3) targets for the left and right hands in the robot's root frame. We generate feasible hand commands by sampling collision-free arm joint configurations and solving forward kinematics. The observation $\mathbf{s}^{\rm upper}_t$ comprises arm joint states, current hand poses, and tracking errors. To improve robustness to payload variation and to the external forces arising from object interaction and lower-body motion, we apply domain randomization over payload mass and joint friction during training.

\subsubsection{Lower-Body Policy $\pi_{\rm lower}$}
Most locomotion policies are velocity-tracking controllers designed for long-distance periodic walking, whereas loco-manipulation demands frequent start-stop transitions and precise torso positioning: a small base-pose error can prevent the hands from reaching objects such as door handles or boxes. A velocity interface does not directly enforce this positional objective, and at low speed it tends to step in place, overshoot, or ignore small commands when a dead zone is applied. We therefore condition $\pi_{\rm lower}$ on a discrete foot-placement command that specifies where each step should land, giving direct control over the number and location of footsteps used to approach a target. Unlike velocity tracking, which accumulates positional error over time, this formulation yields substantially improved foot-placement accuracy and locomotion stability~\cite{wang2025beamdojo, suliman2025reinforcement}.

We represent this command as a \textbf{stance-to-swing command} $\mathbf{g}^{\rm lower}_t$, which encodes the SE(3) pose of the swing foot relative to the stance foot's coordinate frame, a countdown timer that provides temporal context, and a binary flag indicating which foot is swinging. This representation is versatile across command sources: a joystick can issue discrete foothold commands, a VR foot tracker can relay a human operator's relative foot pose in real time, and scripted planners can generate foothold sequences for automated data collection. The full command specification is summarized in \ref{tab:swing-foot-observation}.

\begin{table}[t]
\caption{Stance-to-swing foot-placement command representation.}
\label{tab:swing-foot-observation}
\begin{center}
\begin{tabular}{p{3.5cm} p{7cm} p{2.5cm}}
\hline
\textbf{Command Term} & \textbf{Description} & \textbf{Example} \\
\hline
Swing foot pose in stance foot frame &
Desired pose of swing foot relative to stance foot. Includes position $(x, y, z)$ in meters and orientation as a rotation matrix. &
\begin{tabular}[t]{@{}l@{}}
$[0.12,\ {-0.22},\ 0.08,$ \\
$1,0,0,\ 0,1,0,\ 0,0,1]$
\end{tabular} \\
\hline
Countdown time &
Scalar from $0.8 \to 0.0$. Jump $0.0\!\to\!0.8$ signals a new command. $0.8\!\to\!0.5$: CoM shift to stance foot; $0.5\!\to\!0.0$: swing phase. &
$0.35$ \\
\hline
Swing foot indicator &
Binary flag: $0$ for left swing, $1$ for right swing. &
$1$ \\
\hline
\end{tabular}
\end{center}
\end{table}

We train $\pi_{\rm lower}(\mathbf{a}^{\rm lower}_t \mid \mathbf{s}^{\rm lower}_t, \mathbf{g}^{\rm lower}_t)$ by RL motion imitation~\cite{DeepMimic2018} of model-based lower-body reference trajectories, where $\mathbf{a}^{\rm lower}_t$ denotes an offset from a default lower-body joint configuration. For terrain such as flat ground and stairs, we generate references with a linear inverted pendulum model (LIPM) combined with lower-body inverse kinematics~\cite{liu2024opt2skill}: each reference corresponds to a specific stair rise and tread depth and consists of pre-swing, swing, and post-swing phases. To diversify the motions, we randomize the swing foot's lift-off and landing positions via discretized grid sampling; for each sampled pair, a cubic-spline swing trajectory that avoids collision with the stairs is precomputed, and a trajectory optimization solves the joint-level stepping motion.

\begin{figure}[h]
\centering
\includegraphics[width=0.95\columnwidth]{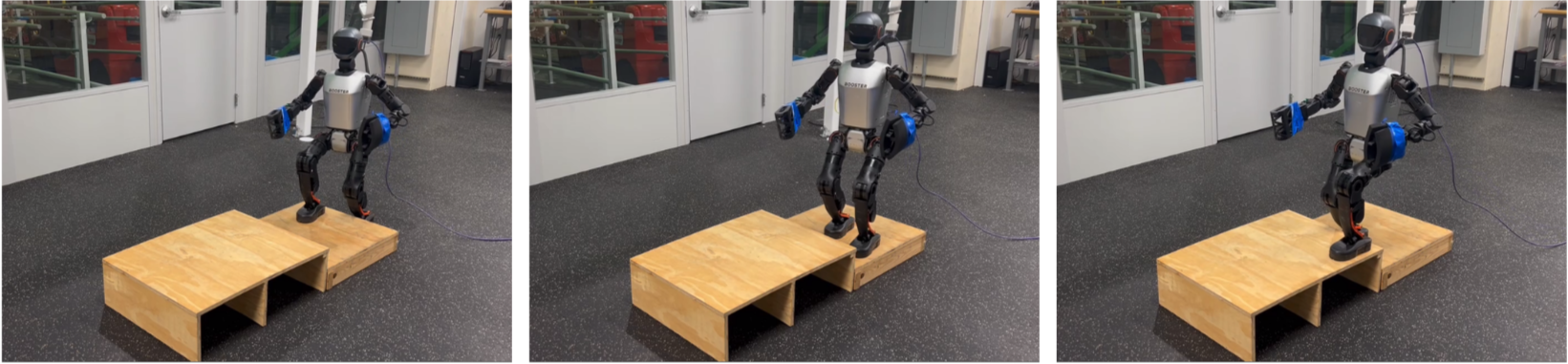}
\caption{The lower-body policy enables the robot to walk on flat ground and stairs.}
\label{fig:stair}
\vspace{-0.05in}
\end{figure}

During training, the robot learns to imitate these reference motions while maintaining balance. Taking stair climbing as a representative example, each parallel simulation environment initializes the robot on a stair step whose rise-and-tread configuration matches the pre-swing and post-swing poses of the sampled reference. We adopt an asymmetric actor-critic network: the actor observation $\mathbf{s}^{\rm lower}_t$ includes joint positions and velocities, projected gravity, base angular velocity, the stance-to-swing command $\mathbf{g}^{\rm lower}_t$, and previous actions, spanning 10 history time steps (0.2 s at 50 Hz) to capture short-term dynamics; the critic additionally accesses privileged information such as base linear velocity and foot contact wrenches. The action space is restricted to the lower-body joints, while the upper-body joints are held at a fixed configuration by PD control. The reward combines foot-tracking terms with regularization terms for accuracy and stability. Driven by these inputs and the motion-imitation objective, the policy learns to execute the targeted swing-foot placements precisely, walking on both flat ground and stairs as shown in \ref{fig:stair}.

\subsubsection{Combined Loco-manipulation Policy $\pi_{\rm {loco\_manip}}$}
Both $\pi_{\rm lower}$ and $\pi_{\rm upper}$ output joint-position offsets relative to a fixed default configuration $\mathbf{q}_{\rm def}$. Stacking the two gives the full action $\mathbf{a}_t = [\mathbf{a}^{\rm lower}_t;\, \mathbf{a}^{\rm upper}_t]$, where $\mathbf{a}^{\rm lower}_t \in \mathbb{R}^{12}$ drives the leg joints and $\mathbf{a}^{\rm upper}_t \in \mathbb{R}^{14}$ drives the arm joints. The combined policy is $\pi_{\text{loco\_manip}}(\mathbf{a}_t \mid \mathbf{s}_t, \mathbf{g}_t)$ with observation $\mathbf{s}_t = [\mathbf{s}^{\rm lower}_t;\, \mathbf{s}^{\rm upper}_t]$ and command $\mathbf{g}_t = [\mathbf{g}^{\rm lower}_t;\, \mathbf{g}^{\rm upper}_t]$. The target joint positions $\mathbf{q}_{\rm target} = \mathbf{q}_{\rm def} + \mathbf{a}_t$ are tracked by proportional-derivative (PD) control. Full observation and reward terms are listed in \ref{tab:loco_manip_obs}, \ref{tab:loco_rewards}, and \ref{tab:manip_rewards}.

\begin{table}[t]
\centering
\caption{Observation terms for RL policy training}
\label{tab:loco_manip_obs}
\small
\begin{tabular}{l c c}
\hline
\textbf{Observation} & \textbf{Actor} & \textbf{Critic} \\
\hline
\multicolumn{3}{l}{\textit{Manipulation}} \\
\hline
Projected gravity & \checkmark & \checkmark \\
Arm joint pos. & \checkmark & \checkmark \\
Arm joint vel. & \checkmark & \checkmark \\
Action & \checkmark & \checkmark \\
Hand pose cmd. ($\mathbf{p}_L^{\mathrm{cmd}}, \mathbf{p}_R^{\mathrm{cmd}}$) & \checkmark & \checkmark \\
Current hand poses ($\mathbf{p}_L, \mathbf{p}_R$) & \checkmark & \checkmark \\
Hand tracking error ($\mathbf{e}_L, \mathbf{e}_R$) & \checkmark & \checkmark \\
\hline
\multicolumn{3}{l}{\textit{Locomotion}} \\
\hline
Foot placement cmd. & \checkmark & \checkmark \\
Base angular vel. & \checkmark & \checkmark \\
Projected gravity & \checkmark & \checkmark \\
Leg joint pos. & \checkmark & \checkmark \\
Leg joint vel. & \checkmark & \checkmark \\
Action & \checkmark & \checkmark \\
Reference foot pos. & & \checkmark \\
Current foot pos. & & \checkmark \\
Base linear vel. & & \checkmark \\
Foot wrench & & \checkmark \\
\hline
\end{tabular}
\end{table}

\begin{table}[t]
\caption{Reward terms for foot-placement tracking policy}
\label{tab:loco_rewards}
\centering
\small
\setlength{\tabcolsep}{3pt}
\renewcommand{\arraystretch}{1.05}
\begin{tabularx}{\columnwidth}{@{}l >{\raggedright\arraybackslash}X c@{}}
\toprule
\textbf{Term} & \textbf{Formulation} & \textbf{Weight} \\
\midrule
\multicolumn{3}{l}{\textit{End-effector reference tracking (per foot)}} \\
Position ($\sigma=0.05$) &
$\exp(-\lVert\mathbf{p}-\mathbf{p}^{\mathrm{ref}}\rVert^{2}/\sigma^{2})$ & $+5.0$ \\
Yaw ($\sigma=0.1$) &
$\exp(-(\psi-\psi^{\mathrm{ref}})^{2}/\sigma^{2})$ & $+3.0$ \\
Base vel. ($\sigma=0.25$) &
$\exp(-\lVert\mathbf{v}-\mathbf{v}^{\mathrm{ref}}\rVert^{2}/\sigma^{2})$ & $+1.0$ \\
\midrule
\multicolumn{3}{l}{\textit{Joint reference tracking}} \\
Position ($\sigma=0.3$) &
$\exp(-\lVert\mathbf{q}-\mathbf{q}^{\mathrm{ref}}\rVert^{2}/\sigma^{2})$ & $+0.5$ \\
Velocity ($\sigma=1.0$) &
$\exp(-\lVert\dot{\mathbf{q}}-\dot{\mathbf{q}}^{\mathrm{ref}}\rVert^{2}/\sigma^{2})$ & $+0.3$ \\
\midrule
\multicolumn{3}{l}{\textit{Regularization (with curriculum $\rightarrow$)}} \\
Action smoothness &
$\lVert\mathbf{a}_t-\mathbf{a}_{t-1}\rVert^{2}$ & $-0.001 \rightarrow -0.5$ \\
Joint acceleration &
$\lVert\ddot{\mathbf{q}}\rVert^{2}$ & $-10^{-7} \rightarrow -10^{-5}$ \\
Torque limits &
$\max(0,\lvert\boldsymbol{\tau}\rvert-\boldsymbol{\tau}_{\max})$ & $-0.5$ \\
Joint limits &
$\max(0,\lvert\mathbf{q}\rvert-\mathbf{q}_{\max})$ & $-10^{-4}$ \\
Ankle posture &
$\lVert\mathbf{q}-\mathbf{q}_{\mathrm{def}}\rVert^{2}$ & $-2.0$ \\
Base ori. ($\sigma=0.2$) &
$\exp(-\lVert\mathbf{g}_{\mathrm{xy}}\rVert^{2}/\sigma^{2})$ & $+3.0$ \\
Foot airtime ($\tau=0.4$) &
$\sum_{i} c_i(t^{\mathrm{air}}_i-\tau)$ & $+20.0$ \\
Base height range &
$\begin{aligned}
&[\max(0,h_{\min}-h)]^2\\[-1pt]
&+[\max(0,h-h_{\max})]^2
\end{aligned}$ & $-3.0$ \\
\bottomrule
\end{tabularx}
\end{table}

\begin{table}[t]
\caption{Reward terms for hand-pose tracking policy}
\label{tab:manip_rewards}
\centering
\small
\setlength{\tabcolsep}{4pt}
\renewcommand{\arraystretch}{1.05}
\begin{tabularx}{\columnwidth}{@{}l >{\raggedright\arraybackslash}X c@{}}
\toprule
\textbf{Term} & \textbf{Formulation} & \textbf{Weight} \\
\midrule
\multicolumn{3}{l}{\textit{End-effector position tracking (per hand)}} \\
Coarse ($\sigma=0.4$)   & $\exp(-\lVert\mathbf{p}-\mathbf{p}^{\mathrm{cmd}}\rVert^2/\sigma^2)$ & $+2.0$ \\
Fine ($\sigma=0.1$)     & $\exp(-\lVert\mathbf{p}-\mathbf{p}^{\mathrm{cmd}}\rVert^2/\sigma^2)$ & $+2.0$ \\
Precise ($\sigma=0.08$) & $\exp(-\lVert\mathbf{p}-\mathbf{p}^{\mathrm{cmd}}\rVert^2/\sigma^2)$ & $+2.0$ \\
\midrule
\multicolumn{3}{l}{\textit{End-effector orientation tracking (per hand)}} \\
Coarse ($\sigma=0.8$)   & $\exp(-\lVert\mathbf{e}_{\mathrm{quat}}\rVert^2/\sigma^2)$ & $+1.0$ \\
Fine ($\sigma=0.5$)     & $\exp(-\lVert\mathbf{e}_{\mathrm{quat}}\rVert^2/\sigma^2)$ & $+1.0$ \\
Precise ($\sigma=0.3$)  & $\exp(-\lVert\mathbf{e}_{\mathrm{quat}}\rVert^2/\sigma^2)$ & $+1.0$ \\
\midrule
\multicolumn{3}{l}{\textit{Regularization (with curriculum $\rightarrow$)}} \\
Posture prior       & $\lVert\mathbf{q}-\mathbf{q}_{\mathrm{nom}}\rVert_1$ & $-0.2$ \\
Action smoothness   & $\lVert\mathbf{a}_t-\mathbf{a}_{t-1}\rVert^2$       & $-0.01 \rightarrow -0.1$ \\
Joint velocity      & $\lVert\dot{\mathbf{q}}\rVert^2$                      & $-10^{-3} \rightarrow -2\times 10^{-3}$ \\
Joint acceleration  & $\lVert\ddot{\mathbf{q}}\rVert^2$                     & $-10^{-6} \rightarrow -3\times 10^{-6}$\\
EE acceleration     & $\lVert\ddot{\mathbf{p}}_{\mathrm{ee}}\rVert^2$         & $-1.4\times 10^{-2}$ \\
Torque limits       & $\max(0,\lvert\boldsymbol{\tau}\rvert-\boldsymbol{\tau}_{\max})$ & $-0.1$ \\
Joint limits        & $\max(0,\lvert\mathbf{q}\rvert-\mathbf{q}_{\max})$   & $-4.0$ \\
\bottomrule
\end{tabularx}
\end{table}

\subsubsection{Data Collection using $\pi_{\rm {loco\_manip}}$}

We collect demonstrations in simulation using two methods, both executed through $\pi_{\text{loco\_manip}}$: VR teleoperation and heuristic planner rollouts. The heuristic planners are designed per task as stage-conditioned behavior transitions, with intermediate waypoints parameterized from the observed object poses. The two methods play complementary roles. Teleoperation captures rich behaviors such as perturbation recovery, but is costly and low-throughput because whole-body humanoid coordination places substantial demands on operator skill. In contrast, heuristic planners are cheaper to roll out at scale, but they typically produce less diverse behaviors. Balancing this trade-off, we augment 50 teleoperated trajectories with heuristic planner rollouts to expand the dataset to 1000 trajectories: the teleoperated trajectories define the core behavioral modes, including recovery responses, while the heuristic rollouts broaden coverage.

To cover diverse scenarios, we randomize both the initial object configurations and the robot's initial states during collection; further diversity arises from locomotion drift, which randomizes the torso pose and the manipulation timing. We retain only successful trajectories, each a sequence of state-action pairs. The recorded state includes the robot's hand and foot poses, the gripper state, and the target object's pose, all expressed in the body frame. The action specifies the desired hand poses, the gripper state, and a base-velocity command, which a velocity-to-footstep planner converts into a sequence of footstep commands.

\subsection{Diffusion Policy Pre-Training}

The collected loco-manipulation data is used to pre-train a diffusion policy $\bar{\pi}_\theta$~\cite{diffusion_policy}, where $\theta$ denotes the policy parameters. Let $\mathbf{S}_t$ denote the chunk of observations at environment time step $t$, let $\mathbf{A}_t^k$ denote the corresponding action chunk at denoising step $k$, and let $K$ denote the total number of denoising steps. Formally, a DP models the conditional action distribution
\[
p_\theta(\mathbf{A}_t^{0:K} \mid \mathbf{S}_t)
=
p(\mathbf{A}_t^{K})
\prod_{k=1}^{K} p_\theta(\mathbf{A}_t^{k-1} \mid \mathbf{A}_t^{k}, \mathbf{S}_t),
\]
where $\mathbf{A}_t^{0:K}$ denotes the full denoising chain from the clean action chunk $\mathbf{A}_t^0$ to the initial noise chunk $\mathbf{A}_t^K$, and $p(\mathbf{A}_t^K)$ is the standard Gaussian prior over the initial noise chunk.

The action distribution is learned by modeling a sequence of progressively corrupted actions generated through a forward diffusion process, where Gaussian noise is incrementally added:
\begin{equation}
q(\mathbf{A}_t^{k} \mid \mathbf{A}_t^{k-1}) = \mathcal{N}\!\left(\sqrt{\alpha_k}\mathbf{A}_t^{k-1},\, (1-\alpha_k)\mathbf{I}\right),
\end{equation}
where $q(\mathbf{A}_t^{k} \mid \mathbf{A}_t^{k-1})$ is the forward diffusion kernel, $\{\alpha_k\}_{k=1}^{K}$ is a pre-scheduled decreasing sequence of noise-schedule coefficients, and $\mathbf{I}$ is the identity covariance matrix. In practice, we sample noise-corrupted actions directly using the closed-form marginal:
\begin{equation}
\mathbf{A}_t^{k}
= \sqrt{\bar{\alpha}_k}\,\mathbf{A}_t^{0}
+ \sqrt{1 - \bar{\alpha}_k}\,\epsilon, \quad
\bar{\alpha}_k = \prod_{j=1}^k \alpha_j, \quad \epsilon \sim \mathcal{N}(0, \mathbf{I}).
\end{equation}
Here $\bar{\alpha}_k$ is the cumulative product of the schedule coefficients through denoising step $k$, $j$ indexes the coefficients in this product, and $\epsilon$ is standard Gaussian noise. Given the dataset $\mathcal{D}=\{(\mathbf{S}_i, \mathbf{A}_i^0)\}_{i=1}^N$, where $i$ indexes demonstrations and $N$ is the number of demonstration chunks, we train a noise prediction network $\epsilon_\theta$ by minimizing the noise prediction error:
\begin{equation}
\mathcal{L}_{\text{diff}}(\theta) = \mathbb{E}_{\mathbf{A}_t^0, \mathbf{S}_t, k, \epsilon}\!\left[\|\epsilon - \epsilon_\theta(\mathbf{A}_t^{k}, \mathbf{S}_t, k)\|^2\right].
\end{equation}
Here $\mathcal{L}_{\text{diff}}(\theta)$ is the diffusion training loss, $\mathbb{E}$ is the expectation over training chunks, denoising steps, and sampled noise, and $\epsilon_\theta(\mathbf{A}_t^{k}, \mathbf{S}_t, k)$ is the network prediction of the target noise $\epsilon$ conditioned on the noisy action chunk, observation chunk, and denoising step.
At inference time, the policy samples an initial Gaussian noise $\mathbf{A}_t^{K} \sim \mathcal{N}(0, \mathbf{I})$ and progressively denoises it to produce the clean action $\mathbf{A}_t^{0}$. At each denoising step, $\epsilon_\theta$ predicts the noise component that refines a noisy action sample $\mathbf{A}_t^{k}$ toward the next action $\mathbf{A}_t^{k-1}$:
\begin{equation}
\label{eq:denoise}
p_\theta(\mathbf{A}_t^{k-1} \mid \mathbf{A}_t^{k}, \mathbf{S}_t)
= \mathcal{N}\!\left(\mathbf{A}_t^{k-1};\,\boldsymbol{\mu}_\theta(\mathbf{A}_t^k,\mathbf{S}_t),\, \boldsymbol{\sigma}_k^2 \mathbf{I}\right),
\end{equation}
where the reverse-process mean $\boldsymbol{\mu}_\theta$ and variance coefficient $\boldsymbol{\sigma}_k^2$ are
\[
\boldsymbol{\mu}_\theta(\mathbf{A}_t^k,\mathbf{S}_t)
= \frac{1}{\sqrt{\alpha_k}}\!\left(\mathbf{A}_t^{k} - \frac{1-\alpha_k}{\sqrt{1-\bar{\alpha}_k}}\epsilon_\theta(\mathbf{A}_t^{k}, \mathbf{S}_t, k)\right),
\quad
\boldsymbol{\sigma}_k^2 = \frac{1-\bar{\alpha}_{k-1}}{1-\bar{\alpha}_k}\cdot(1-\alpha_k).
\]

The DP uses a Diffusion Transformer (DiT) backbone with observation horizon 8 and action chunk size 12, running at 10 Hz. We choose $\bar{\alpha}_k$ to follow the cosine schedule~\cite{nichol_improved_2021} for pre-training. The loco-manipulation controller runs at 50 Hz.

\subsection{Diffusion Policy Fine-Tuning via DPPO}
\label{sec:dpft}

While using the DP as a diffusion planner achieves a decent success rate, its performance remains insufficient for reliable task execution. The pre-trained planner does not explicitly account for the robot's closed-loop dynamics and execution errors, resulting in deviations that accumulate over long horizons and occasional task failure. We therefore employ RL to fine-tune the pre-trained DP, which offers two key advantages: (i) improving task success through trial-and-error interaction with the simulation environment, and (ii) enhancing robustness to out-of-distribution conditions through domain randomization, enabling zero-shot sim-to-real transfer.

Conventional policy gradient methods such as PPO~\cite{PPO} require evaluating a policy density $\bar{\pi}_\theta(\mathbf{A}_t^0 \mid \mathbf{S}_t)$, which is intractable for diffusion policies. We therefore adopt Diffusion Policy Policy Optimization (DPPO)~\cite{ren_diffusion_2024}, which augments the environment MDP by incorporating the denoising process into the MDP and treating each denoising step as a decision step. Since denoising transitions follow a tractable Gaussian distribution, this augmented MDP enables likelihood-based policy gradients.

Formally, we define the environment MDP as $\mathcal{M}_{\text{ENV}} := (\mathcal{S}, \mathcal{A}, P_0, P, R)$, where $\mathcal{S}$ is the state space, $\mathcal{A}$ is the action space, $P_0$ is the initial state distribution, $P$ is the transition kernel, and $R$ is the reward function. We then define a diffusion-process-augmented MDP $\mathcal{M}_\text{DP}:= (\bar{\mathcal{S}}, \bar{\mathcal{A}}, \bar{P}_0, \bar{P}, \bar{R})$, where the barred quantities denote the corresponding augmented state space, action space, initial distribution, transition kernel, and reward function. This augmented MDP expands $\mathcal{M}_{\text{ENV}}$ by inserting one full denoising process per environment time step and defining a unified index $\bar{t}(t,k) = tK + (K-1-k)$ for environment time step $t$ and denoising step $k \in \{0,\ldots,K-1\}$. The augmented state $\bar{s}_{\bar{t}}$, action $\bar{a}_{\bar{t}}$, and initial distribution $\bar{P}_0$ are:
\[
\bar{s}_{\bar{t}(t,k)} = (\mathbf{S}_t, \mathbf{A}_t^{k+1}),\quad
\bar{a}_{\bar{t}(t,k)} = \mathbf{A}_t^k,\quad
\bar{P}_0 = P_0 \otimes \mathcal{N}(0,\mathbf{I}).
\]
Here $\otimes$ denotes the product distribution between the environment initial state and the initial Gaussian diffusion noise.
The augmented transition dynamics are:
\begin{equation}
\bar{P}\!\left(\bar{s}_{\bar{t}+1} \mid \bar{s}_{\bar{t}}, \bar{a}_{\bar{t}}\right) =
\begin{cases}
(\mathbf{S}_t, \mathbf{A}_t^{k}) \sim \boldsymbol{\delta}_{\mathbf{S}_t,\mathbf{A}_t^k}, & k>0 \\[4pt]
(\mathbf{S}_{t+1}, \mathbf{A}_{t+1}^{K}) \sim P(\mathbf{S}_{t+1} \mid \mathbf{S}_{t}, \mathbf{A}_{t}^{0}) \otimes \mathcal{N}(0, \mathbf{I}), & k=0
\end{cases}
\end{equation}
where $\boldsymbol{\delta}$ is the Dirac distribution. Intermediate denoising steps ($k>0$) receive zero reward; only the final denoised action $\mathbf{A}_t^0$ is executed in $\mathcal{M}_{\text{ENV}}$ and receives the task reward:
\begin{equation}
\bar{R}_{\bar{t}(t, k)}\!\left(\bar{s}_{\bar{t}}, \bar{a}_{\bar{t}}\right)=
\begin{cases}
0, & k > 0 \\
R_t(\mathbf{S}_{t}, \mathbf{A}_{t}^{0}), & k = 0.
\end{cases}
\end{equation}
Here $R_t(\mathbf{S}_{t}, \mathbf{A}_{t}^{0})$ is the environment reward at time step $t$ after executing the clean action chunk $\mathbf{A}_t^0$. For the DP update, we optimize the policy gradient objective:
\begin{align}
\nabla_{\theta} \bar{\mathcal{J}}(\bar{\pi}_{\theta})
&= \mathbb{E}_{\bar{\pi}_{\theta}, \bar{P}, \bar{P}_0} \!\left[
    \sum_{\bar{t} \geq 0}
    \nabla_{\theta} \log \bar{\pi}_{\theta}(\bar{a}_{\bar{t}} \mid \bar{s}_{\bar{t}}) \,
    \bar{r}(\bar{s}_{\bar{t}}, \bar{a}_{\bar{t}})
\right],
\label{eq:objective}
\end{align}
where $\bar{\mathcal{J}}$ is the augmented-MDP return objective, $\bar{\pi}_{\theta}(\bar{a}_{\bar{t}(t,k)}\mid\bar{s}_{\bar{t}(t,k)}) = p_\theta(\mathbf{A}_{t}^k\mid\mathbf{A}_t^{k+1}, \mathbf{S}_t)$ is the tractable denoising-step policy density, and $\bar{r}(\bar{s}_{\bar{t}}, \bar{a}_{\bar{t}}) := \sum_{\tau \geq \bar{t}} \gamma(\tau)\,\bar{R}_{\tau}(\bar{s}_{\tau}, \bar{a}_{\tau})$ is the discounted return from augmented time $\bar{t}$. In this return, $\tau$ indexes future augmented time steps and $\gamma(\tau)$ denotes the corresponding discount factor. We use generalized advantage estimation (GAE)~\cite{schulman_high-dimensional_2018} to calculate the advantage estimate from a replay buffer and optimize the DP using a PPO-style clipped surrogate loss over the denoising chain $\mathbf{A}_t^{K:0}$.

\subsection{Joint Optimization of the Diffusion and RL Policies}

Fine-tuning the DP alone can degrade controller tracking performance, as the DP can issue excessive hand-pose commands to compensate for the limited tracking capability of a frozen loco-manipulation controller. Large command errors can increase contact forces and produce jerky motion when contact changes unexpectedly. To address this, REFINE-DP jointly optimizes both the DP planner and the loco-manipulation controller, matching the controller's tracking capability to the planner's evolving command distribution.

As shown in \ref{fig:dpft_framework}(c), the augmented MDP $\mathcal{M}_\text{DP}$ produces augmented DP states $\bar{s}_{t}$, loco-manipulation controller states $\mathbf{s}_{t}$, DP rewards $\bar{R}_{t}$, and controller tracking rewards $\hat{R}_{t}$. The reward $\bar{R}_{t}$ is used to update the DP $\bar{\pi}_{\theta}$ for higher task success, while $\hat{R}_{t}$ is used to update $\pi_{\rm loco\_manip}$ to accurately track the planner's evolving command distribution. Both policies are optimized via PPO.

Joint optimization enables the tracking controller to accurately follow dynamic, task-relevant commands generated by the DP planner. Unlike the independently sampled, stationary commands used during controller pre-training, DP commands represent a moving target along a continuous trajectory, creating a distribution mismatch that degrades tracking performance. Joint optimization alleviates this mismatch by exposing the controller to planner-generated commands, bringing them in-distribution and substantially improving tracking accuracy. The full algorithm is detailed in \ref{alg:co-train}.

\begin{algorithm}[tb]
\caption{REFINE-DP: Joint Optimization of DP and RL}
\label{alg:co-train}
\SetAlgoLined
\KwIn{Pre-trained diffusion policy $\bar{\pi}_\theta(\mathbf{A}_t^0 |\mathbf{S}_t)$, loco-manipulation policy $\pi_\textrm{loco\_manip}$, augmented MDP $\mathcal{M}_\text{DP}$, outer-loop count $L$, controller update count $M$, and DP update count $N$}
Initialize state $\bar{s}_{t}$ from $\mathcal{M}_\text{DP}$\;
\For{$iter=0$ \KwTo $L$}{
   Rollout in $\mathcal{M}_\text{DP}$ to collect replay buffer $D$\;
   \For{$iter=0$ \KwTo $M$}{
      Sample mini-batch $D_k$ from $D$\;
      Update $\pi_\textrm{loco\_manip}$ using $D_k$ via PPO~\cite{PPO}\;
   }
   Rollout in $\mathcal{M}_\text{DP}$ to collect replay buffer $\bar{D}$\;
   \For{$iter=0$ \KwTo $N$}{
      Sample mini-batch $\bar{D}_k$ from $\bar{D}$\;
      Update $\bar{\pi}_\theta(\mathbf{A}_t^0 |\mathbf{S}_t)$ using $\bar{D}_k$ via DPPO (\ref{eq:objective})\;
   }
}
\end{algorithm}
% Here $D$ is the controller replay buffer, $D_k$ is a mini-batch sampled from $D$, $\bar{D}$ is the DP replay buffer, and $\bar{D}_k$ is a mini-batch sampled from $\bar{D}$.

\section{Experiments}

\subsection{Experiment Setup}
\label{sec:experiment}

\begin{figure}[t]
\centering
\includegraphics[width=0.90\textwidth]{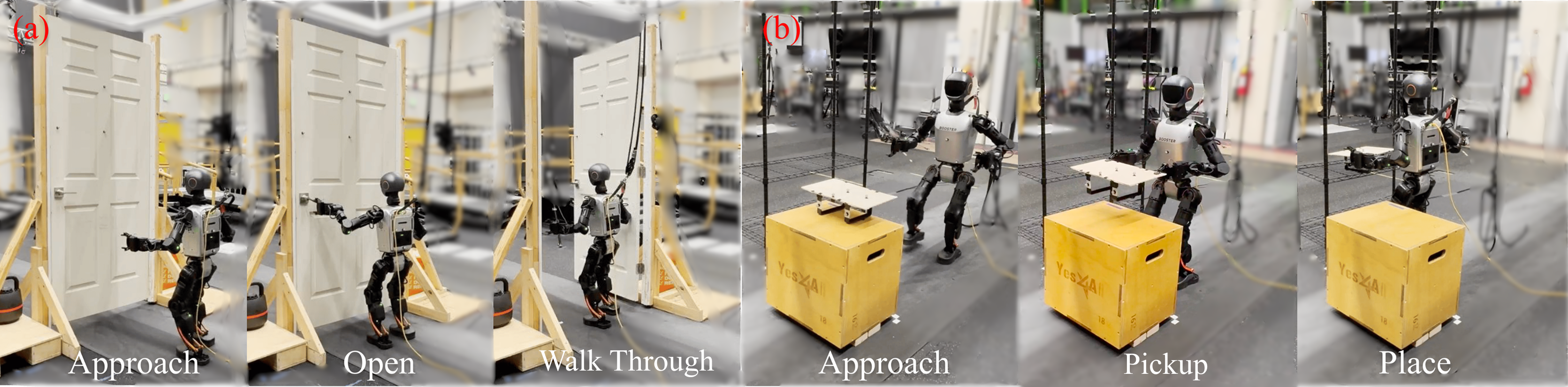}
\caption{The robot autonomously executes loco-manipulation tasks on a Booster T1 humanoid: (a) door opening and traversal, (b) long-horizon object transportation.}
\label{fig:hardware}
\vspace{-0.05in}
\end{figure}

Hardware experiments are conducted on a Booster T1 humanoid robot with 29 degrees of freedom. Both the DP motion planner $\bar{\pi}_{\theta}$ and the RL-based loco-manipulation controller $\pi_{\text{loco\_manip}}$ run on a policy computer with an AMD 7945HX CPU and an NVIDIA RTX 4060 GPU. This policy computer receives the robot's proprioceptive states (joint angles and IMU measurements) from T1's embedded computer and performs inference for $\bar{\pi}_{\theta}$. The DP outputs Cartesian action chunks for $\pi_{\text{loco\_manip}}$, which converts them into upper- and lower-body joint-position references. We use a state-based DP that relies on the pre-processed pose of the target object relative to the robot torso, captured using either a motion capture (MoCap) system streaming at 90 Hz or an onboard RGB camera. In the camera setting, a head-mounted Intel RealSense D435i provides video for 6-DoF fiducial-marker-based~\cite{olson2011apriltag} object-pose estimation at 30 Hz, demonstrating the capability to operate without privileged state information. For sim-to-real RL policy training, expert demonstration collection, and fine-tuning, we use IsaacLab~\cite{mittal2025isaaclab} as the simulator.

The DP $\bar{\pi}_{\theta}$ has an observation horizon of 8 and an action chunk size of 12, with a 0.1 s interval between each observation and action. The DP $\bar{\pi}_{\theta}$ runs at 10 Hz using NVIDIA TensorRT. The loco-manipulation policy $\pi_{\rm loco\_manip}$ runs at 50 Hz. All policies are trained on an NVIDIA H200 GPU. RL pre-training of $\pi_{\rm loco\_manip}$ takes about 10 h with 4096 parallel environments, while DP pre-training on a 100-trajectory dataset (3 h of demonstrations) requires about 18 h, for a total of 22 h of wall-clock time from pre-training to the fully fine-tuned policy. Joint optimization (\ref{alg:co-train}) uses $L=2$ iterations, each costing about 9 h, split equally between updating $\pi_{\rm loco\_manip}$ and $\bar{\pi}_{\theta}$.

As shown in \ref{fig:hardware}, we evaluate REFINE-DP on four loco-manipulation tasks. Each task requires multiple stages of walking and manipulation. Unlike prior work that requires explicit stage signals~\cite{StageACT_Lee2025_door}, our DP policy infers task stages from object locations and implicitly learns stage transitions, enabling loco-manipulation without explicit stage supervision.

\begin{itemize}
\item \textbf{Task 1 – Object pickup:} Walk to a table and pick up a box.
\item \textbf{Task 2 – Long-horizon pick-and-place:} Pick up a box and place it on another table, with a longer horizon of $40$ s and an expanded observation of the target table.
\item \textbf{Task 3 – Door opening and traversal:} Walk toward a door, grasp the handle, open, and pass through. We are the first to apply DP to humanoid door traversal.
\item \textbf{Task 4 – Stair-assisted object retrieval:} Step onto an elevated platform to retrieve an object, requiring combined locomotion over uneven terrain and manipulation.
\end{itemize}

\subsection{Baseline Methods and Ablation Study}
\label{sec:baseline}

We compare REFINE-DP with several baselines that use different planner architectures or fine-tuning strategies. Specifically, we consider three pre-trained planner backbones: (i) a diffusion policy with a transformer backbone (DiT), (ii) a recurrent LSTM diffusion policy, and (iii) a deterministic multilayer perceptron (MLP) planner; and two fine-tuning baselines: (iv) RL fine-tuning of the pre-trained MLP planner (MLP-FT) and (v) residual RL~\cite{ankile2024imitationrefinementresidual} on top of the frozen diffusion policy. These experiments benchmark the DP motion planner, with $\pi_{\text{loco\_manip}}$ remaining unchanged.

\textbf{DiT.}
The DP baseline uses the standard formulation~\cite{diffusion_policy} with a transformer backbone. This evaluates the performance of a purely pre-trained DP within our hierarchical framework.

\textbf{LSTM.}
The LSTM baseline serves as a recurrent counterpart to the transformer-based diffusion policy (DiT).

\textbf{MLP.}
The MLP planner is introduced as a separate planner baseline under the same planner-controller interface, rather than as a component inside the DiT architecture or a replacement module inside the learned denoising diffusion process. It directly regresses actions from observations and therefore produces deterministic action predictions at inference time. In contrast, the DiT-based diffusion planner samples action trajectories through a denoising diffusion process, which enables stochastic trajectory generation and can better capture multimodal action distributions.

\textbf{MLP-FT.}
MLP-FT extends the deterministic MLP baseline by fine-tuning via RL, serving as an ablation of the planner backbone. Since PPO-style policy fine-tuning relies on stochastic exploration, we convert the pre-trained deterministic MLP into a stochastic behavior policy during fine-tuning by interpreting its output as the mean action and sampling around it with an Ornstein-Uhlenbeck (OU) process. During fine-tuning, the OU process acts as a mean-reverting diffusion process, producing temporally correlated and smoothly varying perturbations that are suitable for continuous-control exploration:
\[
\mathbf{A}_t^{k-1} = (1 - \lambda_k)\mathbf{A}_t^{k}
+ \lambda_k \tilde{\boldsymbol{\mu}}_\theta(\mathbf{S}_t)
+ \tilde{\boldsymbol{\sigma}}_k \boldsymbol{\epsilon}_k,
\quad \boldsymbol{\epsilon}_k \sim \mathcal{N}(0, \mathbf{I}),
\]
where $\tilde{\boldsymbol{\mu}}_\theta(\mathbf{S}_t)$ is the MLP mean action prediction, $\lambda_k$ is the interpolation coefficient at denoising step $k$, $\tilde{\boldsymbol{\sigma}}_k$ is the exploration-noise scale, and $\boldsymbol{\epsilon}_k$ is a standard Gaussian noise vector. We adopt a linear schedule for $\lambda_k$ and a decreasing noise schedule for $\tilde{\boldsymbol{\sigma}}_k$, encouraging exploration in early steps and concentration around the mean in later steps. This design enables RL fine-tuning of the MLP baseline while preserving its deterministic structure, allowing us to isolate the contribution of the diffusion-based policy formulation from the fine-tuning procedure itself.

\textbf{Residual RL.}
Residual RL~\cite{ankile2024imitationrefinementresidual} improves a pre-trained DP by learning a lightweight corrective residual added to the output of the frozen DP. A residual Gaussian policy is optimized with PPO to predict small additive adjustments to the frozen DP's actions.

\subsection{Quantitative Results}

\begin{figure}[t]
\centering
\includegraphics[width=0.85\textwidth]{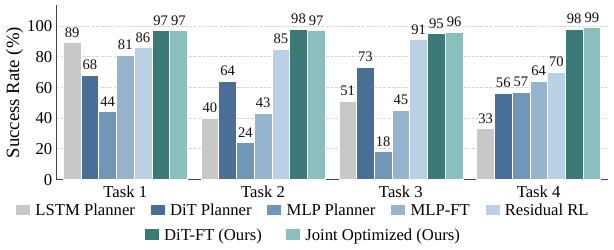}
\caption{Comparison of task success rates across all tasks and methods. REFINE-DP (fine-tuning and joint optimization) substantially outperforms all baselines, achieving over 90\% across all tasks.}
\label{fig:sr_table}
\vspace{-0.05in}
\end{figure}

\paragraph{Task Success Rate.}
Task success rate (SR) results are reported in \ref{fig:sr_table}. REFINE-DP, including both fine-tuning and joint optimization, substantially outperforms all baselines across all loco-manipulation tasks, achieving more than 90\% SR by fine-tuning from pre-trained policies of 50--70\%. Since the fine-tuned SR is already high, joint optimization does not further improve SR; instead it improves motion quality.

We observe a consistent performance improvement of the transformer-based diffusion (DiT) over the MLP-based diffusion, which we attribute to DiT's ability to model multimodal action distributions and generate diverse feasible trajectories. In contrast, MLPs perform deterministic regression and tend to average across demonstrations, producing actions that may not correspond to viable trajectories. MLP-FT partially mitigates this by injecting stochasticity during fine-tuning, but the achievable improvement remains limited by the representational capacity of the backbone. The LSTM baseline attains a success rate comparable to DiT on the short-horizon Task 1, but degrades sharply on the long-horizon tasks, indicating that its recurrent state is insufficient to capture the full action distribution over extended horizons.

Pre-training is essential for successful fine-tuning. Fine-tuning is effective when a pre-trained policy achieves a moderate success rate (approximately 50--70\%). In contrast, fine-tuning from random initialization shows limited improvement, as sparse rewards rarely provide informative learning signals when task success is infrequent.

\begin{figure}[t]
  \centering
  \includegraphics[width=0.8\textwidth]{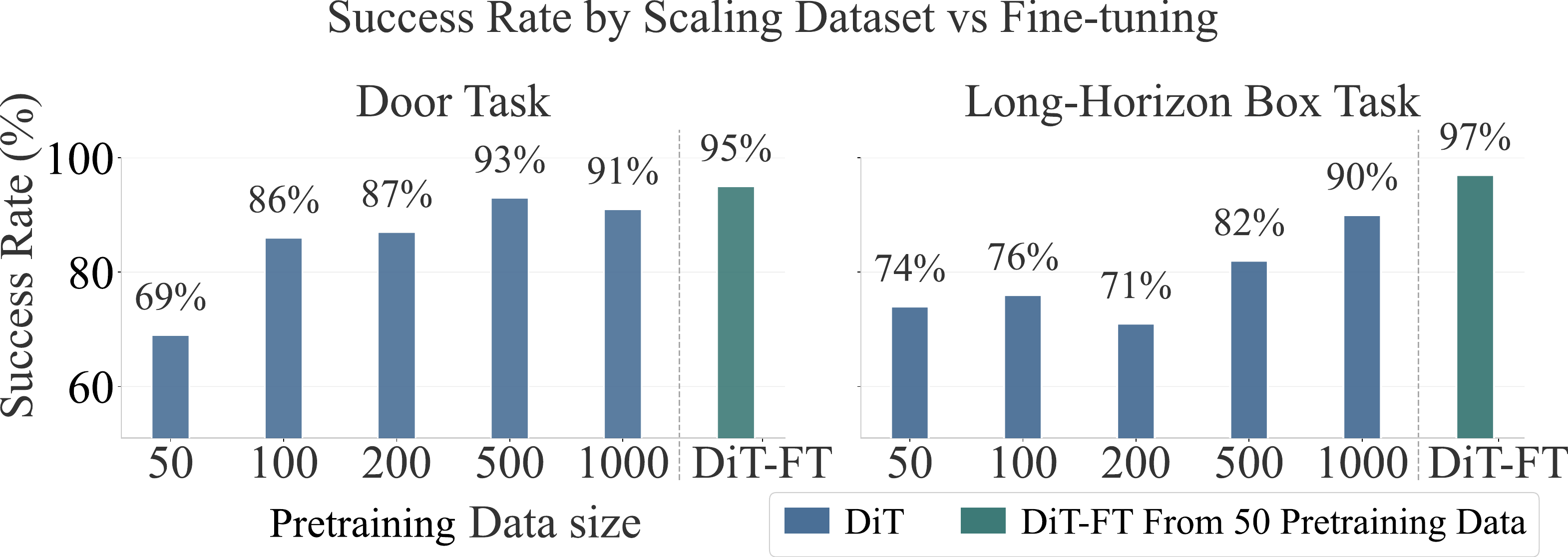}
  \caption{Comparison of success rate between scaling pre-training data and fine-tuning. Fine-tuning achieves higher SR with substantially fewer demonstrations.}
  \label{fig:scale_vs_ft}
  \vspace{-0.05in}
\end{figure}

\paragraph{Data Efficiency.}
We evaluate the data efficiency of fine-tuning by measuring SR as a function of offline dataset size. The results in \ref{fig:scale_vs_ft} show that while SR improves with more pre-training data, fine-tuning achieves higher performance with substantially fewer demonstrations. In particular, reaching 90\% SR through pre-training alone requires approximately 1000 trajectories. In contrast, a DP pre-trained on only 50 trajectories achieves up to 95\% SR after fine-tuning, demonstrating that our approach substantially reduces the amount of expensive pre-training data required while achieving a higher final SR.

\paragraph{Joint Optimization and Motion Quality.}

\begin{figure}[t]
  \centering
  \includegraphics[width=0.75\textwidth]{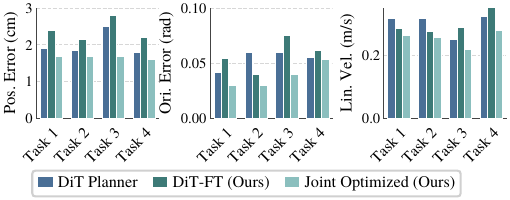}
  \caption{Joint optimization achieves better end-effector tracking and smoother velocities across all tasks. The jointly optimized controller reduces orientation error by up to 50\% compared to the pre-trained controller and decreases end-effector velocity by 15\%, indicating smoother manipulation.}
  \label{fig:door_cotraining_vs_ft}
  \vspace{-0.05in}
\end{figure}

Optimizing the loco-manipulation controller provides three benefits. First, optimizing the RL controller alone using pre-trained DP rollouts already improves SR by 18\% on the long-horizon pick-and-place task. Second, the jointly optimized controller $\pi'_{\rm loco\_manip}$ improves the training efficiency of DP fine-tuning, requiring approximately half as many iterations (20 instead of 40) to achieve 90\% SR compared to fine-tuning with the pre-trained controller. Third, joint optimization improves motion quality as shown in \ref{fig:door_cotraining_vs_ft}: $\pi'_{\rm loco\_manip}$ achieves the lowest position error and reduces orientation error by up to 50\% compared to the pre-trained controller, with an average 15\% decrease in end-effector velocity reflecting smoother manipulation. On the other hand, fine-tuning the DP alone can degrade controller tracking performance because the planner may issue excessive commands against a frozen controller, which increases command-tracking error and can produce jerky motion under sudden contact changes.

\begin{figure}[t]
  \centering
  \includegraphics[width=0.75\textwidth]{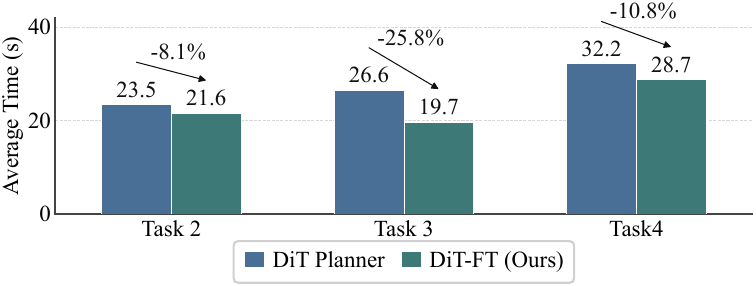}
  \caption{Fine-tuning reduces task execution time in both simulation and hardware experiments by eliminating hesitant motions.}
  \label{fig:throughput}
  \vspace{-0.05in}
\end{figure}

\paragraph{Task Completion Efficiency.}
The fine-tuned policy demonstrates higher throughput, completing tasks in less time. As shown in \ref{fig:throughput}, fine-tuning reduces task completion time by 15\% on average across Tasks 2--4 by eliminating hesitant and indecisive motions produced by the pre-trained policy.

\begin{figure}[t]
\centering
\includegraphics[width=0.9\columnwidth]{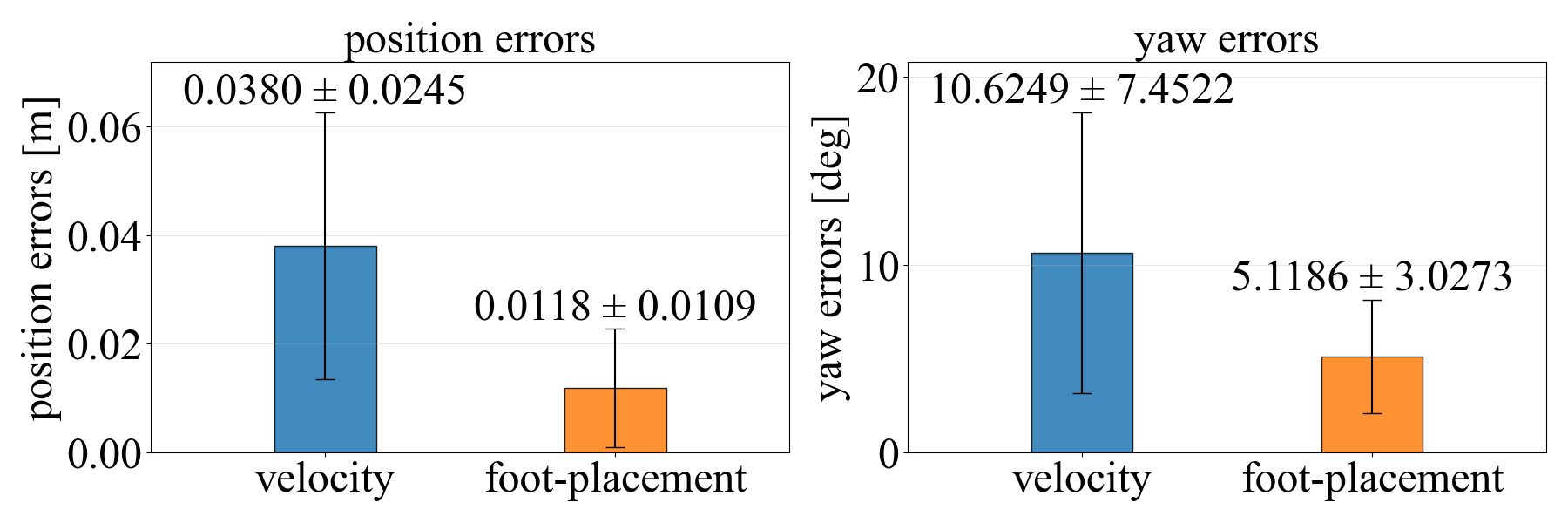}
\caption{Position and yaw tracking errors across 10 trials for velocity-based and foot-placement-based lower-body policies.}
\label{fig:vel_stair_plot}
\vspace{-0.05in}
\end{figure}

\paragraph{Foot-Placement vs. Velocity-Tracking.}
As motivated in \ref{sec:RL_train}, the loco-manipulation tasks require precise torso positioning, which the foot-placement interface enforces more directly than the velocity-tracking interface. To quantify this effect, we conduct a torso-position-tracking ablation that isolates the lower-body command interface: we keep all upper-body components fixed and use the same PD feedback controller for torso-position tracking, varying only the lower-body policy. For the foot-placement policy, the feedback base-velocity command is converted into explicit footstep targets by the velocity-to-footstep planner; for the velocity-tracking policy, the same base-velocity command is used directly. As shown in \ref{fig:vel_stair_plot}, the foot-placement policy achieves lower torso position and yaw error and more consistent target reaching across 10 trials, supporting our choice of foot-placement tracking over velocity tracking for precise loco-manipulation.

\paragraph{Out-of-Distribution Generalization.}
To extend the pre-trained policy to out-of-distribution (OOD) scenarios, we employ domain randomization to expose it to scenarios beyond those observed during pre-training. For the object transport task, we fine-tune in unseen robot initial locations parameterized by radial distance, polar angle, and heading angle relative to the target object. Fine-tuning expands the randomization ranges to 320\%, 125\%, and 600\% of the corresponding pre-training ranges, respectively. While the pre-trained policy achieves 0\% SR at the maximum randomization level, curriculum-based fine-tuning improves SR to over 80\%.

\begin{figure}[t]
  \centering
  \begin{minipage}{0.48\textwidth}
    \centering
    \includegraphics[width=\linewidth]{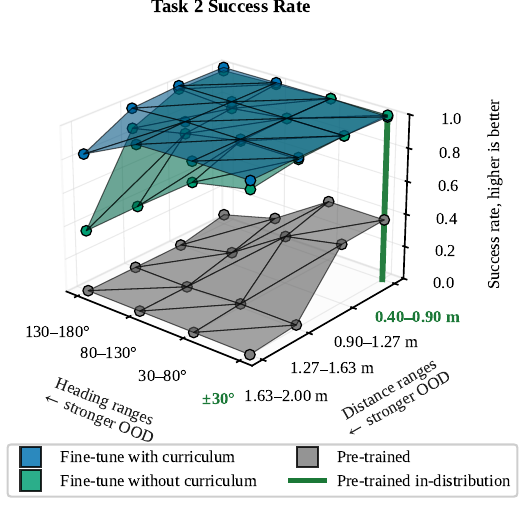}\\[2pt]
    {\small (a) Success rate}
  \end{minipage}\hfill
  \begin{minipage}{0.48\textwidth}
    \centering
    \includegraphics[width=\linewidth]{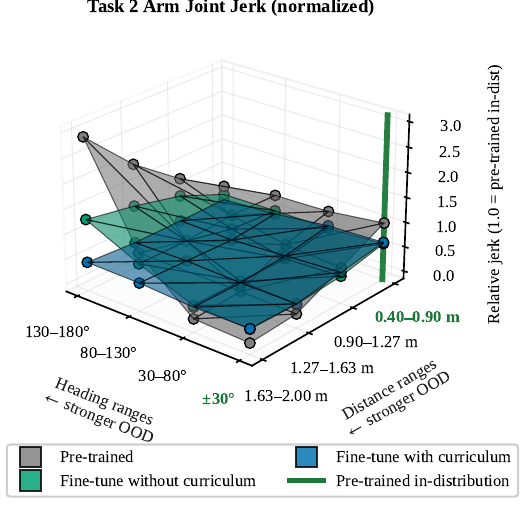}\\[2pt]
    {\small (b) Average jerk}
  \end{minipage}
  \caption{OOD generalization under different robot initialization ranges. (a) The curriculum checkpoint achieves the highest success rate in high-OOD regions. (b) The curriculum checkpoint maintains lower upper-body jerk across in-distribution and OOD regions, averaged over 100 rollout environments.}
  \label{fig:ood_sr}
  \vspace{-0.05in}
\end{figure}

\paragraph{Curriculum-Based Fine-Tuning Improves Motion Quality.}
The curriculum progressively increases the randomization ranges by 10\% once the policy reaches 90\% SR at the current level, allowing the policy to expand coverage through increasingly difficult initial-condition distributions. To evaluate this effect, we compare the pre-trained, curriculum-fine-tuned, and no-curriculum-fine-tuned checkpoints across regions of the full randomization space, defined by heading and distance intervals, using identical hyperparameters for both fine-tuned policies. As shown in \ref{fig:ood_sr}(a), directly training on the maximum randomization ranges can improve SR, but the learned arm behavior exhibits substantially more jitter in OOD scenarios because early exposure to high-randomization conditions amplifies stochastic PPO rollout actions. \ref{fig:ood_sr}(b) reports the squared sum of upper-body joint jerk averaged over 100 rollout environments: the curriculum-fine-tuned checkpoint maintains stable jerk across both in-distribution and OOD scenarios, whereas the no-curriculum-fine-tuned checkpoint exhibits much higher jerk, comparable to the pre-trained checkpoint, when deployed to OOD settings.

\subsection{Hardware Results}
\label{sec:dpft_hardware_results}

On the Booster T1 hardware, REFINE-DP achieves a success rate of 70\% (Task 1, object pickup), 50\% (Task 2, long-horizon pick-and-place), and 75\% (Task 3, door opening) over $N=20$ trials each. The main failure modes are occasional tripping of the lower-body RL controller and unsuccessful grasps on the box and door handles. The sim-to-real gap arises from two primary sources: (i) object observations exhibit systematic offsets from pose calibration and become biased under MoCap or perception occlusions, especially during contact-rich phases such as grasping a box or door handle; and (ii) sim-to-real dynamics mismatch causes locomotion trembling and sliding.

Despite these challenges, the fine-tuned policy demonstrates improved robustness to noisy observations and execution errors. REFINE-DP adapts to environmental perturbations during execution by re-adjusting and re-attempting. When the target object location is altered mid-task, the robot re-adjusts its plan on-the-fly and continues. When the robot misses the door handle, it takes small corrective footsteps to get closer before re-attempting the manipulation. Such corrective behaviors occur consistently under the fine-tuned policy, whereas the pre-trained policy often becomes stuck and fails to recover. REFINE-DP achieves average speedups of around 10\% and 20\% for the box pickup and door-opening tasks, respectively.

\paragraph{Observation-Noise Sensitivity.}
We further test whether injecting observation noise during fine-tuning improves the DP's robustness to the sim-to-real gap. We add zero-mean Gaussian noise to the observed state under several noise levels. Counterintuitively, noise with standard deviations of 1~cm and 3~cm reduces the real-world Task 1 success rate from 70\% to 25\% and 20\%, respectively. The dominant failure mode is excessive, chaotic motion in the predicted hand poses. We attribute this to the generative nature of the DP: perturbing the observed state corrupts the conditioning signal, so the planner learns to map noisy, inconsistent conditions to actions and produces unstable action predictions.

\section{Conclusion and Future Work}

REFINE-DP introduces a hierarchical fine-tuning framework that jointly optimizes a diffusion-policy planner and an RL-based loco-manipulation controller. RL fine-tuning of the DP substantially improves task success rate and data efficiency, enabling robust generalization to out-of-distribution scenarios. The joint optimization further reduces the distributional mismatch between the planner and controller, leading to higher tracking accuracy, smoother motion, and faster fine-tuning convergence. Together with PPF in \ref{chapter:PPF}, these results demonstrate that reinforcement learning fine-tuning can reduce the quantity and quality requirements of robot demonstrations while expanding pre-trained capabilities across both locomotion and autonomous loco-manipulation.

A current limitation is that the planner remains state-conditioned, which requires an explicit object pose and limits the type of objects it can handle. Future work will focus on an end-to-end RGB-conditioned diffusion-policy planner, more versatile skills in a unified whole-body controller, and a richer planner-controller interface.

Another important direction is hardware-informed fine-tuning to bridge the sim-to-real gap. Directly fine-tuning REFINE-DP on the robot remains difficult because the DPPO used in this chapter is on-policy and depends on large numbers of rollout samples; collecting such data on a balancing humanoid would be slow and safety-critical, especially during exploratory rollouts full of failed grasps and missed contacts. More sample-efficient alternatives include off-policy residual adaptation from small amounts of hardware data, reward or dynamics learning from real rollouts, and real-sim-real loops that use hardware measurements to recalibrate simulation before additional fine-tuning. 

In addition, this chapter relies on costly VR teleoperation and per-task heuristic rollouts; supplementing these with more scalable data sources, such as egocentric human data, is a promising direction for reducing the data burden.

%% file: chapters/conclusion.tex
\chapter{Conclusion}
\label{chapter:conclusion}

\section{Dissertation Contributions and Lessons Learned}

The goal of this dissertation is to advance robust, contact-aware, and autonomous humanoid loco-manipulation. To this end, it develops model-based and learning-based methods that progressively expand humanoid capabilities from robust locomotion, to contact-aware whole-body manipulation, and toward autonomous loco-manipulation. The contributions and lessons learned are summarized below.

\paragraph{Formal guarantees on locomotion robustness.}
We demonstrate that locomotion robustness can be specified, synthesized, and optimized through formal specifications and model-based optimization in LTL-BT and STL-MPC. Systematic hardware evaluation, rather than ad hoc perturbation, serves as the true touchstone for validating hardware robustness. Dynamic robust locomotion emerges from system-level design: optimization efficiency, communication latency, and state estimation accuracy. Each component across the control stack must be optimized to deliver the robust behavior on hardware.

\paragraph{Force-aware whole-body manipulation.}
We show that whole-body tactile sensing is the key to unlocking contact-rich whole-body manipulation. Human data is particularly valuable because it captures natural force interactions at scale without requiring the robot in the loop. 
However, the lack of action labels is the major bottleneck in human data. To address this challenge, we develop a human-data correction module that converts human motions into robot-executable actions while preserving the natural force interactions.
We achieve force-supervised planning and tactile-based admittance control and improve contact stability beyond motion prediction alone.

\paragraph{Reinforcement learning fine-tuning.}
We show that reinforcement learning fine-tuning provides a more efficient and effective path for improving humanoid capabilities than simply scaling demonstration data. By adapting pre-trained policies to out-of-distribution scenarios, fine-tuning expands performance beyond the coverage and limitations of the initial demonstrations. For hierarchical frameworks, we show that jointly fine-tuning the entire framework improves cohesion among components, yielding better motion quality than fine-tuning a single component alone.

\section{Broader Impacts}

This dissertation has several broader impacts for humanoid robotics research.

First, we conducted rigorous hardware experiments to validate locomotion robustness under realistic disturbances. In particular, we demonstrated the effectiveness of our approach through a field test on a real naval ship, the \textit{Stiletto}, in Virginia Beach, VA. This experiment moved beyond laboratory validation and showed that formally specified robustness can translate to challenging real-world applications.

Second, the major systems developed in this dissertation, including STL-MPC, WT-UMI, and REFINE-DP, have been open-sourced to support reproducible humanoid robotics research. Beyond algorithm development, we invested substantial effort in building tools that are useful to the broader research community. These include a whole-body controller that has inspired and accelerated subsequent research, a data collection and processing pipeline for tactile-based whole-body manipulation, and a reinforcement learning fine-tuning framework for adapting pre-trained policies to new scenarios. In addition to releasing code, we also provide datasets and pre-trained models, helping other researchers reproduce our results and build upon our work more efficiently.

Third, this dissertation is complemented by our survey paper on humanoid locomotion and manipulation~\cite{gu_humanoid_2025}, which provides a roadmap for researchers entering the rapidly evolving field of humanoid robotics. The survey organizes progress across model-based planning and control, reinforcement learning, imitation learning, foundation models, and whole-body tactile sensing, clarifying how these directions relate to humanoid loco-manipulation. By synthesizing the field's current capabilities, limitations, and open challenges, the survey helps researchers identify promising problems, compare technical approaches, and understand how individual advances fit into the broader path toward general-purpose humanoid robots.

\section{Designing Hierarchical System and the Command Interface}

Across the methods developed in this dissertation, a recurring design pattern is to separate planning and control. The planner reasons over a longer horizon, typically on the order of seconds, and produces compact motion objectives that encode task intent. The controller operates at a much shorter time scale, often at hundreds or thousands of hertz, and converts these objectives into high-dimensional joint-level commands while enforcing balance, tracking, and hardware constraints. In the locomotion thrust, this interface consists of desired center-of-mass and swing-foot trajectories; in WT-UMI and REFINE-DP, it consists of hand-pose targets, base-motion commands, and, when contact regulation is required, force references. This structure keeps high-level decision making tractable while allowing the low-level controller to absorb fast dynamics and contact disturbances.

This hierarchy is a design choice rather than a fundamental requirement. A unified end-to-end policy could, in principle, map perception and task context directly to joint-level actions, and recent work~\cite{wu2026perceptivehumanoidparkourchaining} has shown that such policies can acquire impressive humanoid behaviors. The cost, however, is reduced controllability and interpretability. When a single policy selects both the task motion and the joint-level response, it becomes harder to demand a specific behavior, isolate failure modes, or identify whether an error originates from planning, tracking, sensing, or hardware execution. The interfaces used in this dissertation are therefore chosen to be physically meaningful and directly monitorable. Center-of-mass trajectories, foot placements, hand poses, base velocities, and force references expose the intended behavior at the interface between the planner and the controller, which makes the system easier to command, debug, and improve.

There is no universal best interface. A good communication signal depends on the task, the required degree of whole-body coordination, and the available data. Full-body joint targets or body keypoints are suitable when the goal is tightly coordinated whole-body motion. Latent representations are useful for representing multiple skills in a single policy, though they sacrifice physical transparency. The interfaces adopted in this dissertation favor interpretability, simplicity, and teleoperation compatibility, which match the needs of robustness evaluation, tactile manipulation, and autonomous loco-manipulation. Future humanoid systems may increasingly blend these choices, learning richer interfaces that represent more versatile behaviors.

\section{Reflections on Humanoid System Reliability}

Across the projects in this dissertation, I have worked with a wide range of humanoid platforms, from Agility Robotics' Cassie and Digit, to Figure 02 during my internship at Figure AI, and finally to Unitree G1 and Booster T1. These experiences have made one lesson especially clear: hardware reliability cannot be taken for granted. For humanoid robots to support sustained research and real-world deployment, they must be practical and capable. They should be easy to operate without repeated calibration, reliable in remote communication, convenient to transport for field tests, transparent in their low-level torque interfaces, and compatible with widely adopted software stacks. They should also be safe for human interaction, lightweight, energy efficient, thermally stable, tolerant to impact, resilient to falls, and supported by a dependable supply of spare parts.

This perspective has also deepened my appreciation for deployment-centered system design in robotics. For a robot company, earning user trust requires more than demonstrating impressive capabilities; it requires delivering a system that users can depend on every day. Responsiveness, maintainability, serviceability, and usability are core to earning that trust.

Hardware reliability and availability have been persistent bottlenecks throughout my humanoid research. Experiments were often shaped by the pressure of hardware breakdowns, limited spare parts, and long repair cycles. These constraints were not only a blocker for individual projects, but also a universal barrier to entry into the broader humanoid research community. The lack of reliable and affordable humanoid platforms has restricted many researchers who wish to explore this field.

Looking forward, the emergence of more accessible, robust, and affordable humanoid platforms will accelerate progress in the field. As hardware becomes more reliable and widely available, researchers will be able to spend less effort maintaining the robot and more effort advancing general-purpose physical intelligence.

\section{The Future of Whole-Body Tactile Sensing}

Whole-body tactile sensing represents an important direction for future humanoid robot perception. Existing tactile-sensing research has largely concentrated on hands and fingertips, where both commercial products and in-house sensor designs have supported progress in dexterous manipulation. In contrast, tactile sensing over larger body areas, including the arms, torso, and lower body, remains considerably less mature. Commercial solutions for large-area upper-body tactile sensing are still limited, and custom-built systems remain costly and labor-intensive. For example, the complete WT-UMI sensing suit costs approximately \$4,000, illustrating the current gap between laboratory prototypes and scalable deployment.

Large-area tactile sensing is particularly important for whole-body manipulation, where contacts often occur beyond the hands. Tasks such as lifting heavy objects, transporting bulky items, collaborating with humans, and interacting with cluttered environments require robots to perceive and regulate distributed physical contact. In these settings, tactile sensing can transform contact from an unmodeled disturbance into a structured source of feedback. A humanoid robot equipped with whole-body tactile sensing can estimate contact locations and interaction forces, detect slips and impacts, regulate distributed loads, and coordinate locomotion and manipulation through continuous physical feedback.

In the long term, tactile sensing may become a standard modality for humanoid robots, comparable in importance to vision. Covering larger portions of the robot body with tactile sensors would provide a physical feedback layer for human-level loco-manipulation, supporting stability, safety, and force-aware interaction during complex tasks. However, several technical challenges must be addressed before whole-body tactile sensing can become widely deployable. Large sensor arrays require reliable calibration, synchronization, and drift mitigation over long-duration operation. Sensors placed on the lower body must tolerate high loads, impacts, abrasion, and repeated deformation. System designers must also determine whether tactile sensors should be embedded into the robot structure, integrated as compliant artificial skin, or implemented as modular wearable layers. In addition, scalable manufacturing, low-cost maintenance, and easy replacement are necessary for practical deployment.

These open questions define a critical research frontier for humanoid robotics. Progress in scalable, durable, and affordable tactile systems would substantially expand the role of physical feedback in robot autonomy. Whole-body touch has the potential to become a foundational sensing modality through which humanoid robots perceive contact, regulate force, and adapt their behavior during real-world loco-manipulation.

\section{Limitations and Future Directions}

The model-based locomotion methods in this dissertation provide structured robustness guarantees, but these guarantees remain tied to modeling assumptions and bounded perturbation sets. LTL-BT reasons over a discretized keyframe space, and STL-MPC optimizes robustness with reduced-order dynamics before whole-body execution. These abstractions make real-time planning tractable, but they cannot represent every contact mode, actuator limit, sensing error, or terrain interaction encountered on hardware. Future work should more tightly integrate formal specifications with full-body dynamics models, so that formal robustness measures more accurately reflect the behavior of the full-order robot interacting with its environment.

The whole-body manipulation and loco-manipulation policies developed in this dissertation remain task-specific. They have not yet achieved versatility, defined here as the ability to accomplish diverse tasks under a single set of policy weights. Future work will explore how multiple skills can be integrated into a unified policy that generalizes across tasks, objects, and environments. One straightforward direction is to use a task planner that orchestrates multiple task-specific policies. Another possibility is to train a mixture-of-experts policy distilled from several specialized policies. A more ambitious direction is to train a single versatile policy directly from a mixed dataset of diverse task demonstrations. This direction is particularly challenging because it requires the policy to learn a shared representation across heterogeneous tasks and adapt its behavior based on the current context.

The current loco-manipulation skills are not yet fully whole-body. Locomotion and manipulation are often separated into distinct stages, which reduces execution efficiency and produces behavior that is less human-like. In contrast, humans naturally coordinate walking and manipulation simultaneously. Future research should investigate how humanoid robots can achieve simultaneous locomotion and manipulation through tightly coupled whole-body control and motion planning. One potential direction is to redesign the whole-body controller interface, such as using whole-body joint targets or body keypoints. During teleoperation, human whole-body poses could then serve as the command input. The planner would also need to learn a high-dimensional whole-body action space following the same controller interface.

Hardware fine-tuning remains challenging because DPPO is an on-policy algorithm and therefore requires substantial interaction data. As a result, fine-tuning in this dissertation is restricted to simulation, while direct hardware fine-tuning remains costly and risky for full-size humanoids. Future work should investigate data-efficient adaptation methods that require only limited real-world trials, such as off-policy reinforcement learning, real-to-sim fine-tuning, and safe exploration strategies. This direction is especially important for long-horizon loco-manipulation, where hardware data are scarce, task failures can be expensive, and repeated rollouts may impose significant wear on the robot.

Although this dissertation corrects human demonstrations into robot-executable actions, the corrected data still depends on a small amount of teleoperation data. Future work should reduce or eliminate this dependence, since teleoperation is often less natural and contact-rich than direct human demonstration. One promising direction is to incorporate vision-based human pose estimation, which could provide a more scalable and cost-effective way to capture natural whole-body manipulation behaviors without requiring robot-in-the-loop teleoperation.

Sensing remains another important bottleneck for autonomous loco-manipulation. WT-UMI demonstrates the value of tactile force information, but richer physical interaction will require broader tactile coverage across the whole body, multi-axis force estimation, and tighter vision-tactile integration. These sensing capabilities would support more diverse contact modes and more anticipatory planning before contact occurs. When integrated with autonomous loco-manipulation policies, they could help robots decide where and how to establish contact, regulate distributed forces during interaction, and recover when contact unexpectedly breaks or emerges.

Autonomy also still relies on structured state observations and task-specific training. Future humanoid systems should integrate perception-conditioned policies, task-level input, and closed-loop recovery so that robots can decide plans from visual and tactile context, sequence multiple skills, and update plans when objects move, grasps fail, or contacts change. This direction will require tighter integration between foundation-model-style semantic reasoning, a versatile motion planner, and a whole-body loco-manipulation controller that respects balance, contact, and hardware constraints.

\section{Closing Remarks}

The central arc of this dissertation is the conversion of structured knowledge and experience into deployable humanoid capability. For instance, formal specifications and dynamics models provide a mechanism for specifying and optimizing robust locomotion; human demonstrations and whole-body tactile sensing provide force-aware supervision for contact-rich manipulation; robot demonstrations and reinforcement learning fine-tuning provide a path from pre-trained behaviors to autonomous loco-manipulation. Together, these contributions advance humanoid robots from robust mobility, to contact-aware whole-body interaction, and toward autonomous loco-manipulation in real-world environments.

The long-term challenge is to unite versatile skills of walking and manipulation, continually acquire physical intelligence through interaction with the world, and reason over long-horizon tasks fully autonomously. The methods developed in this dissertation pave a way toward this vision.